\documentclass[11pt, logo, copyright]{deepmind}
\usepackage{kantlipsum, lipsum}
\usepackage{dm-colors}

\usepackage[title]{appendix}
\usepackage{graphicx}
\usepackage{subfigure}
\usepackage{booktabs} 
\usepackage{overpic}
\usepackage{amssymb}
\usepackage{pstricks, pst-node}
\usepackage{verbatim}
\usepackage{floatrow}
\usepackage{tikz}
\usepackage{xspace}
\usepackage{tikz-cd}
\usepackage{tkz-graph}
\usepackage{color}
\usepackage{amsmath}
\usepackage{titletoc}
\usepackage[colorlinks=true, allcolors=blue]{hyperref}
\usetikzlibrary{fit,positioning,bayesnet}

\newcommand{\dermassist}{\textsc{DermAssist}\xspace}
\newcommand{\enceladus}{\textsc{Enceladus}\xspace}
\newcommand{\narvi}{\textsc{Narvi}\xspace}
\newcommand{\aegir}{\textsc{Aegir}\xspace}

\newcommand{\datasettrain}{$D_\texttt{train}$\xspace}
\newcommand{\ptrain}{$p_\texttt{train}$\xspace}

\definecolor{TartOrange}{HTML}{ff2e35}
\definecolor{Orange}{HTML}{ff7825}
\definecolor{Mango}{HTML}{ffc013}
\definecolor{AppleGreen}{HTML}{7cb81b}
\definecolor{Blue}{HTML}{1173b0}
\definecolor{BdazzledBlue}{HTML}{2e58a5}
\definecolor{Purple}{HTML}{5b3590}
\definecolor{Sunglow}{HTML}{FFCA3A}
\definecolor{TableRow}{gray}{0.9}
\newcommand{\addcomment}[3]{{\color{#1}\textsuperscript{\textbf{#3:}}#2}}

\newcommand{\olivia}[1]{\addcomment{red}{#1}{Olivia}}

\usepackage{mathtools}
\usepackage{dsfont}
\usepackage[dvipsnames]{xcolor}
\usepackage[colorinlistoftodos]{todonotes}
\usepackage{booktabs}
\usepackage{xfrac}
\usepackage{bbm}

\usepackage{algorithm}
\usepackage{algorithmicx}

\usepackage[most]{tcolorbox}
\usepackage{xparse}
\usepackage{lipsum}
\usepackage{changepage}
\usepackage{enumitem}

\newcommand{\squishlist}{
   \begin{list}{$\bullet$}
    { \setlength{\itemsep}{0pt}      \setlength{\parsep}{3pt}
      \setlength{\topsep}{3pt}       \setlength{\partopsep}{0pt}
      \setlength{\leftmargin}{1.5em} \setlength{\labelwidth}{1em}
      \setlength{\labelsep}{0.5em} } }

\newcommand{\squishlisttwo}{
   \begin{list}{$\bullet$}
    { \setlength{\itemsep}{0pt}    \setlength{\parsep}{0pt}
      \setlength{\topsep}{0pt}     \setlength{\partopsep}{0pt}
      \setlength{\leftmargin}{2em} \setlength{\labelwidth}{1.5em}
      \setlength{\labelsep}{0.5em} } }

\newcommand{\squishend}{
    \end{list}  }

\usepackage{amsthm}

\newcommand{\x}{x}
\newcommand{\z}{z}
\newcommand{\vx}{\mathbf{x}}

\newcommand{\y}{y}

\newcommand{\two}{(2)}

\newcommand{\va}{\myvec{a}}

\DeclareMathAlphabet{\mathpzc}{OT1}{pzc}{m}{n}

\usepackage{amsmath,amsfonts,bm}

\def\eqref#1{equation~\ref{#1}}

\def\1{\bm{1}}

\def\va{{\bm{a}}}

\def\vx{{\bm{x}}}

\DeclareMathAlphabet{\mathsfit}{\encodingdefault}{\sfdefault}{m}{sl}
\SetMathAlphabet{\mathsfit}{bold}{\encodingdefault}{\sfdefault}{bx}{n}

\def\sE{{\mathbb{E}}}

\usepackage[authoryear, sort&compress, round]{natbib} 

\graphicspath{{figures/}}

\title{Generative models improve fairness of medical classifiers under distribution shifts}

\keywords{diffusion models, augmentations, out-of-distribution generalization, statistical fairness} 
\paperurl{deepmind.com/papers/2023/genmodels.pdf}

\renewcommand{\today}{2023-04-18} 

\author[*,1]{Ira Ktena}
\author[*,1]{Olivia Wiles}
\author[1]{Isabela Albuquerque}
\author[1]{Sylvestre-Alvise Rebuffi}
\author[1]{Ryutaro Tanno}
\author[2]{Abhijit Guha Roy}
\author[2]{Shekoofeh Azizi}
\author[1]{Danielle Belgrave}
\author[1]{Pushmeet Kohli}
\author[2]{Alan Karthikesalingam}
\author[1]{Taylan Cemgil}
\author[1]{Sven Gowal}

\affil[*]{Equal contributions}
\affil[1]{DeepMind}
\affil[2]{Google}

\begin{abstract}
A ubiquitous challenge in machine learning is the problem of domain generalisation. This can have serious implications as it can exacerbate bias against groups or labels that are underrepresented in the datasets used for model development. Model bias can lead to unintended harms, especially in safety-critical applications such as healthcare. Furthermore, the challenge is compounded by the difficulty of obtaining labelled data due to high cost or lack of readily available domain expertise. In our work, we show that learning realistic augmentations automatically from data is possible in a label-efficient manner using generative models (e.g., diffusion probabilistic models). In particular, we leverage the higher abundance of unlabelled data to capture the underlying data distribution of different conditions and subgroups for an imaging modality. By conditioning generative models on appropriate labels (e.g., diagnostic labels and / or sensitive attribute labels), we can steer the distribution of synthetic examples according to specific requirements. We demonstrate that these learned augmentations can surpass heuristic, manually implemented ones by making models more robust and statistically fair in- and out-of-distribution. To evaluate the generality of our approach, we study three distinct medical imaging contexts of varying difficulty: (i) histopathology images from a publicly available and widely adopted generalisation benchmark, (ii) chest X-rays from publicly available clinical datasets, and (iii) dermatology images characterised by complex shifts and imaging conditions. The latter constitutes a particularly unstructured domain with various challenges. Two of these imaging modalities further require operating at a high-resolution, which requires developing faithful super-resolution techniques to recover fine details of each health condition. Complementing real training samples with synthetic ones improves the robustness of models in all three medical tasks and increases fairness by improving the accuracy of clinical diagnosis within underrepresented groups. Our proposed approach leads to stark improvements out-of-distribution across modalities: 7.7\% prediction accuracy improvement in histopathology, 5.2\% in chest radiology with 44.6\% lower fairness gap and a striking 63.5\% improvement in high-risk sensitivity for dermatology with a $7.5\times$ reduction in fairness gap.
\end{abstract}

\begin{document}

\maketitle

\section{Introduction}
The advent of machine learning (ML) in healthcare has led to many advances in various facets of care and in a wide range of applications~\citep{Esteva17,Ardila19,DeFauw18}.
For example, in dermatology, skin disease impacts at least 30\% of people worldwide \citep{Hay2014}, leading to 85 million doctor visits a year and a cost of \$75 billion~\citep{Lim17} in the US alone.
AI dermatological tools (e.g., \citealp{Esteva17,Liu20}) have the potential to allow patients to assess their conditions better and improve diagnostic accuracy \citep{Jain21}. Similarly, ML technologies have unlocked new capabilities in computational pathology which have the ability to handle the gigantic quantity of data created throughout the patient care lifecycle and improve classification, prediction, and prognostication of diseases~\citep{cui2021artificial}. These solutions are often motivated by the global shortage of expert clinicians, e.g., in the case of radiologists~\citep{rimmer2017radiologist}, and demonstrate that machine learning models can facilitate detection of conditions~\citep{rajpurkar2017chexnet}. Despite these rapid methodological developments and the promise of transformative impact in different areas of healthcare~\citep{liu2019comparison}, few of these approaches (if any) have achieved the ambitious goal of fostering clinical progress~\citep{varoquaux2022machine}. As~\cite{wilkinson2020time} highlight, only 24\% of published studies evaluate the performance of their proposed algorithms on external cohorts or compare this out-of-sample performance with that of clinical experts. Many studies do not validate the efficacy of algorithms in multiple settings and, the ones that do, often perform poorly when introduced to new environments not represented in the training data. 

Building a method that is robust across populations and subgroups, such that model performance does not degrade and benefits can be transferred when applied across groups, is a non-trivial task. This is due to data scarcity~\citep{castro2020causality}, challenges in the acquisition strategies of evaluation datasets, and the limitations of evaluation metrics. We list a number of key challenges here.
\textbf{(i)} \textit{Disease prevalence} may differ between demographic subgroups.
For example, melanoma is 26\% more likely to occur in white patients than black patients \citep{AmericanCancer}. Additionally, disease prevalence in the training data may not be reflective of the general population~\citep{Kaushal20}, which can be particularly problematic when due to disparities in access to healthcare~\citep{Khan21}.
However, over-reliance on such an attribute may lead to the model learning spurious correlations between those features and the diagnostic label or relying on `shortcuts'~\citep{Degrave2021,brown2022detecting}.
\textbf{(ii)} \textit{Data scarcity}. 
While we may be able to mitigate poor performance among subgroups or new domains by collecting more data, this can be infeasible due to disease scarcity, at odds with protecting patients privacy or just not sufficient for better and more generalizable solutions~\citep{varoquaux2022machine}.
\textbf{(iii)} \textit{Poor evaluation datasets}. It is vital to evaluate methods on datasets that reflect realistic shifts. For example, a population shift~\citep{Quinonero2008dataset} can lead to performance drop of the machine learning model across environments. In a realistic setting, we have limited control over the complexity of the shifts that arise, as shifts over multiple axes can occur simultaneously (e.g., both the acquisition protocol and the population at a new hospital may be different in a new geographic location). In healthcare, machine learning systems are often trained on data from a limited number of hospitals with the hope that they will generalise well to new unseen sites. However, if we focus on simpler synthetic settings, our conclusions may not generalise as demonstrated by \cite{Gulrajani20,Wiles21}.
\textbf{(iv)} \textit{High performance on overall accuracy metrics}, while important to track, \textit{may not always expose subtle problems}. For instance, it is possible to improve on top-1 accuracy by improving performance of the most prevalent class at the expense of performance on the minority classes. 

\begin{figure}[t]
\includegraphics[clip, trim=0.0cm 2cm 0.0cm 1cm, width=\textwidth]{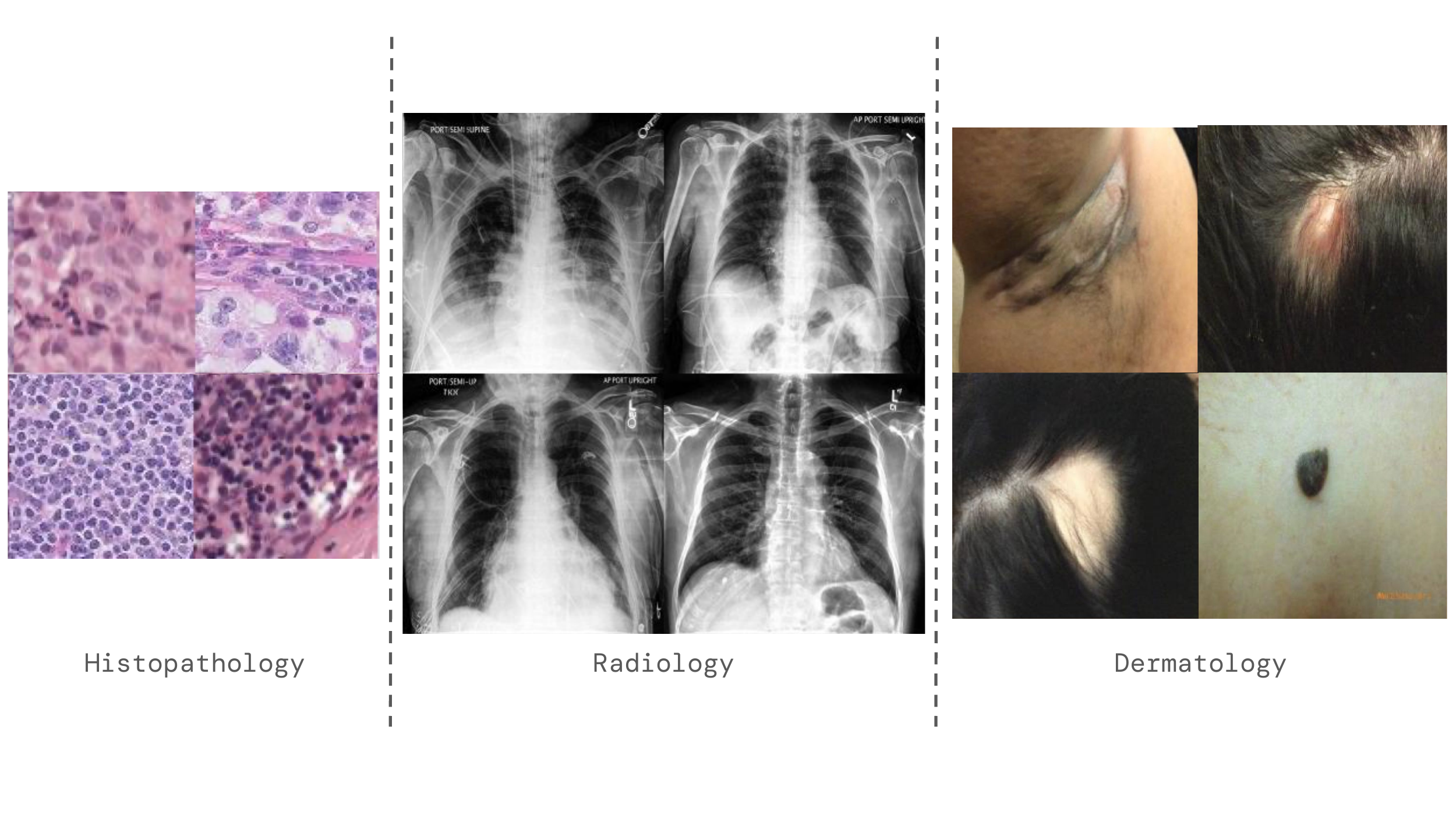}
\caption{Samples generated by our conditional diffusion model for different imaging modalities.}
\label{fig:all_generated_modalities}
\end{figure}

Prior work has shown that a developed model may perform unexpectedly poorly on underrepresented populations or population subgroups in radiology~\citep{Larrazabal20,Seyyed21}, histopathology~\citep{yu2018classify} and dermatology~\citep{abbasi2020risk}. However, the issues of robustness to distribution shifts and statistical fairness have rarely been tackled together. In this work, we leverage generative models and potentially available unlabelled data to capture the underlying data distribution and \textit{augment} real samples when training diagnostic models across these three modalities. We show that combining synthetic and real data can lead to significant improvements in top-level performance, while closing the fairness gap with respect to different sensitive attributes under distribution shifts\footnote{We borrow the term \textit{sensitive attribute} from the fairness literature to describe demographic attributes we want the model to be fair against. All of the data used in this research was de-identified before DeepMind and Google gained access to it.}. Finally, we show that diffusion models are able to generate high quality images (see~\autoref{fig:all_generated_modalities}) across modalities and perform an in-depth analysis to shed light on the mechanisms that improve generalization capabilities of the downstream classifiers.

\section{Background}

Generative models, especially generative adversarial networks (GANs)~\citep{Goodfellow14}, have been employed by various studies to improve performance in different medical imaging tasks~\citep{Frid18,Ju21,Li19,Baur18,Rashid19}  and, in particular, for underrepresented conditions~\citep{Han20,Havaei21}. GAN-based augmentation techniques have not only been used for whole-image downstream tasks, but also for pixel-wise classification tasks~\citep{Zhao18,Uzunova20} with a more thorough review of those techniques provided by~\cite{Chen22}. Data obtained by exploring different latent image attributes through a generative  model has also been shown to improve adversarial robustness of image classifiers~\cite{Gowal_2020_CVPR}.

More recently, diffusion probabilistic models (DDPM)~\citep{Ho20, Nichol21,Nichol2021glide,Ho22} presented an outstanding performance in image generation tasks and have been probed for medical knowledge by~\cite{Kather22} in different medical domains. Other works extended diffusion models to 3D MR and CT images~\citep{khader2022medical} and demonstrated that they can be conditioned on text prompts for chest X-ray generation~\citep{chambon2022roentgen}. Given the ethical questions around the use of synthetic images in medicine and healthcare~\citep{Kather22,Chen21}, it is important to make a distinction between using generative models to \textit{augment} the original training dataset and \textit{replacing} real images with synthetic ones, especially in the absence of privacy guarantees. None of these works claims that the latter would be preferable, but rather come to the rescue when obtaining more abundant real data is either expensive or infeasible (e.g. in the case of rare conditions), even if this solution is not a panacea~\citep{Zhang22}. It is worth noting that while some studies view generative models as a means of replacing real data with `anonymized' synthetic data, we abstain from such claims as greater care needs to be taken in order to ensure that generative models are trained with privacy guarantees, as shown by~\cite{carlini2023extracting, Somepalli2022diffusion}.

Recently, machine learning systems used for computer-aided diagnosis and clinical decision making have been scrutinized to understand their effect on sub-populations based on demographic or socioeconomic traits. Studies led by~\cite{Larrazabal20,Seyyed21,Puyol22} have investigated and identified discrepancies across groups based on gender / sex, age, race / ethnicity and insurance type (as a proxy of socioeconomic status), as well as their intersections.~\cite{Gianfrancesco18} performed a similar analysis for models operating on electronic health records. When evaluating machine learning systems in terms of certain fairness criteria, it is important to keep in mind that ensuring fairness in a source domain does not guarantee fairness in a different target domain under significant distribution shifts~\citep{Schrouff22}. Last but not least, there are multiple definitions of fairness in recent literature and different fairness metrics are often at odds with each other as noted by~\cite{Ricci22}. A more thorough review of related work is provided in~\autoref{sec:related_work}. In this work, we evaluate fairness both in- and out-of-distribution and aim to improve on fairness metrics without compromising top-level accuracy across all settings.




\section{Results}

\subsection{Overview of the proposed approach and experimental setting}

\begin{figure}[t]
\includegraphics[width=\textwidth]{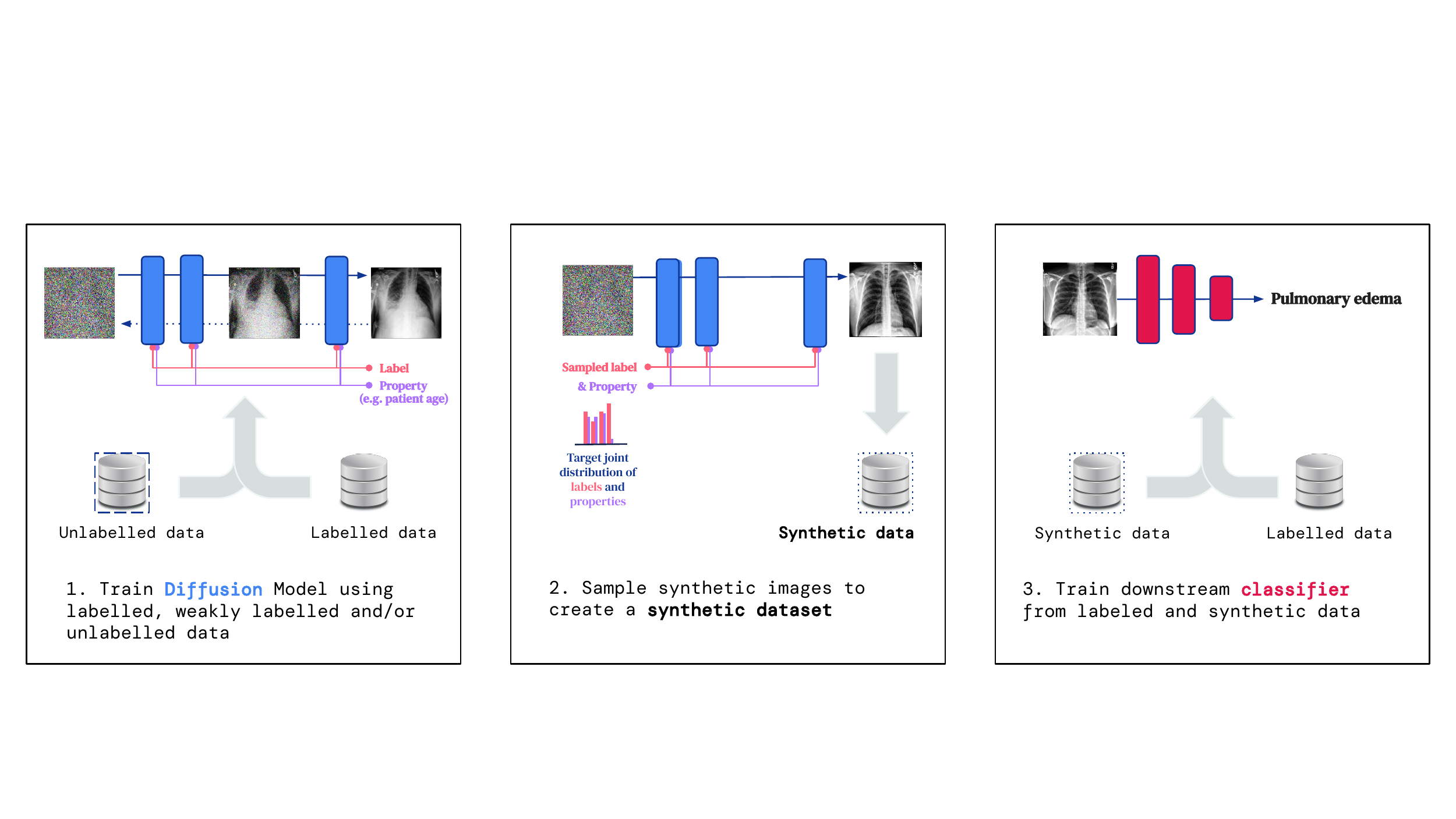}
\caption{\textbf{Method overview.} In the proposed approach we first train a diffusion model on both labelled and unlabelled data (if available). In a general setting, unlabelled data may comprise of in- or out-of-distribution data (e.g., from an unseen hospital) for which we do not have expert labels. Subsequently, we sample synthetic images from the diffusion model according to particular specifications (e.g., an image of a female individual with pulmonary edema). Finally, we train a downstream classifier on a combination of the real labelled images and the synthetic images sampled from the diffusion model.}
\label{fig:overview}
\end{figure}

Our proposed approach (illustrated in~\autoref{fig:overview}) leverages generative models for learning augmentations of the data to improve robustness and fairness of medical machine learning models. It comprises three main steps: \textbf{(1)} We train a generative diffusion model given the available labelled and unlabelled data; we assume that labelled data is available only for a single, source domain, while additional unlabelled data can be from any domain (in- or out-of-distribution). We either condition the generative model only on the diagnostic label or on both the diagnostic label and a property (e.g., hospital id or sensitive attribute label). If high-resolution images are required ($>96 \times 96$ resolution), we further train an upsampling diffusion model in a similar manner. It is worth highlighting that both the low-resolution generative model and the upsampler are trained with the same conditioning vector (i.e. either with label or label \& property conditioning). \textbf{(2)} We sample from the generative model according to a fair sampling strategy. To do this, we sample uniformly from the sensitive attribute distribution, and preserve the original diagnostic label distribution in order to preserve the original disease prevalence. Sampling multiple times from the generative model allows us to obtain different augmentations for a given condition (and property) and increase the diversity of training samples for the downstream classifier. \textbf{(3)} We combine the synthetic images sampled from the generative model with the labelled data from the source domain and train a downstream classifier. We treat the mixing ratio of real to synthetic data as a hyperparameter that is application- and modality-specific. The classifier may have multiple heads and a shared backbone in the scenario where we require a separate prediction per diagnostic class.

\paragraph{Experimental protocol.}
We evaluate this approach using diffusion probabilistic models on different medical contexts and track top-level performance (e.g. accuracy) and fairness (when relevant) in- and out-of-distribution. The evaluation on out-of-distribution datasets is equivalent to developing a machine learning model on a certain population (e.g., from a particular hospital or geographic location) and testing its performance on a population from an unseen hospital or acquired under novel conditions. In all contexts, we consider the strongest and most relevant heuristic augmentations as a baseline. It is worth noting that these augmentations (heuristic or learned) can be combined with any alternative learning algorithm that aims to improve model generalization. For the sake of our experiments we use empirical risk minimization (ERM)~\citep{Vapnik1991principles}, as there exists no single method found to consistently outperform it under distribution shifts~\citep{Wiles21}. Even though our experiments and analysis focus on diffusion probabilistic models for generation, any conditional generative model that produces high-quality and diverse samples can be used.

\paragraph{Evaluation metrics.}
To measure the performance of the different baselines and the proposed method we use two sets of metrics: one set is more focused on accuracy (i.e. top-1 accuracy for histopathology, ROC-AUC for radiology and high-risk sensitivity for dermatology), while the second set is more geared towards fairness. The performance metrics vary depending on the classification task performed for each modality (i.e. binary vs. multi-class vs. multi-label) and consider label imbalance (tracking raw accuracy in a heavily imbalanced binary setting is not very insightful). For fairness we look at the performance gap (depending on the performance metric of interest) in the binary attribute setting and the difference between the worst and best subgroup performance for categorical attributes. For continuous sensitive attributes, like age, we discretize them into appropriate buckets (specified in~\autoref{sec:datasets}).

\subsection{Clinical Tasks and Datasets}

\subsubsection{Histopathology}

The first setting that we consider is histopathology. Different staining procedures followed by different hospitals lead to distribution shifts that can challenge a machine learning model that has only encountered images from a particular hospital. The CAMELYON17 challenge by~\cite{Bandi2018detection} aims to improve generalization capabilities of automated solutions and reduce the workload on pathologists that have to manually label those cases. The corresponding dataset contains whole-slide images from five different hospitals and the task is to predict whether the histological lymph node sections captured by the images contain cancerous cells, indicating breast cancer metastases. Two of the hospital datasets provided by the challenge are held-out for out-of-distribution evaluation and three are considered in-distribution, because they use similar staining procedures. We consider this as the simplest setting for our experiments, because there are no extreme prevalence or demographic shifts.  Additionally, the considered image resolution ($96 \times 96$) is smaller in comparison to the imaging modalities presented later, which allows generation directly at that resolution without requiring an upsampler. The labelled dataset contains $455,954$ images, while the unlabelled dataset contains $1.8$ million images from the three training hospitals; full statistics are given in \autoref{tab:camelyondatasetstatistics}.
The unlabelled dataset {\em does} contain the hospital identifier, but not the diagnostic label.

\begin{figure}[t]
    \centering
    \subfigure[In-distribution results]{ \includegraphics[width=0.62\linewidth]{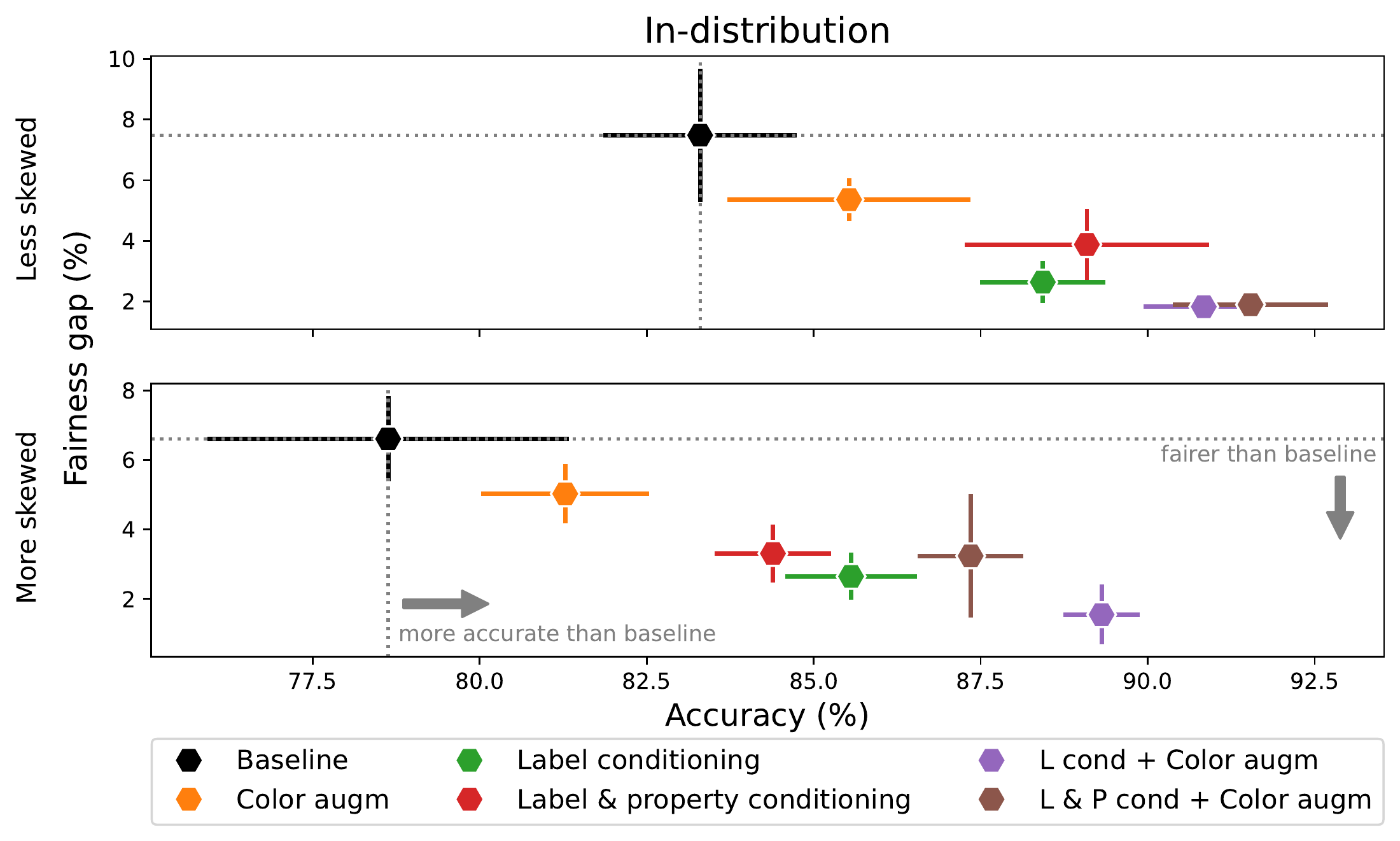}}
    \subfigure[Out-of-distribution results]{ \includegraphics[clip, trim=0.0cm 6.2cm 0.0cm 0.0cm,width=0.35\linewidth]{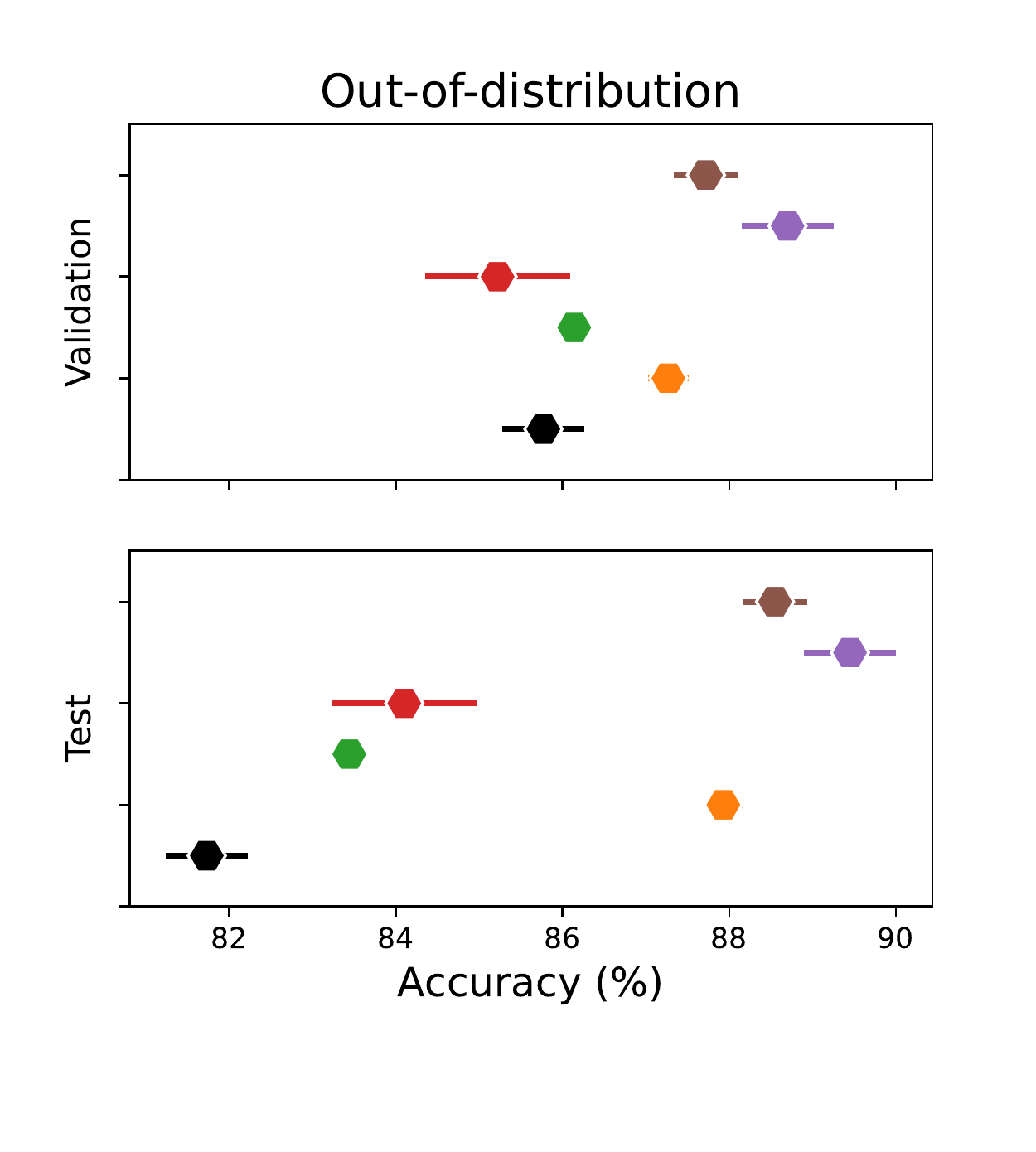}}
    \caption{(a) In-distribution fairness gap (in percentage) between the best and worst performing hospital vs. overall prediction accuracy for the presence of breast cancer metastases in histopathology images. (b) Prediction accuracy (x-axis) on the validation and test hospitals when training the generative model on all in-distribution labelled examples (the y-axis corresponds to method index). Note that the validation set is used for model selection, given that its distribution is more similar to the training distribution. We compare the following methods: \textit{Baseline} model with no augmentations; \textit{Color augm} for a model that uses color augmentations; \textit{Label conditioning} and \textit{Label \& property conditioning} for our proposed approach of a generative model conditioned on the diagnostic label and both the diagnostic label and the hospital id, respectively; \textit{L cond + Color augm}, \textit{L \& P cond + Color augm} for applying color augmentations on the images generated with diffusion models. Combining color augmentation with synthetic data performs best across all settings.}
    \label{fig:camelyon_results}
\end{figure}

In order to understand the impact of the number of labelled examples on fairness and overall performance, we create different variants of the labelled training set, where we vary the number of samples from two of the three training hospitals (3 and 4). The number of labelled examples from one hospital remains constant. For each setting, we train a diffusion model using the labelled and unlabelled dataset (using only the diagnostic label whenever available in one case, and the diagnostic label together with the hospital id in the other case).
We, subsequently, sample synthetic samples from the diffusion model and train a downstream classifier that we evaluate on the held out in- and out-of-distribution datasets (results shown in~\autoref{fig:camelyon_results}).
We compare top-level classification accuracy and fairness gap, i.e. best-to-worst accuracy gap between the in-distribution hospitals to different baselines (more details about baselines are provided in~\autoref{subsec:baselines}).
We find that using synthetic data outperforms both baselines in-distribution in the less skewed (with 1000 labelled samples from hospitals 3, 4) and more skewed setting (with only 100 labelled samples) while closing the performance gap between hospitals. We obtain the best accuracy out-of-distribution when using all in-distribution labelled examples as shown in~\autoref{fig:camelyon_results}b (in the OOD setting there is one validation and one test hospital so we do not report a performance gap). 
We find that performing color augmentation on top of the generated samples generalizes best overall, leading to a $7.7\%$ absolute improvement over the baseline model on the test hospital.
This validates that indeed we can use synthetic data to better model the data distribution and outperform variants using real data alone. We also observe that this method is most effective in the low-data regime (i.e. more skewed setting in~\autoref{fig:camelyon_results}a). In~\autoref{fig:generated_image_clinical} we show some examples of healthy and abnormal histopathology images generated at $96 \times 96$ resolution.

\subsubsection{Chest Radiology}

\begin{figure}[t]
    \centering
    \subfigure[In-distribution condition co-occurrence]{ \includegraphics[width=0.49\linewidth]{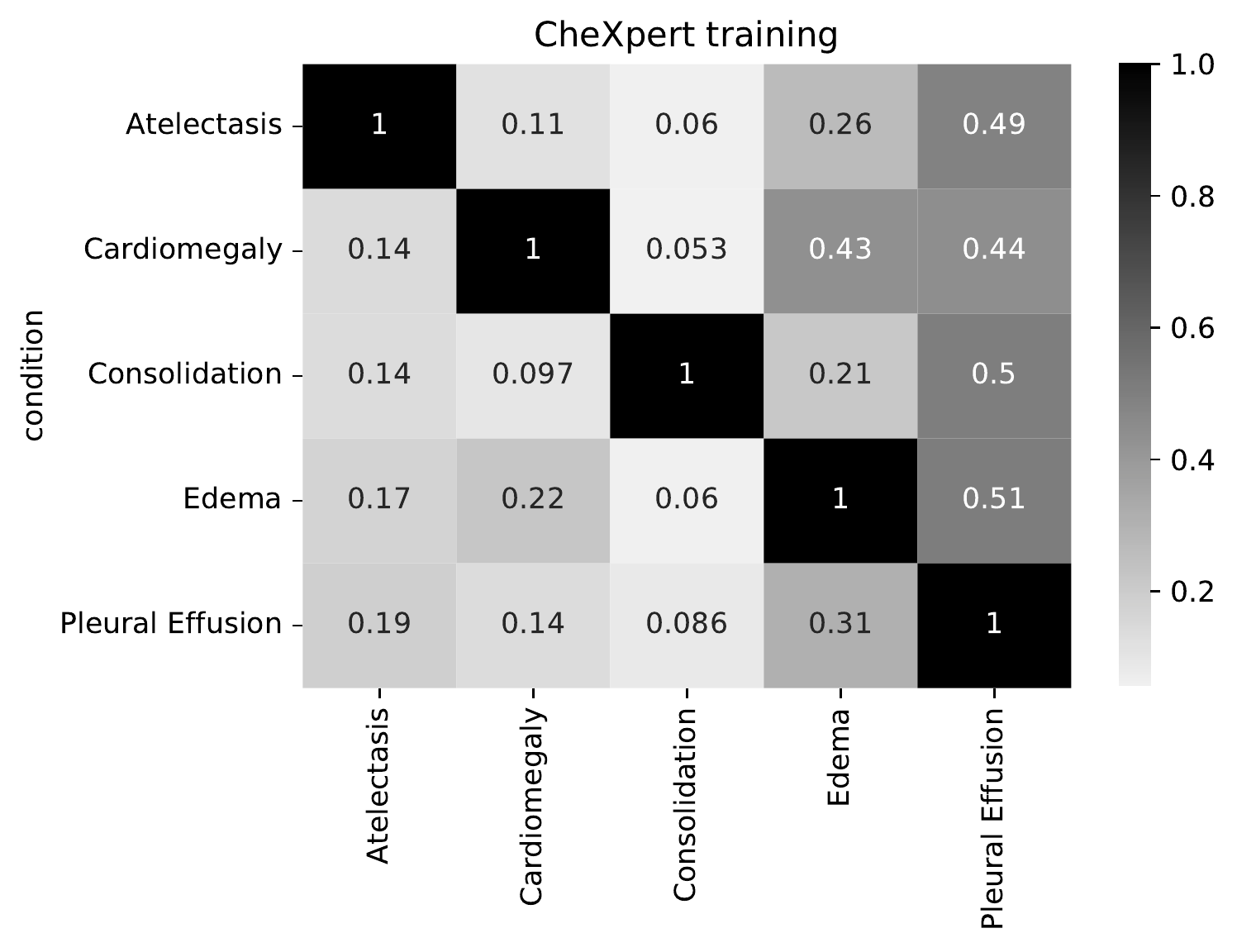}}
    \subfigure[Out-of-distribution condition co-occurrence]{ \includegraphics[width=0.49\linewidth]{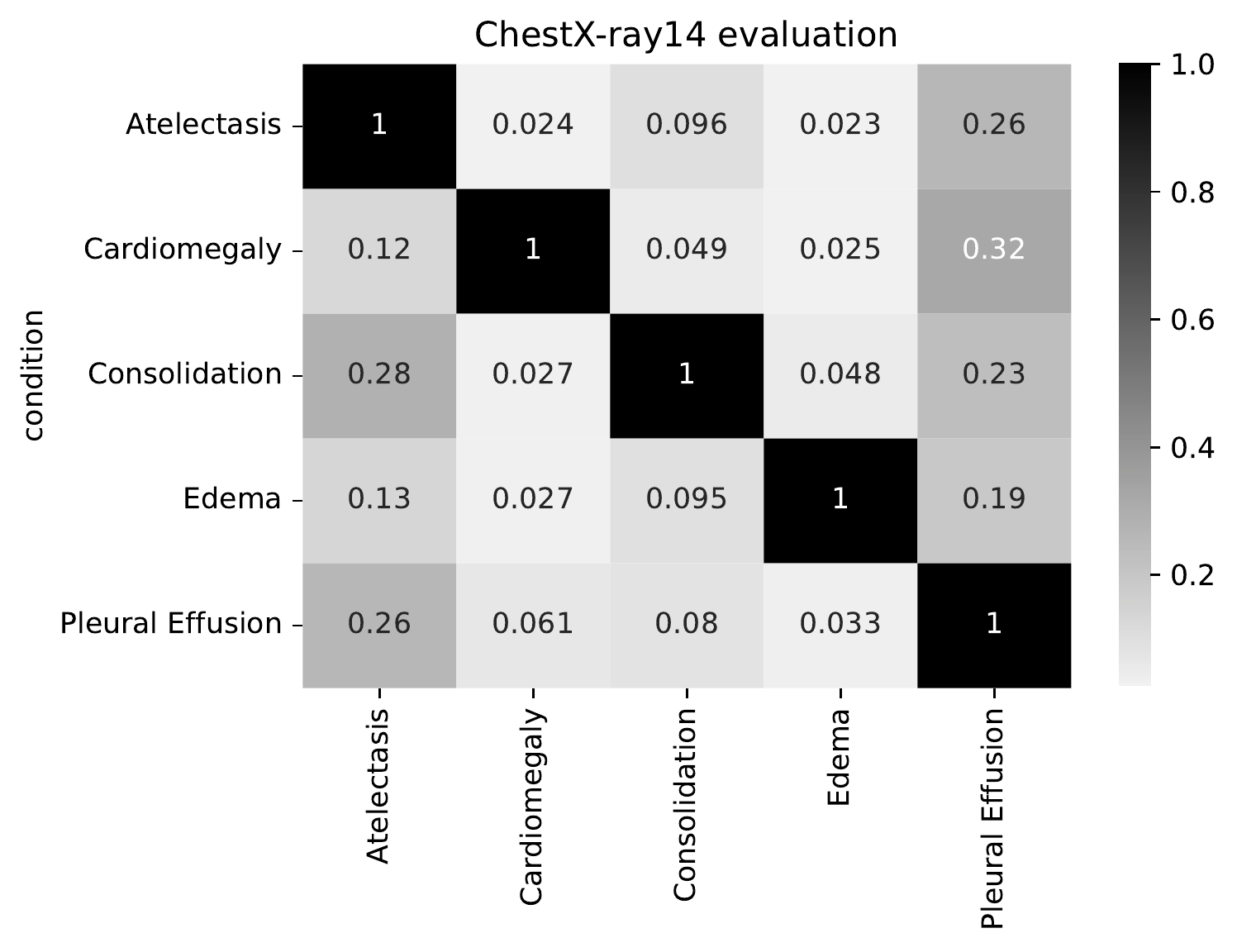}}
    \caption{Heatmaps of normalized co-occurrence of conditions in the (a) CheXpert training and (b) the ChestX-ray14 evaluation datasets. For each condition on the row $r$ of the heatmap, the corresponding column $c$ indicates the ratio of all samples with condition $r$ that also have condition $c$. Note that more than two conditions can be present at once. We observe that in the training set it is much more common that more than one condition is present simultaneously.}
    \label{fig:cxr_condition_coexist}
\end{figure}

The second setting that we consider is radiology. We focus our analysis on two large public radiology datasets, CheXpert~\citep{Irvin2019chexpert} and ChestX-ray14 (US National Institutes of Health (NIH))~\citep{Wang2017chestx}. These datasets have been widely studied by the community~\citep{rajpurkar2017chexnet,Larrazabal20,Seyyed21} for model development and fairness analyses. For these datasets, demographic attributes like sex and age are publicly available, and classification is performed at a higher resolution, i.e. $224 \times 224$ like in~\cite{Azizi2022robust}. After training the generative model and classifier on 201,055 examples of chest X-rays from the CheXpert dataset, we evaluate on a held-out CheXpert test set (containing 13,332 images), which we consider in-distribution, and the test set of ChestX-ray14 (containing 17,723 images), which we consider out-of-distribution (OOD) due to demographic and acquisition shifts. We focus on five conditions for which labels exist in common between the two datasets\footnote{Note that the labelling procedures for the two datasets were defined and enacted separately, which likely increases the complexity of the task.}, i.e., \textit{atelectasis}, \textit{consolidation}, \textit{cardiomegaly}, \textit{pleural effusion} and \textit{pulmonary edema}, while each of these datasets contains more conditions (not necessarily overlapping), as well as examples with no findings, corresponding to healthy controls. In this setting the model backbone is shared across all conditions, while a separate (binary classification) head is trained for each condition, given that multiple conditions can be present at once.~\autoref{fig:cxr_condition_coexist} illustrates how often different conditions co-occur in the training and evaluation samples. It is apparent that capturing the characteristics of a single condition can be challenging given that in most cases they coexist with other conditions. One characteristic example is pleural effusion, which is included in the diagnosis of atelectasis, consolidation and edema in ~50\% of the cases. However, the scenario is slightly different for the OOD ChestX-ray14 dataset, where for most pairs of conditions the corresponding ratio is much lower. It is worth noting that the original CheXpert training set contains positive, negative, uncertain and unmentioned labels. The uncertain samples are not considered when learning the classification model, but they are used for training the diffusion model. The unmentioned label is considered a negative (i.e. the condition is not present) which yields a highly imbalanced dataset. Therefore, we report area under the receiver operating characteristic (ROC-AUC) curve in line with the CheXpert leaderboard, as raw accuracy is not very informative for such imbalanced settings.

\begin{figure}[t]
\includegraphics[width=\textwidth]{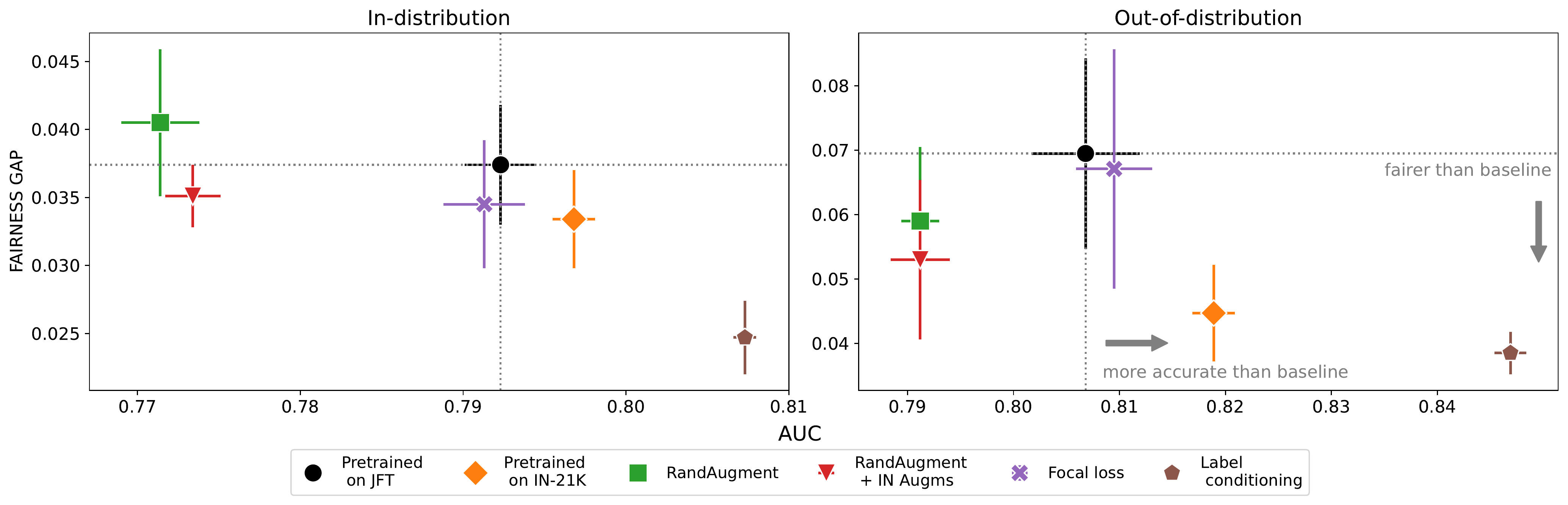}
\caption{Comparison of average AUC vs. fairness (AUC) gap across different baselines for radiology. We report results in- (left column) and out-of-distribution (right column) on CheXpert and ChestX-ray14 datasets, respectively. We mark the baseline \textit{Pretrained on JFT} with black. \textit{Label conditioning} corresponds to the model that uses synthetic images from a diffusion model conditioned on only the diagnostic labels. We further compare to other strong contenders, i.e., a BiT-ResNet model pretrained on ImageNet-21K (\textit{Pretrained on IN-21K}), a model pretrained on JFT using RandAugment heuristic augmentations (\textit{RandAugment}), a model trained with RandAugment on top of standard ImageNet augmentations (\textit{RandAugment + IN Augms}) and a model trained with focal loss (\textit{Focal loss}). To ensure a fair comparison all methods are trained / finetuned for the same number of steps and with the same batch size. It is worth noting that for the fairness gap smaller values are preferable.}
\label{fig:eval_auc_fairness_gap}
\end{figure}

We observe that synthetic images improve the average AUC for the five conditions of interest in-distribution, but even more so out-of-distribution. Improvements are particularly striking for cardiomegaly, where the model trained purely with synthetic images improves AUC by $21.1\%$ (see~\autoref{fig:radiology_per_condition}). Overall, we observe an improvement of $5.2\%$ on average AUC OOD and a $44.6\%$ improvement in sex fairness gap (see~\autoref{fig:eval_auc_fairness_gap}). We show some examples of generated augmentations by the diffusion model for a model conditioned on the diagnostic label in~\autoref{fig:generated_image_clinical}. Higher resolution images are generated in comparison to histopathology with the use of a cascaded diffusion model that upsamples images generated at $64 \times 64$ resolution to $224 \times 224$.

\subsubsection{Dermatology}

\begin{figure}[t]
\includegraphics[width=\textwidth]{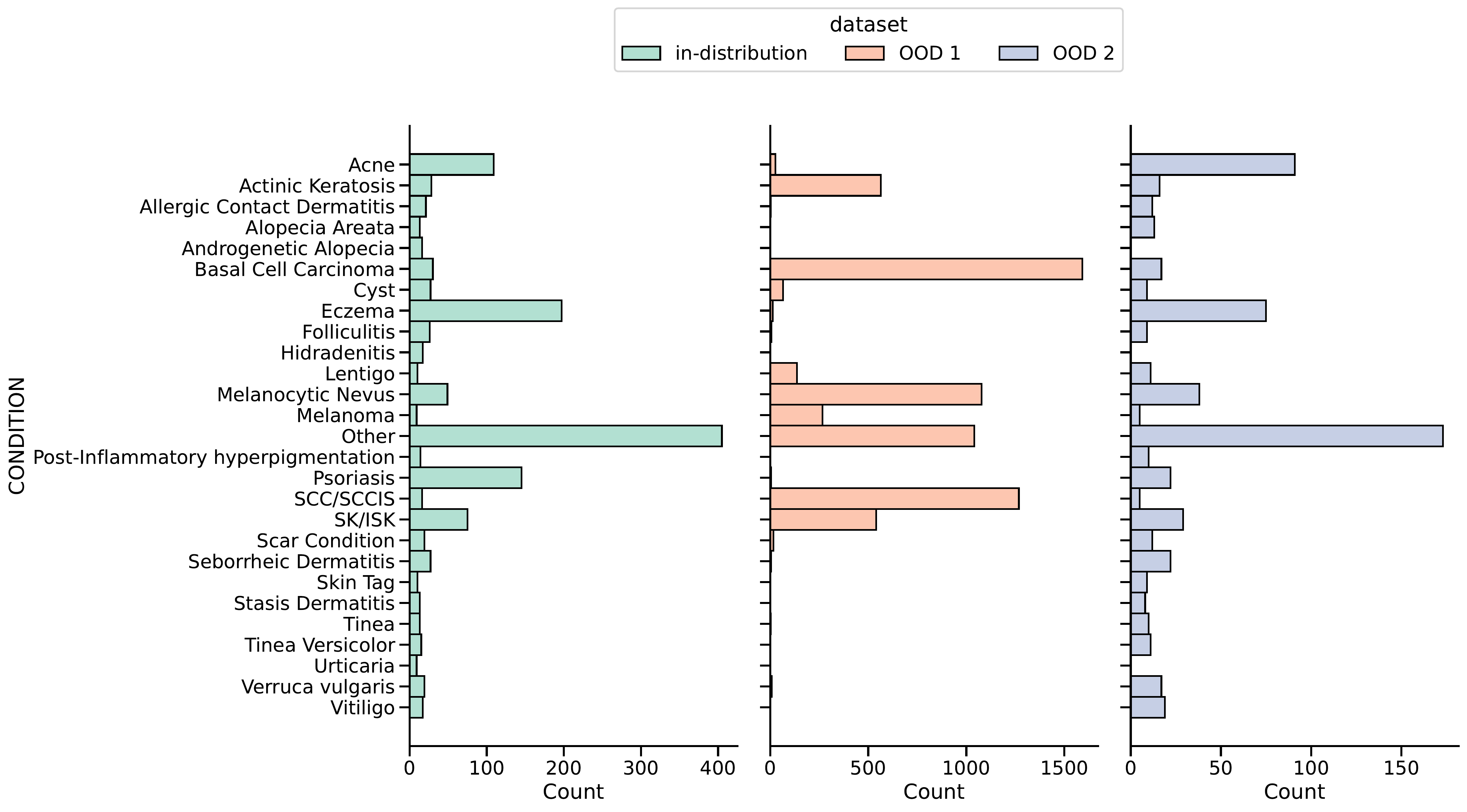}
\caption{Condition distributions for in-distribution and out-of-distribution dermatology datasets. In-domain and OOD 2 distributions are much more similar in comparison to OOD 1. In particular, 3 out of 4 of the most prevalent conditions (i.e. acne, eczema and other) in the in-distribution dataset are also the most prevalent in OOD 2. However, there are only few examples of high-risk conditions like basal cell carcinoma and SCC/SCCIS, which are the two most prevalent conditions in OOD 1. We can see that overall the right-most dataset has a similar label distribution to the training dataset and, hence, is `less' out-of-distribution than the other one.}
\label{fig:eval_condition_stats}
\end{figure}

\begin{figure}[t]
\includegraphics[width=\textwidth]{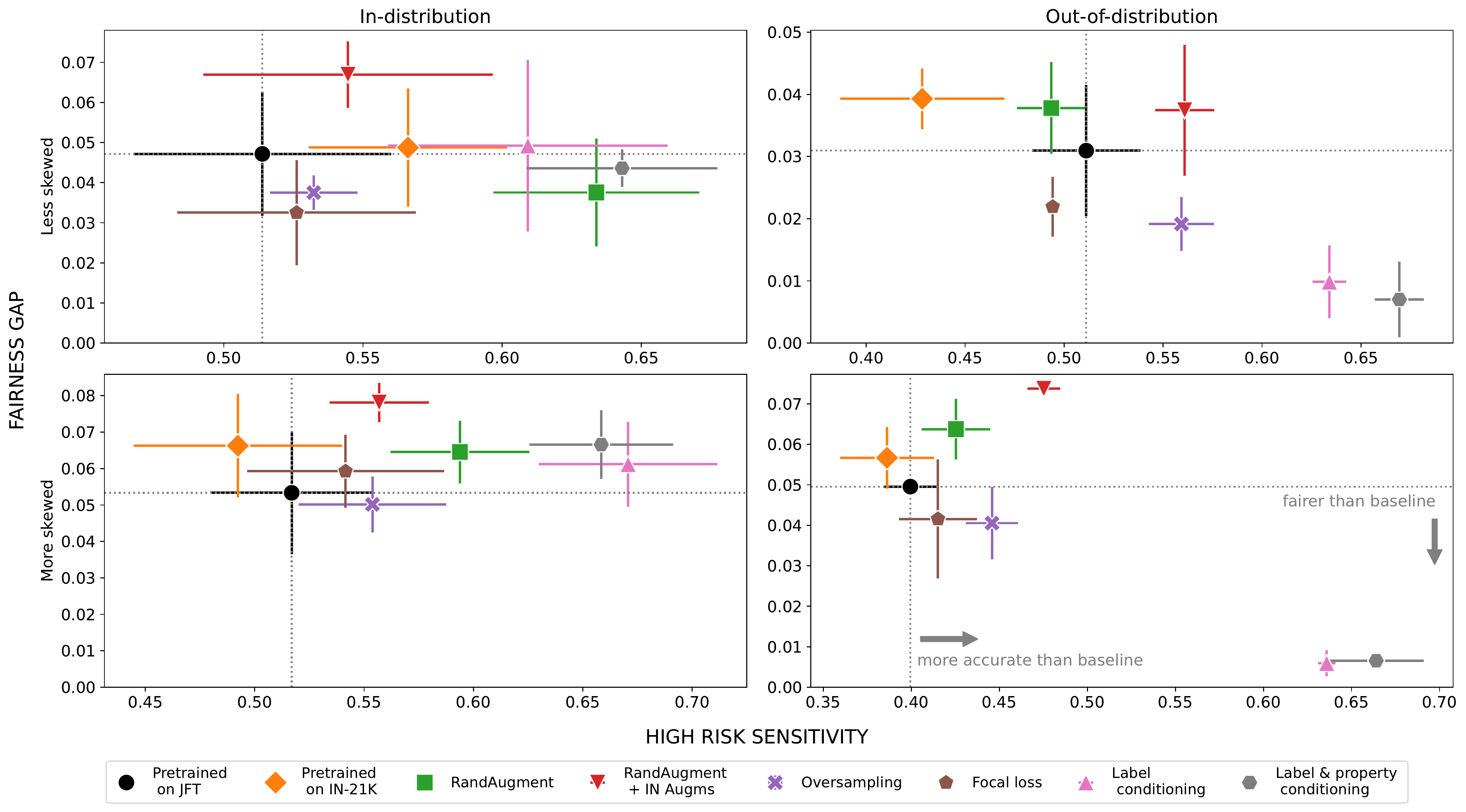}
\caption{Comparison of high-risk sensitivity (for \texttt{basal cell carcinoma}, \texttt{melanoma}, \texttt{squamous cell carcinoma (SCC/SCCIS)} and \texttt{urticaria}) vs. fairness gap w.r.t. sex in dermatology across different baselines. We report results in- (left column) and out-of-distribution for OOD 1 (right column), as well as for the less skewed (top row) and more skewed (bottom row) setting. We mark the baseline \textit{Pretrained on JFT} with black. \textit{Label conditioning} and \textit{Label \& property conditioning} correspond to the models that use synthetic images sampled from a diffusion model conditioned on only the label, and the label and sensitive attribute, respectively. We further compare to other strong contenders, i.e., a BiT-ResNet model pretrained on ImageNet-21K (\textit{Pretrained on IN-21K}), a model pretrained on JFT using RandAugment heuristic augmentations (\textit{RandAugment}), a model trained with RandAugment on top of standard ImageNet augmentations (\textit{RandAugment + IN Augms}), a model trained on a resampled version of the training dataset that is more balanced w.r.t. to the sensitive attribute (\textit{Oversampling}) and a model trained with focal loss (\textit{Focal loss}). To ensure a fair comparison all methods are trained / finetuned for the same number of steps and with the same batch size. It is worth noting that for the fairness gap, smaller values are preferable.}
\label{fig:eval_high_risk_fairness}
\end{figure}

For the dermatology setting, we consider  a dermatology dataset of images at $256 \times 256$ resolution grouped into 27 labelled conditions ranging from low risk (e.g.~acne, verruca vulgaris) to high risk (e.g.~melanoma). Out of these conditions, four are considered to be high-risk: basal cell carcinoma, melanoma, squamous cell carcinoma (SCC/SCCIS) and urticaria. The imaging samples are often accompanied with metadata that include attributes, like biological sex, age, and skin tone.
Skin tone is labelled according to the Fitzpatrick scale\footnote{https://dermnetnz.org/topics/skin-phototype}, which gives rise to 6 categories (plus unknown). The ground truth labels for the condition are the result of aggregation of clinical assessments by multiple experts, who provide a list of top-3 conditions along with a confidence score (between 1-5). A weighted aggregate of these labels gives rise to soft labels that we use for training the diffusion and downstream classifier models.
For the purposes of our experiments we consider three datasets: the in-distribution dataset featuring 16,530 cases from a tele-dermatology dataset acquired from a population in the US (Hawaii and California); OOD 1 dataset featuring 6,639 images of clinical type focusing mostly on high-risk conditions from an Australian population, and OOD 2 featuring 3,900 tele-dermatology images acquired in Colombia.
These datasets are characterized by complex shifts with respect to each other as the label distribution, demographic distribution and capture process may all vary across them. To demonstrate the severity of the prevalence shift across locations, we visualise the distribution of conditions in the evaluation datasets in~\autoref{fig:eval_condition_stats}.
For training the downstream classifier, we use labelled samples from only one of these datasets (in-distribution), while we include unlabelled images from the other two distributions when training the diffusion model. We evaluate on a held-out slice of the in-distribution dataset and two out-of-distribution sets to investigate how well models generalize. We present results for OOD 2 only in Supplementary material~\ref{sec:supp_different_splits_derm}, as it has similar label distribution to the in-distribution dataset and is less challenging.

We explore whether the proposed approach can be used to {\em not only} improve out-of-distribution accuracy {\em but also} fairness over the different label predictions and attributes for the in-distribution distribution. Given that images considered for dermatology are high resolution, we train a cascaded diffusion model that upsamples images generated at $64 \times 64$ resolution to $256 \times 256$.
While the datasets are already imbalanced with respect to different labels and sensitive attributes, we also investigate how the performance varies as a dataset becomes more or less skewed along a single one of these axes. This allows us to better understand to what extent conditioning generative models on the axis of interest can help alleviate biases with regard to the corresponding attribute. For example, if our original dataset is skewed towards younger age groups, conditioning the generative model on age and (over)sampling from older ages can potentially help close the performance gap between younger and older populations\footnote{To study this aspect, we cannot rebalance our datasets as we have too few samples from the long tail of our distribution with regards to the label or sensitive attribute.}. We skew the training labelled dataset to make it progressively more biased (by removing instances from the least represented subgroups) and investigate how performance suffers as a result of the skewing.
For each sensitive attribute, we create new versions of the in-distribution dataset that are progressively more skewed to the high data regions. We show how the resulting training dataset are skewed with respect to each of the sensitive attributes in~\autoref{tab:sens_attr_splits_train}. 

In~\autoref{fig:eval_high_risk_fairness},  we illustrate for a single axis of interest how different methods compare with regards to sensitivity for the four high-risk conditions mentioned above and fairness. In the more skewed setting the training dataset contains a maximum of 100 samples from the underrepresented subgroup regardless of the underlying condition, while in the less skewed setting it contains maximum 1000 samples. We compare all methods in four different settings: in- and out-of-distribution as well as less and more skewed with respect to the sensitive attribute of interest, i.e. sex. We observe that in all settings, combining heuristic augmentations as in \textit{RandAugment + IN Augms} does improve the predictive performance across the board, but harms fairness of the model. Pretraining on a different dataset, on the other hand, has a negative impact on both performance and fairness (except for some performance improvement in the less skewed setting). Using \textit{RandAugment} alone is beneficial for high-risk sensitivity in-distribution, but not out-of-distribution, but it harms fairness in the OOD setting. \textit{Oversampling} slightly closes the fairness gap across the board while improving performance, as expected. The approaches that leverage synthetic data, \textit{Label conditioning} and \textit{Label \& property conditioning}, improve on high-risk sensitivity in-distribution without reducing fairness, while they yield a significant improvement in the OOD setting on both axes. In the more skewed setting, in particular, \textit{Label \& property conditioning} leads to $27.3\%$ better high-risk sensitivity compared to the baseline in-distribution and a striking $63.5\%$ OOD, while closing the fairness gap by $7.5\times$ OOD. It is worth noting that the underrepresented group in the training set and the ID evaluation set is over-represented in the OOD evaluation set. Our approach shows improvements in accuracy and fairness metrics with respect to different sensitive attributes, while being able to generalize these improvements out-of-distribution as shown in~\ref{sec:supp_different_splits_derm}.

\begin{figure}
    \centering
    \subfigure[Cyst, melanolytic nevus, seborrheic dermatitis]{\includegraphics[clip, trim=3cm 0cm 3cm 0cm,width=0.49\linewidth]{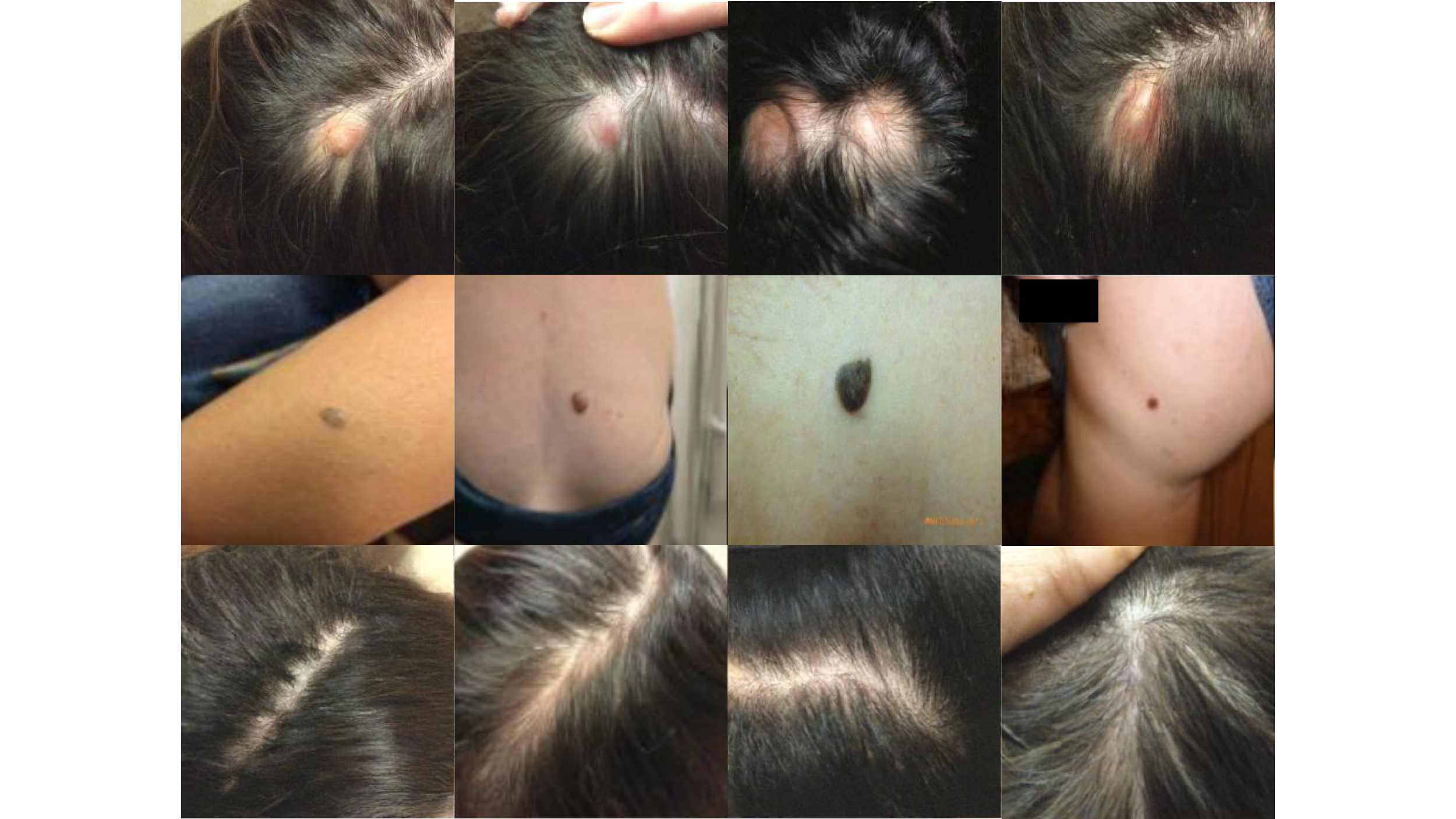}}
    \subfigure[Folliculitis, hidradenitis, alopecia areata]{ \includegraphics[clip, trim=3cm 0cm 3cm 0cm,width=0.49\linewidth]{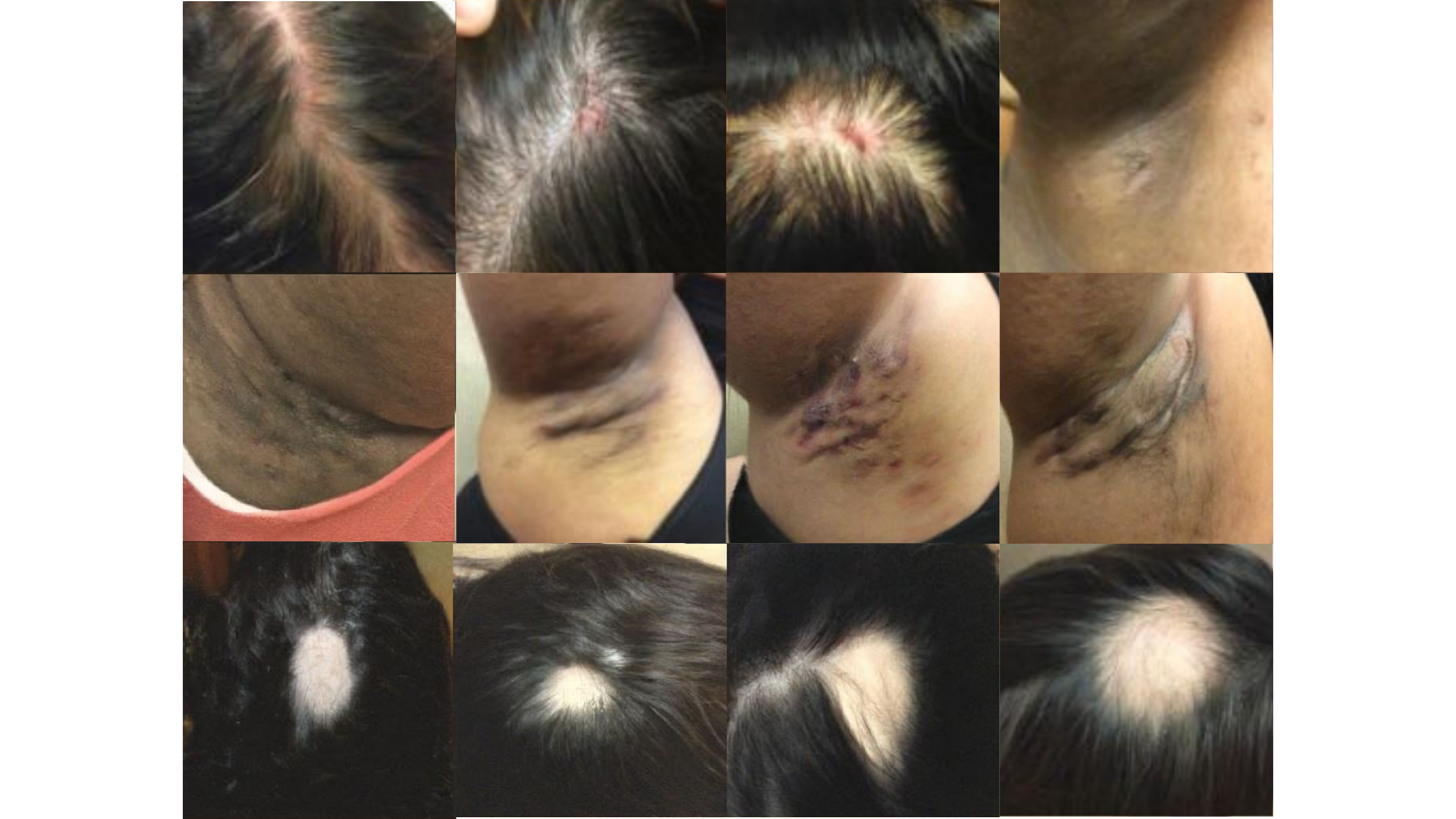}}
    \caption{Generated images in dermatology setting; each row of images corresponds to a different condition.}
    \label{fig:generated_image_natural}
\end{figure}

\subsection{In depth analysis for dermatology}
In this section our analysis focuses on the last modality of dermatology.

\subsubsection*{Generated images are diverse}
First, we show images generated at $256 \times 256$ resolution for this challenging, natural setting and a number of dermatological conditions in~\autoref{fig:generated_image_natural}. We highlight that our conditional generative model does capture the characteristics well for multiple, diverse conditions, even for cases that are more scarce in the dataset, such as \texttt{seborrheic dermatitis}, \texttt{alopecia areata} and \texttt{hidradenitis}. 

\subsubsection*{Generated images are realistic}
We further evaluate how realistic the generated images are as determined by expert dermatologists to validate that these images do contain properties of the disease used for conditioning.
We note that the synthetic images do not need to be perfect, as we are interested in downstream performance.
However, being able to generate realistic images validates that the generative model is capturing relevant features of the conditions.
To evaluate this, we ask dermatologists to rate a total of 488 synthetic images each, evenly sampled from the four most common classes (\texttt{eczema}, \texttt{psoriasis}, \texttt{acne}, \texttt{SK/ISK}) and four high risk classes (\texttt{melanoma}, \texttt{basal cell carcinoma}, \texttt{urticaria}, \texttt{SCC/SCCIS}). 
They are tasked to first determine if the image is of a sufficient quality to provide a diagnosis.
They are then asked to provide up to three diagnoses from over 20,000 common conditions with an associated confidence score (out of 5, where 5 is most confident).
These 20,000 conditions are mapped to the 27 classes we use in this paper (where one class, \texttt{other}, encompasses all conditions not represented in the other 26 classes).
We report mean and standard deviation for all metrics across the three raters.
$50.0 \pm 12.6\%$ of those images were found to be of a sufficient quality for diagnosis, while dermatologists had an average confidence of $4.13 \pm 0.43$ out of $5$ for their top diagnosis.
They had a top-1 accuracy of $56.0 \pm 11.9$\% on the generated images and a top-3 accuracy of $67.7 \pm 12.5$\%. We compare these numbers to a set of real images of the same eight conditions considered above (for the images considered, the majority of raters consider diagnosis of this disease as most prevalent in the image). Amongst 101 board certified dermatologists rating 789 real images in total\footnote{For this analysis, if an image has been rated by $N$ dermatologists, we consider a single rater's accuracy with respect to the aggregated diagnosis of the remaining $N-1$ raters.}, we found that their top-1 accuracy was $54.0 \pm 21.1$\% and top-3 accuracy $67.1 \pm 22.7$\%; slightly higher performance in terms of top-1 (63\%) and top-3 accuracy (75\%) was shown in \citep{Liu20} across a more diverse set of dermatological conditions.
This demonstrates that, when diagnosable as per experts' evaluation, synthetic images are indeed representative of the condition they are expected to capture; similarly so to the real images. Even though not all generated images are diagnosable, this can be the case for real samples as well, given that images used to train the generative model do not necessarily include the body part or view that best reflects the condition.



\subsubsection*{Generated images are canonical}
\begin{figure}[t]
\includegraphics[clip, trim=0.0cm 4cm 0.0cm 1cm, width=\textwidth]{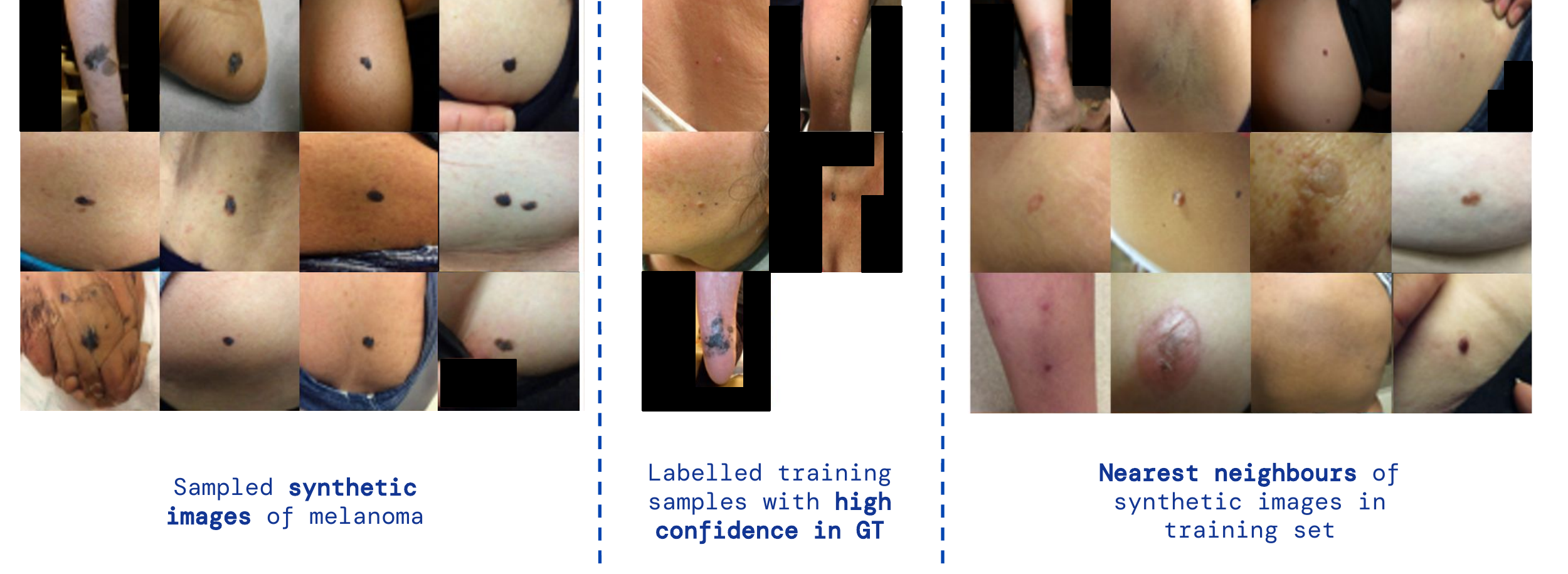}
\caption{(\textit{Left}) The diffusion model can produce an infinite amount of synthetic images for a particular condition that is inherently more scarce (we have fewer than 50 samples of melanoma in our training dataset). (\textit{Middle}) The images that experts have identified with melanoma with high confidence (a combination of individual's confidence in diagnosis and expert consensus). (\textit{Right}) The nearest neighbours from the training samples identified for each of the synthetic images based on $\ell^2$-norm in pixel space.}
\label{fig:canonical_image_diversity}
\end{figure}

We hypothesize that the reason why models become more robust to prevalence shifts is due to synthetic images being more canonical examples of the conditions. To understand how canonical ground truth images for a particular condition are, we investigate cases with high degree of concordance in raters' assessments and compare those to synthetic images for the same condition. More specifically, we threshold the aggregated ground truth values to filter the images within the training data that experts were most confident about presenting a condition. The aggregation function operates as follows: assume we have a set of 4 conditions $\{A, B, C, D\}$; if rater $R_1$ provides the following sequence of (\texttt{condition}, \texttt{confidence}) diagnosis tuples: $\{(A, 4), (B, 3)\}$ and rater $R_2$ provides $\{(A, 3), (D, 4)\}$, then we obtain the following soft labels $\{0.5, 0.167, 0, 0.333\}$ (after weighting each condition with the inverse of its rank for each labeller, summing across labellers and normalising their scores to 1).
If we look for instances for which there is consensus amongst raters and high-confidence that a condition is present we can threshold the corresponding soft label for that condition with a strict threshold, e.g. $t=0.9$. In our example, this doesn't hold for any of the 4 conditions, but if we lowered the threshold to 0.5, then it would hold for condition $A$.
In~\autoref{fig:canonical_image_diversity} we show an example for melanoma. For this particular diagnostic class we are able to generate multiple synthetic instances of the condition, while we recovered only 5 images (out of $>15,000$) that clinicians rated with high confidence, i.e. $t_{\textrm{melanoma}}=0.9$. The nearest neighbours from the training dataset identified based on $\ell^2$-norm are also shown in~\autoref{fig:canonical_image_diversity}.

\subsubsection*{Generated images align feature distributions better}
Previous work on out-of-distribution generalization \citep{Ben2010theory, Muandet2013domain, Albuquerque2019generalizing} has pointed out that several factors can affect the performance of a model on samples from domains beyond the training data. In this analysis, we investigate the models trained with our proposed learned augmentations in terms of changes in distribution alignment between all pairs of distributions measured via the Maximum Mean Discrepancy (MMD)~\citep{Gretton2012kernel}, as previous work has empirically shown that approaches based on learning features that decrease MMD estimates yield improved out-of-distribution generalization~\citep{Li2018domain}. We compute domain mismatches considering the space where decisions are performed, i.e., the output of the penultimate layer of each model. We thus project each data point from the input space to a representation. We find that learned augmentations yield on average 18.6\% lower MMD in comparison to heuristic augmentations (for more details refer to~\ref{sec:supp_different_splits_derm}) which leads to the following conclusions: (i) Data augmentation has a significant effect on distribution alignment. Improvement on OOD performance suggests this is happening via learning better predictive features rather than capturing spurious correlations. (ii) Generated data helps the model to better match different domains by attenuating the overall discrepancy between domains. (iii) Given the minor decline in performance when adding generated data in the less skewed setting as shown in~\autoref{fig:eval_high_risk_fairness}, these findings suggest that learning such features might conflict with learning spurious correlations that were helpful for in-distribution performance.

\subsubsection*{Synthetic images reduce spurious correlations}

To further compare the effect of different augmentation schemes on the features learned by the downstream classifier, we investigate the representation space occupied by all considered datasets, including samples obtained from the generative model. In practice, we project $N$ randomly sampled instances from each dataset to the feature space learned by each model and apply the Principal Component Analysis algorithm~\citep{abdi2010principal}. We then extract the number of principal components required to represent different fractions of the variance across all instances projected to the feature spaces induced by models obtained with heuristic and learned augmentations.
We observe that for a fixed dataset, features from models trained with synthetic data require 5.4\% fewer principal components to retain 90\% of the variance in latent feature space (results for different fractions are provided in~\autoref{fig:pca_analysis}). This indicates that using synthetic data induces more compressed representations in comparison to augmenting the training data in a heuristic manner. Considering this finding in the context of the results in~\autoref{tab:mmd_domains}, we posit the observed effect is due to domain-specific information being attenuated in the feature space learned by models trained with synthetic data. This suggests that our proposed approach is capable of reducing model's reliance upon correlations between inputs and labels that do not generalize out-of-distribution.

\section{Discussion}

In this work, we propose to use conditional generative models for improving robustness and fairness of machine learning systems applied to medical imaging. More specifically, we show that diffusion models can produce useful synthetic images in three different medical settings of varying difficulty, complexity and resolution: histopathology, radiology and dermatolgy. Our experimental evaluation provides extensive evidence that synthetic images can indeed improve statistical fairness, balanced accuracy and high risk sensitivity in a multi-class setting, while improving robustness of models both in- and out-of-distribution. In fact, we observe that generated data can be more beneficial out-of-distribution than in-distribution even in the absence of data from the target domain during training of the generative model (in the case of radiology). Generative models prove to be label efficient in both histopathology and dermatology settings, where we demonstrate that only a few labelled examples are sufficient for the diffusion models to capture the underlying data distribution well. This is particularly impactful in the medical setting, where data for particular conditions or demographic subgroups can be scarce or, even when available, acquiring expert labels can be expensive and time consuming. For the reader that is familiar with regularization techniques, we view diffusion models as another form of regularization, which can be combined with any other architecture or learning method improvements. 

Even though we do not make any assumptions when training the diffusion model, we find interesting dynamics when combining real and synthetic data. In certain settings, i.e., histopathology and radiology, we observe that we can rely purely on generated data and still outperform baselines trained with real labelled data (see~\ref{subsec:hist_label_eff}). In other settings, like dermatology, we observe that real data is more essential for training of the downstream discriminative model. We take a step further and analyze the impact of generated data and the mechanisms underlying the improvements in robustness and fairness that we report. Synthetic samples seem to better align distributions of different domains, while at the same time allowing models to learn more complex decision boundaries that reduce their reliance on spurious correlations. Finally, we highlight some practical benefits (highlighted in {\color{OliveGreen}green}) and discuss a number of potential risks (highlighted in {\color{purple}red}) and limitations (in {\color{BurntOrange}orange}) from relying on generated data.

\paragraph{\color{OliveGreen}Reusability of synthetic data.}
Beyond the analysis and utility of synthetic data for the particular tasks that we consider in this work, there are many other potential applications for which they can be useful. The same synthetic data can be used for data augmentation across different models and, potentially, tasks. For example, hand-crafted augmentations are often employed to introduce invariances and learn better representations in a self-supervised manner for a variety of downstream tasks.

\paragraph{\color{OliveGreen}Scalable approach.}
As we demonstrate in~\autoref{sec:toy_example}, if we have a perfect generative model then we can perform perfectly under the fair distribution.
Moreover, the better the generative model, the more our results should improve.
As a result, as generative modelling improves or as more data becomes available, results should improve accordingly.

\paragraph{\color{OliveGreen}Utility for leveraging private data sources.}
Combining this technique with privacy-preserving technologies holds a lot of promise in the medical field. One of the main reasons why transformative AI technologies have not yet demonstrated equivalent impact in the healthcare domain is due to regulations and limited data access. There is preliminary evidence that federated learning can be used to learn classification models from multiple institutions~\citep{Kaissis2021end} and if it were possible to generate private synthetic data, this synthetic data could be used for data augmentation along with a smaller, public dataset to improve performance.
This could have practical benefits when data sharing to protect personally identifiable information (PII) while achieving high quality performance. Such an approach would of course be associated with its own risks, some of which are discussed by~\cite{Cheng21}.

\paragraph{\color{purple}Overconfidence in the model.}
Even though we show that diffusion models can be particularly label efficient, this should not encourage practitioners to abandon their data and label acquisition efforts; nor does it imply that generated data can replace real data under any circumstances. What this research demonstrates is that, when labelled data and resources are limited, there are ways to make more of the available labelled and unlabelled data. There is also the potential that using generative models may lead to overconfidence in an AI system, because images look realistic to a non-expert.
Additional data collection will always be important, along with comprehensive analysis of the underlying data and its caveats.
Synthetic data from a generative model should {\em only} be used as a complement to additional data collection and accompanied by rigorous evaluation on real data, ideally outside the main source domain to understand generalization capabilities of the models. In other words, synthetic data is one solution to increase diversity, but not a substitution of efforts to increase data representation for underrepresented conditions and populations.

\paragraph{\color{purple}Bias in the training data.}
If the generative model is of poor quality or biased, then we may end up exacerbating problems of bias in the downstream model.
The generative model may be unable to generate images of a certain label and sensitive attribute. In other settings, the model may always generate a specific part of the distribution for a certain label and sensitive attribute instead of capturing the true image distribution.
The generative model may also create incorrect images of a given label and sensitive attribute, leading the classification model to make mistakes confidently in those regions. Therefore, it is particularly important that the evaluation data is unbiased.

\paragraph{\color{purple}Bias in the evaluation.}
The insights that we obtain by analyzing the model are only as good as our evaluation setup. If the evaluation datasets are not diverse enough, do not capture high-risk conditions well or are not representative of the population, then any conclusions we draw from these results will be limited. Therefore, care needs to be taken in order to report and understand what each of the evaluation setups is capturing. For example, as~\cite{varoquaux2022machine} highlight, clinician-level performance is often overstated without validating models out-of-distribution.

\paragraph{\color{BurntOrange}Categorical and unobserved attributes.} Sensitive attributes are not always observed or explicitly tracked and reported~\citep{Tomasev2021fairness}, often to protect people's privacy. At the same time, the way labels are assigned may have its own limitations. For example, using binary gender and sex attributes (or using the two interchangeably) does not represent people that identify as non-binary. Similarly, researchers have criticized the Fitzpatrick Skin Type because it is less accurate on shades of darker skin tones, which could cause models to misidentify or misrepresent people with darker skin. Similarly, there are other unobserved characteristics that can influence disease and are not accounted for in a visual image of skin, for example, like social determintants of health. One instance of this is how dermatitis on a person who lives in a communal setting could have a different differential diagnosis than dermatitis in a high on a high income individual. These are important considerations when relying on such attributes to condition learned augmentations or to perform fairness analyses.

\paragraph{\color{BurntOrange}Transparency when handling synthetic data.} Synthetic images should be handled with caution as they may perpetuate biases in the original training data. It is important to tag and identify when a synthetic image has been added to a database, especially when considering to reuse the dataset in a different setting or by different practitioners.

We see potential here for future work that improves fairness and out-of-distribution generalization by leveraging powerful generative models but without explicitly relying on pre-defined categorical labels. When we consider synthetic images as an option for addressing performance gaps across subgroups, the following challenges still need to be addressed: reducing memorization for rare attributes and conditions, providing privacy guarantees and accounting for unobserved characteristics.

\section{Acknowledgements}
We would like to thank Mikolaj Binkowski for his input on the data preprocessing for the diffusion upsampler and William Isaac for his input on the ethical risks of this work. We would also like to thank Florian Stimberg, Jan Freyberg, Terry Spitz, Vivek Natarajan, Yun Liu, and David Warde-Farley for providing feedback at different stages of the project, as well as Sophie Elster, Zahra Ahmed, Nina Anderson and Patricia Strachan for their organisational support. Last, but not least, we thank Jessica Schrouff, Yuan Liu, Heather Cole-Lewis and Naama Hammel for the technical feedback they provided on the manuscript.

\section{Author Contributions}

O.W., S.G. and P.K. initiated the project. O.W.,  I.K. and S.G. contributed to the design of the method and experiments. O.W., S.G. and T.C. contributed to the formulation of the method. A.G.R. provided pointers to the datasets. I.K, O.W., S.G. and A.G.R. contributed to software engineering. I.A. performed in-depth analysis on distribution matching and spurious correlations. R.T. and O.W. performed analysis of different sampling schemes. I.K. trained upsamplers and produced high-resolution images. O.W. performed nearest-neighbour analysis for dermatology. A.K. helped formulate the problem in the clinical setting. I.K. and O.W. performed experiments on different modalities. I.K. and O.W. analysed results from expert evaluations in dermatology. S.A.R. performed analysis on mis-classification rates for high-risk individual samples. I.K., O.W., I.A., R.T., S.A.R. and S.G. contributed to the evaluation of the work and performed analysis. I.K., O.W., I.A., R.T., S.A.R., A.G.R., A.K. and S.G. contributed to the interpretation of the results. D.B. and P.K. advised on the work. I.K., O.W., I.A. and R.T. wrote the paper. S.G., P.K., D.B. and A.K. revised the manuscript.


\bibliographystyle{abbrvnat}
\setlength{\bibsep}{5pt} 
\setlength{\bibhang}{0pt}
\bibliography{references}








\newpage

\section*{Supplementary Materials}
\appendix
\renewcommand{\contentsname}{Contents}

\renewcommand{\baselinestretch}{0.5}\normalsize 
\startcontents
\printcontents{l}{1}{\small \section*{\contentsname}}
\renewcommand{\baselinestretch}{1.0}\normalsize 

\counterwithin{figure}{section}
\counterwithin{table}{section}
\renewcommand\thefigure{\thesection\arabic{figure}}
\renewcommand\thetable{\thesection\arabic{table}}

\newpage
\section{Datasets}
\label{sec:datasets}
In this section, we describe the datasets that we used for training downstream classifiers and diffusion models across the different modalities and medical contexts. Three different datasets were used, all of which are de-identified.

\subsection{Histopathology}
We use data from the CAMELYON17 challenge~\citep{Bandi2018detection} that includes labelled and unlabelled data from three different hospitals for training, as well as one in-distribution and one out-of-distribution validation hospitals. Data from different hospitals differs due to the staining procedure used. The task is to estimate the presence of breast cancer metastases in the images which are patches of whole-slide images of histological lymph node sections. The number of samples per hospital are given in \autoref{tab:camelyondatasetstatistics}; all subsets are approximately evenly split into those containing tumours and those that do not. We use the training data (302,436 examples) and the unlabelled data (1.8M examples) in order to train the diffusion model. 

\begin{table}[h]
    \centering
    \footnotesize
    \begin{tabular}{lcccccr}
    \toprule 
    & \textbf{Hospital 0} & \textbf{Hospital 1} & \textbf{Hospital 2} & \textbf{Hospital 3} & \textbf{Hospital 4} & \textbf{Total} \\ \toprule
    & \multicolumn{5}{c}{Labelled Data}  \\
    \textbf{Train} & 53,425 & -- & -- & 116,959 & 132,052 & 302,436 \\
    \textbf{ID (Validation)} & 6,011 &  -- & --  & 12,879 & 14,670 & 33,560 \\
    \textbf{OOD (Validation)} & -- & -- & 34,904 & -- & -- & 34,904 \\
    \textbf{OOD (Test)} & -- & 85,054 & -- & -- & -- & 85,054 \\ \midrule
    & \multicolumn{5}{c}{Unlabelled Data} \\
    \textbf{Train} & 599,187 & -- & -- &  600,030 & 600,030 & 1,799,247 \\ \bottomrule
    \end{tabular}
    \caption{Dataset statistics for CAMELYON17. For all hospitals, the labelled data is approximately evenly split between tumorous and non-tumorous images.}
    \label{tab:camelyondatasetstatistics}
\end{table}

\subsection{Chest Radiology}
\begin{figure}[h]
\centering
\subfigure[Age histogram]{\includegraphics[clip, trim=0cm -0.5cm 0cm 0cm,width=0.28\textwidth]{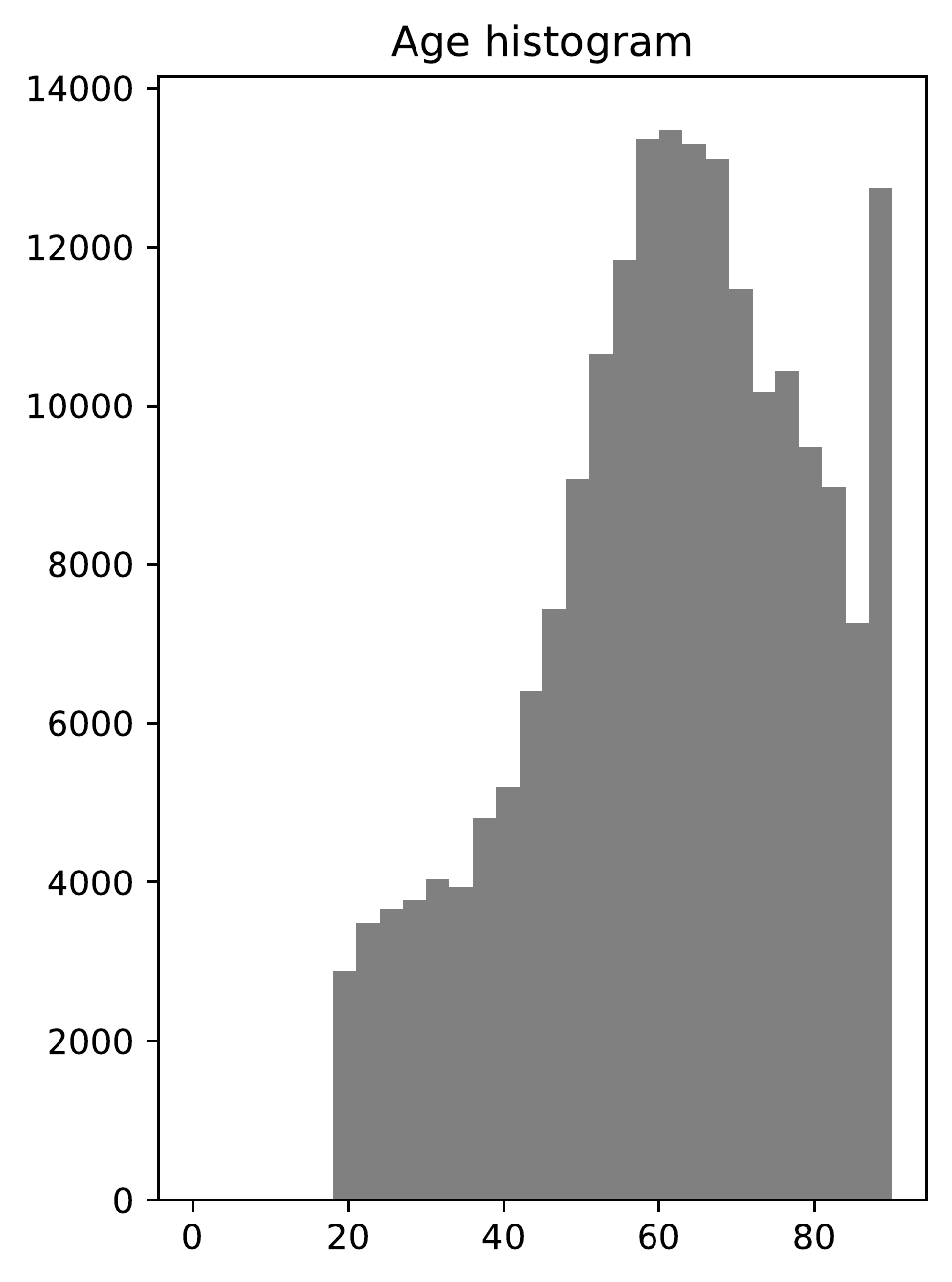}}
\subfigure[Label distribution]{\includegraphics[width=0.62\textwidth]{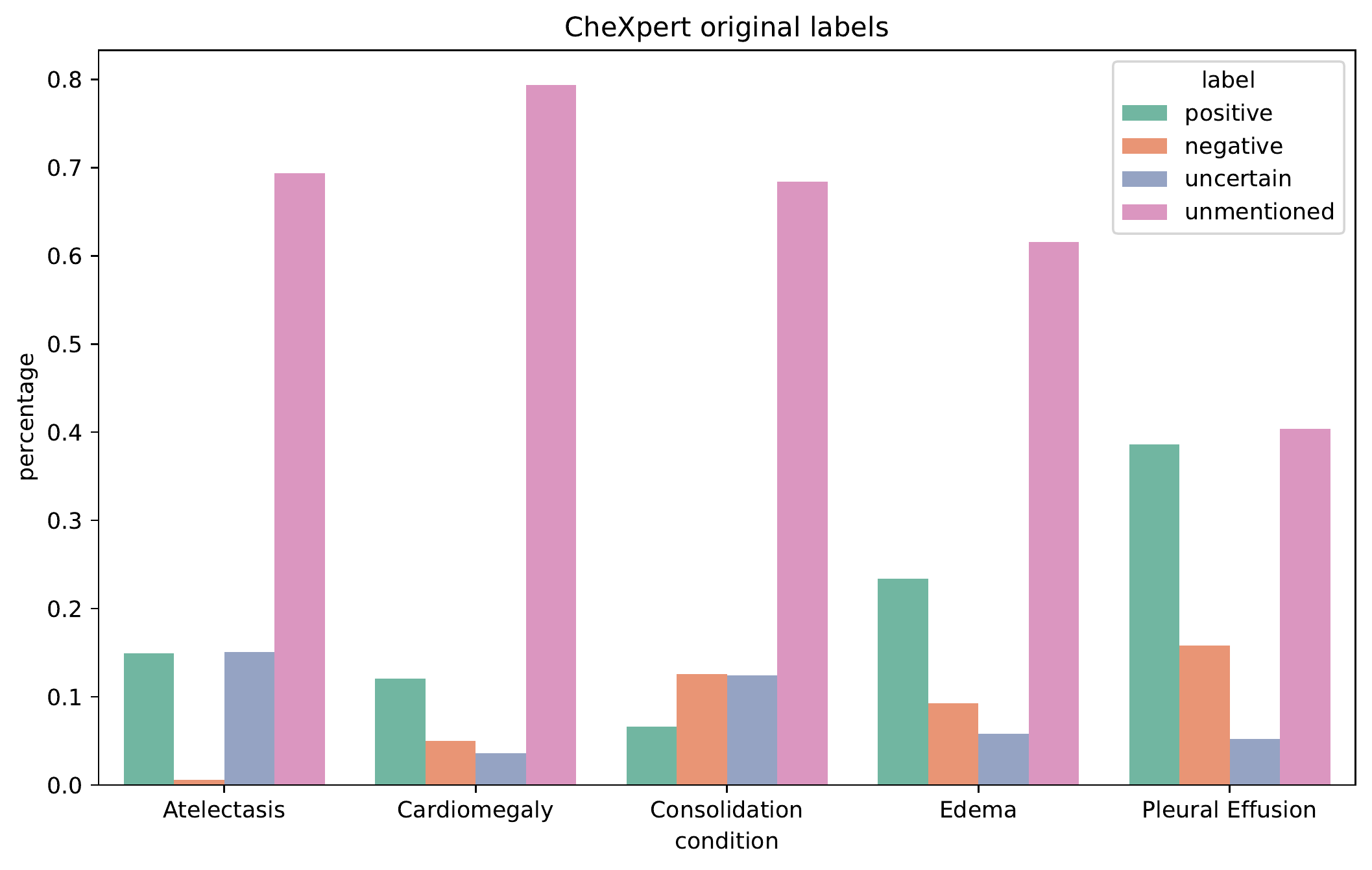}}
\caption{Age histogram and normalized label distributions for the five conditions we consider in the CheXpert training dataset.}
\label{fig:chexpert_training_stats}
\end{figure}

We train the cascaded diffusion and downstream discriminative model on a total of 201,055 samples from the CheXpert database~\citep{Irvin2019chexpert}, with 119,352 individuals annotated as male and 81,703 as female (the dataset only contains binary gender labels). We show the age and original label distribution in~\autoref{fig:chexpert_training_stats}. The uncertain samples are only used for training of the diffusion model. The unmentioned label is mapped to negative, which yields a highly imbalanced dataset. The evaluation NIH dataset~\citep{Wang2017chestx} denoted as out-of-distribution consists of 17,723 individuals, out of which 10,228 are male and 7,495 are female.

\subsection{Dermatology}
\label{subsec:dataset_dermatology}
The original dermatology dataset is characterized by complex shifts. In order to disentangle the effect of each of those shifts, we artificially skew the source dataset along three sensitive attribute axes: sex, skin tone and age. Skewing the dataset allows us to understand which methods perform better as the distribution shifts become more severe. We report how skewing the training dataset impacts the number of samples from the low data regions of the distribution in~\autoref{tab:sens_attr_splits_train}. We also report similar demographic statistics for the 3 evaluation datasets in~\autoref{tab:sens_attr_splits_eval}. The cascaded diffusion model is always trained on the union of the labelled training data and the total of unlabelled data across the three available domains. The discriminative model is always evaluated on the same three evaluation datasets (one in-distribution held-out dataset and two out-of-distribution datasets) for consistency.

\begin{table}[h]
\subtable[Sex splits]{
\begin{tabular}{lcc}
\hline
\textbf{Setting}        & \textbf{Female}       & \textbf{Male} \\ \hline
\textbf{Less skewed} & 8972  & 1157	\\
\textbf{More skewed} & 8972 & 115 \\
\textbf{Most skewed} & 8972 & 19 \\ 
\hline
\end{tabular}
\label{tab:sex_splits}
}
\hfill%
\subtable[Skin tone splits (Fitzpatrick scale)]{
\begin{tabular}{lccccccc}
\hline
\textbf{Setting} & \textbf{Pale white}       & \textbf{White} & \textbf{Beige} & \textbf{Brown}  & \textbf{Dark brown}  & \textbf{Black} & \textbf{Unknown} \\ \hline
\textbf{Less skewed} & 14  & 2433	 & 5668 & 975 & 103 & 4 & 1031 \\
\textbf{More skewed} & 14  & 2433	 & 5668 & 114 & 12 & 0 & 1031 \\
\textbf{Most skewed} & 14  & 2433	 & 5668 & 10 & 1 & 0 & 1031 \\ 
\hline
\end{tabular}
\label{tab:skintone_split}
}
\hfill%
\subtable[Age splits]{
\begin{tabular}{lccccccc}
\hline
\textbf{Setting}        & \textbf{(15, 25]} & \textbf{(25, 35]} & \textbf{(35, 45]} & \textbf{(45, 55]} & \textbf{(55, 65]} & \textbf{(65, 75]} & \textbf{(75, 90]} \\ \hline
\textbf{Less skewed} & 2662  & 2700 & 2163 & 2401 & 2098 & 250 &	76 \\
\textbf{More skewed} & 3054  & 3145 & 2541 & 2495 & 510 & 14 &	2 \\
\textbf{Most skewed} & 3054  & 3145 & 2541 & 2632 & 226 & 0 & 0 \\ 
\hline
\end{tabular}
\label{tab:age_split}
}
\caption{Number of training samples annotated with the corresponding sensitive attribute label across the different simulated splits.}
\label{tab:sens_attr_splits_train}
\end{table}

\begin{table}[h]
\subtable[Sex distribution]{
\begin{tabular}{lcc}
\hline
\textbf{Setting} & \textbf{Female} & \textbf{Male} \\ \hline
\textbf{in-distribution} & 804  & 545 \\
\textbf{OOD 1} & 3153 & 3486 \\
\textbf{OOD 2} & 396 & 246 \\ 
\hline
\end{tabular}
\label{tab:sex_splits_eval}
}
\hfill%
\subtable[Skin tone distribution (Fitzpatrick scale)]{
\begin{tabular}{lccccccc}
\hline
\textbf{Setting} & \textbf{Pale white}       & \textbf{White} & \textbf{Beige} & \textbf{Brown}  & \textbf{Dark brown}  & \textbf{Black} & \textbf{Unknown} \\ \hline
\textbf{in-distribution} & 20 & 193 & 528 & 439 & 52 & 20 & 97 \\
\textbf{OOD 1} & 7 & 99 & 207 & 19 & 0 & 0 & 6307 \\
\textbf{OOD 2} & 5 & 99 & 249 & 220 & 43 & 1 & 25 \\ 
\hline
\end{tabular}
\label{tab:skintone_split_eval}
}
\hfill%
\subtable[Age distribution]{
\begin{tabular}{lccccccc}
\hline
\textbf{Setting}        & \textbf{(15, 25]} & \textbf{(25, 35]} & \textbf{(35, 45]} & \textbf{(45, 55]} & \textbf{(55, 65]} & \textbf{(65, 75]} & \textbf{(75, 90]} \\ \hline
\textbf{in-distribution} & 213  & 212 & 207 & 217 & 213 & 180 & 107 \\
\textbf{OOD 1} & 108  & 295 & 552 & 1005 & 1637 & 1971 & 1065 \\
\textbf{OOD 2} & 98 & 98 & 64 & 98 & 44 & 52 & 32 \\ 
\hline
\end{tabular}
\label{tab:age_split_eval}
}
\caption{Number of evaluation samples annotated with the corresponding sensitive attribute label across the different domains.}
\label{tab:sens_attr_splits_eval}
\end{table}

\section{Related work}
\label{sec:related_work}

\subsection{Learning augmentations with generative models in Health}
In the clinical setting, generative adversarial networks (GANs) have been employed by various studies to improve performance in different tasks, e.g. disease diagnosis, in scenarios where few labelled samples are available. Such models have been used to augment medical images for liver lesion classification~\citep{Frid18}, classification of diabetic retinopathy from fundus images~\citep{Ju21} and breast mass diagnosis in mammography~\citep{Li19}. In dermoscopic imaging~\cite{Baur18} introduced a progressive generative model to produce realistic high-resolution synthetic images, while~\cite{Rashid19} focused on improving balanced multiclass accuracy and, in particular, sensitivity for high-risk underrepresented diagnostic labels like melanoma.~\cite{Han20} focused on a similar approach for chest X-rays by combining real and synthetic images generated with GANs to improve classifier accuracy for rare diseases.~\cite{Havaei21} use conditional image generation in scenarios where the conditioning vector is not always available to disentangle image content and image style information. They apply the method on dermatoscopic images (HAM10000 dataset) corresponding to seven types of skin lesions and lung CT scans from the Lung Image Database Consortium (LIDC-IDRI).

Apart from whole-image downstream tasks, GAN-based augmentation techniques have been used to improve performance on pixel-wise classification tasks, e.g. vessel contour segmentation on fundus images~\citep{Zhao18}, brain lesion segmentation~\citep{Uzunova20}. Given that pixel-wise downstream tasks are not within the scope of our study, we refer the reader to a more thorough review of GAN-based methods in medical image augmentation by~\cite{Chen22}.~\cite{Bissoto2021gan}, in turn, provide an overview of GAN-based augmentation techniques with a main focus on skin lesion augmentation and anonymization.

Despite the wide variety of health applications that have adopted GAN-based generative models to produce learned augmentations of images, these are often characterised by limited diversity and quality~\citep{Zhang22}. More recently, denoising diffusion probabilistic models (DDPM) trained on large scale data by~\cite{Ho20,Nichol21,Ho22} presented outstanding performance both qualitatively and quantitatively in image generation tasks. They can further produce more diverse images than traditional GAN-based approaches. The large text-guided diffusion model GLIDE~\citep{Nichol2021glide} in combination with contrastive language-image pretraining (CLIP) introduced by~\cite{Radford2021learning} inspired researchers to probe GLIDE for medical knowledge when prompted with relevant clinical text (e.g. ``a histopathological image of the brain'').~\cite{Kather22} found that GLIDE captures representations of key topics in oncology relatively sufficiently, while they lack knowledge in other domains, like radiology. However, with appropriate fine-tuning these large models hold a lot of promise in clinical practice.~\cite{khader2022medical} extended diffusion models to 3D MR and CT images and demonstrated their utility in segmentation tasks, while assessing image quality by radiologists. Latent diffusion models can also be conditioned on text prompts (instead of label vectors) and have recently been used for chest X-ray generation~\citep{chambon2022roentgen}. However, both~\cite{Kather22} and~\cite{Chen21} raise ethical questions around privacy and data biases for the use of synthetic images in medicine and healthcare.

\subsection{Exploring fairness in Health}
Many scholars have recently scrutinised machine learning systems and surfaced different types of biases that emerge through the machine learning pipeline, including problems due to data acquisition protocols, flawed human decision making, missing features, and label scarcity.~\cite{Rajkomar18} identified and characterised various biases that can emerge during model development and exacerbated during model deployment as well as in clinical interactions, while they argue that ensuring fairness in those contexts is essential for the path to advancing health equity. Relevant literature discussed below was inspired by the realisation that, if we break down performance of automated systems that rely on machine learning algorithms (e.g. computer vision, judicial systems) based on certain demographic or socioeconomic traits, there can be vast discrepancies in predictive accuracy across these subgroups. This is alarming for applications influencing human life, and particularly concerning in the context of computer-aided diagnosis and clinical decision making.

One of the first studies to dive into the effect of the training data composition on model performance across genders when using chest X-rays to diagnose thoracic diseases was the one led by~\cite{Larrazabal20}. They found that the prevalence of a particular gender in the training set is directly linked to the predictive accuracy of the model for the same group at test time. In other words, a model trained on a set highly skewed towards female patients would demonstrate higher accuracy for female patients at test time compared on a counterpart trained on a male-dominated set of images. Even though this finding might not come as a surprise, one would expect that a machine learning model used in clinical practice across geographical locations be robust to demographic shifts of this kind. In a similar vein,~\cite{Seyyed21} further explored how differences in age, race / ethinicity and insurance type (as a proxy of socioeconomic status) are manifested in the performance of a classifier operating on chest radiographs. A crucial finding was that the algorithm would exhibit higher false positive rate, i.e. underdiagnose, ethnic minorities. These effects were compounded for intersectional identities (i.e. false positive rate was higher for Black female patients in comparison to Black male patients). Similar findings were reported by~\cite{Puyol22} in a cardiac segmentation task with respect to sex and racial biases, and by~\cite{Gianfrancesco18} in a different modality (electronic health records) for patients with low socioeconomic status.

\section{Method}
We motivate the use of generated data and demonstrate its utility in a number of toy settings, which simulate the problem of having only a few number of samples from the underlying distribution or parts of the underlying distribution. We wish to have high performance despite this lack of data. We demonstrate that even in these toy settings, synthetic data is useful.

We assume we have a dataset $D_\texttt{train} = \{ (\vx_i, y_i, \va_i) \}^N_{i=1}$  where $\vx_i, y_i$ is an image and label pair and $\va_i$ is a list of attributes about the datapoint.
The attributes may include attributes (such as sex, skin type, and age) or the hospital id (in the case of histopathology).
We have an additional dataset $D_\texttt{u} = \{ \hat{\vx_j} \}^M_{j=1}$ of unlabelled images which can be used as desired.
We have a generative model $\hat{p}$ trained with $D_\texttt{train}$ and $D_u$ (we make $\tilde{\theta}$ implicit in the following).
We drop the subscripts in the following for simplicity where obvious.

In order to achieve fairness, we assume we have some ``fair'' dataset $D_f = \{(\vx_i, y_i, \va_i) \}_{i=1}^F$ which consists of samples from the ``fair'' distribution $p_f$ over which we aim to minimize the expectation of the loss.
$f_\theta(\vx)$ is the classifier and $L$ the loss function (e.g. binary cross-entropy). We aim to optimize the following objective:

\begin{equation}
    \min_\theta \sE_{D_f} \left( L(f_\theta(\vx), y, \va) \right)
    \label{eq:expectationfairness}
\end{equation}

We can decompose the data generating process into $p_f(\vx|\va, y)p_f(\va|y)p_f(y)$.
For example, we may have created $D_f$ by sampling uniformly over an attribute (such as sex) and labels.
We assume that the training dataset \datasettrain $\subset D_f$ is sampled from a distribution \ptrain where $p_\texttt{train}(x|\va, y) = p_f(\vx|\va, y)$.
When $p_\texttt{train}(y, \va) \neq p_f(y, \va)$ then we have a  distribution shift between the training and fair distribution (e.g.~the training distribution is more likely to generate images of a particular attribute or combinations of label and attribute than the fair distribution). 

We aim to combine the training dataset \datasettrain and synthetic data sampled from the generative model $\hat{p}$ in order to mimic most closely the fair distribution and improve fairness.
We construct a new dataset $\hat{D}_u$ according to a distribution $\hat{p}_u$ from these distributions using some probability parameter $\alpha$.
\begin{equation}
    (\vx, \va, y) \sim p' = 
    \begin{cases}
     (\vx, \va, y) \sim D_{\texttt{train}} \quad :  \alpha \\
    (\vx, \va, y), x \sim \hat{p}(\vx | y, \va), (\va, y) \sim \hat{p}(\va, y) \quad : (1 - \alpha)  
    \end{cases}
\end{equation}
So instead of minimizing \autoref{eq:expectationfairness}, we minimize the following sum of expectations:
\begin{equation}
\label{eq:loss}
    \min_\theta \alpha \sE_{(\vx, \va, y) \sim D_{\texttt{train}}} \left( L(f_\theta(\vx), \va, y) \right) + (1 - \alpha) \sE_{(\vx, \va, y) \sim \hat{p}} \left( L(f_\theta(\vx), \va, y) \right) 
\end{equation}

The question is then how do we choose $\alpha$ and $\hat{p}(\va, y)$.
For all settings in the main paper, we maintain the label distribution $\hat{p}(y) = p(y)$ but sample uniformly over the attributes $\va$. We validate this choice on dermatology in \ref{app:samplingschemes}. We treat $\alpha$ as a hyperparameter in all settings.

\section{Results on simplified settings}

To build an intuition of why synthetic data is beneficial, we experiment with some toy setups. We assume that to minimise the loss in \autoref{eq:loss}, it is sufficient to minimize the KL divergence between the fair distribution $p_f$ and the combined distribution: $p'$.
We consider the simplified case where the training set is sampled from the fair distribution $p_f$.
The distribution shift arises from the fact that $D_\texttt{train}$ is finite; this may cause the actual distribution of  $p_\texttt{train}$ and $p_f$ to differ.

\subsection{A Bernoulli distribution over a single variable.}
We assume there is a single variable $y$ we are trying to model that takes one of two values: $y \in \{0, 1\}$. 
Because $D_\texttt{train}$ is finite and consists of $N = |D_\texttt{train}|$ samples, its distribution $p_\texttt{train}(y)$ (which we refer to as $t(y)$ for conciseness in the following) may drift from $p_f$ (which we refer to as $p$ in the following).
We also have a generative model $\hat{p}(y)$.
We assume the marginals of these distributions are fixed, so we can only manipulate $\alpha$.

\begin{figure}
    \centering
    \subfigure[{A single run.} We set $N=6, p(y=0) = 0.2, \hat{p}(y=0)=0.5$. To the left we plot the original distributions (real $p$, generated $\hat{p}$ and the  distribution $t$ obtained by sampling $6$ points from $p$). To the right we plot the combined distribution using the optimal $\alpha$. We can see that in this case, we find $\alpha=0.1$ is optimal and with the combined distribution, we can perfectly match the true distribution.  ]{\includegraphics[clip, trim=0cm 0cm 0cm 0cm,width=0.49\linewidth]{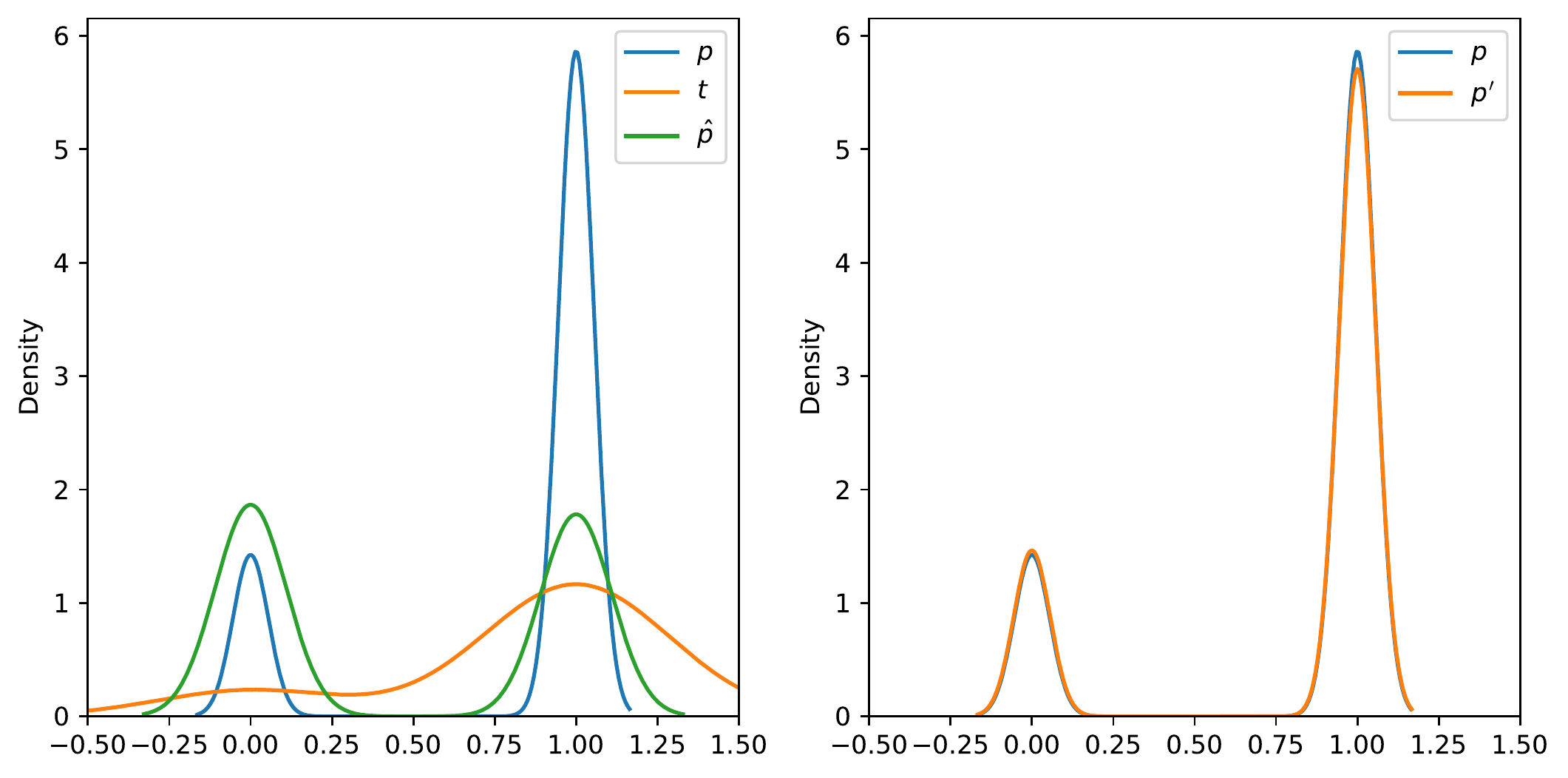} \label{fig:bernoulli_single}} \hspace{1em}
    \subfigure[{Comprehensive runs.} We set $N=6, p(y=0) = p = 0.2, \hat{p}(y=0) = \hat{p} = 0.5$. We then vary each value in turn and plot the optimal $\alpha$ along with the $95\%$ confidence interval. We vary $p, \hat{p}$ between $(0, 1)$ and $N$ between $(0, 100)$. The x-axis plots the normalized value of each variable between (0, 1).  ]{\includegraphics[width=0.4\linewidth]{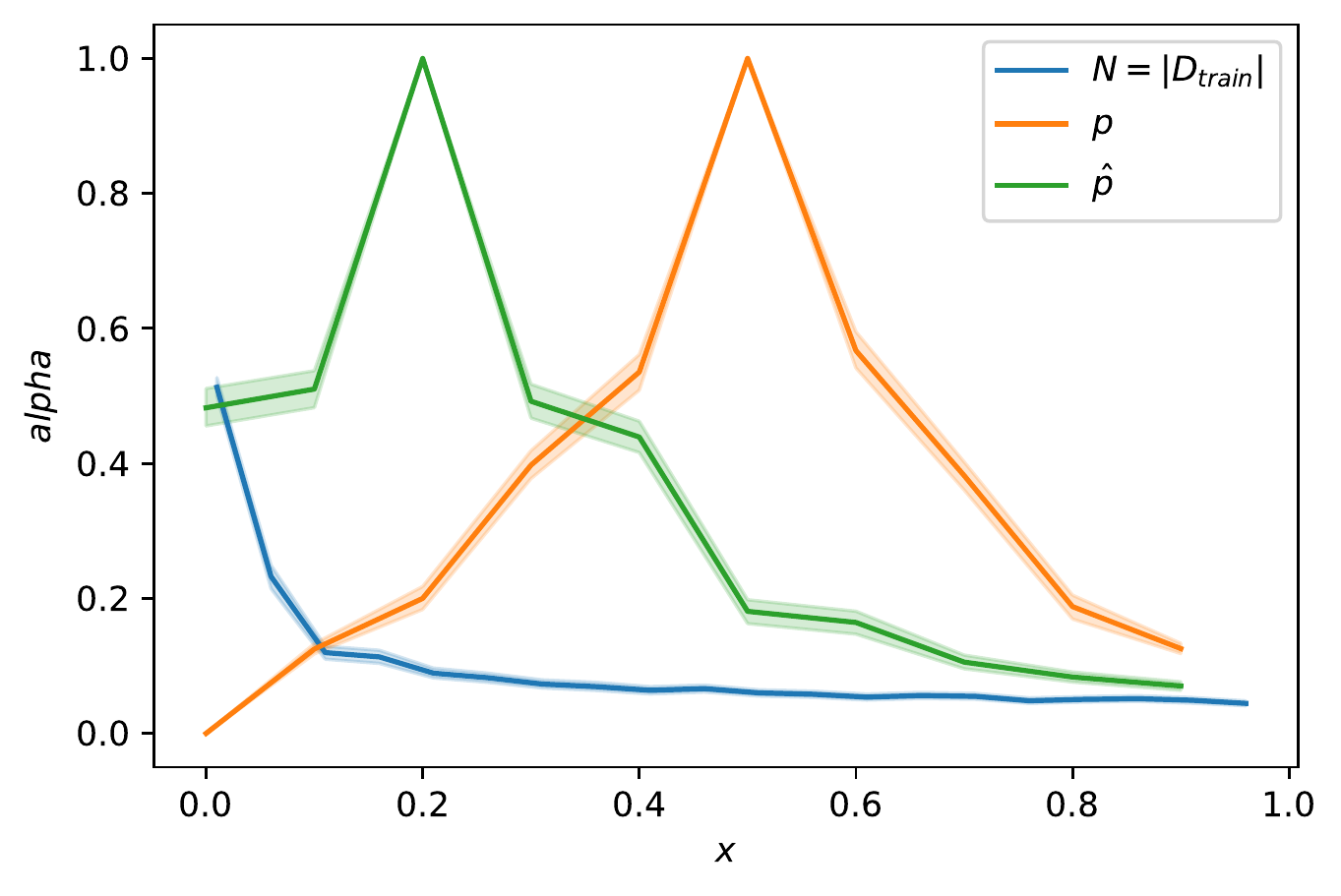}  \label{fig:bernoulli_comprehensive}}
    \caption{{A Bernoulli distribution.} We visualise results for a single run and a comprehensive set of runs to demonstrate that we can leverage generated data to better model the true distribution if there is a distribution shift between the data and true distribution.}
    \label{fig:bernoulli}
\end{figure}

We want to explore how we should combine $t(y)$ and $\hat{p}(y)$.
We can model this by minimizing the KL divergence between $p$ and $p'(y) = t(y) (1 - \alpha) + \hat{p}(y) \alpha$: $KL(p || p')$ where $\alpha$ is bounded by $[0, 1]$.
For example, if $p = \hat{p}$, then we should just sample from the generative model irrespective of $t$. Otherwise, there is a trade-off, and we can find the optimal $\alpha$ analytically or experimentally.
We compare the distributions with and without generated data for a specific example in \autoref{fig:bernoulli_single}.
By combining real and generated data we can perfectly match the $p$ distribution.

Next, we consider the more general setting.
We evaluate how using a generative model improves performance as we vary (1) the number of samples $N$ from $p$ when creating $D_\texttt{train}$; (2) $\hat{p}(y)$ and (3) $p(y)$.
For each choice of values, we resample $t(y)$ 1000 times and experimentally find the optimal $\alpha$.
We plot the optimal $\alpha$ as we vary these values in \autoref{fig:bernoulli_comprehensive}.
As can be seen, using generated data is generally helpful (by the fact that $\alpha > 0$).
We can also see various properties.
First, as $N$ increases (e.g.~there is less distribution shift between $D_\texttt{train}$ and $p$), using generated data becomes less helpful (as $\alpha$ goes to $0$).
Second, generated data is most useful when its distribution is more similar to the true distribution $p$, but generated data is helpful even in the face of some distribution shift between $p$ and $\hat{p}$.
This is perhaps surprising: even in this very simple setting we can see the value of combining the distribution of the training set with another (that of the generated data) in order to mimic the true underlying distribution.

\subsection{A Bernoulli distribution over a labelled and hidden variable.}
The above case does not take into account that we can resample the dataset and use the conditioning variable of the generative model to make $D_\texttt{train}$ and $\hat{p}$ more similar to $p$.
However, in reality while we may be able to resample on a labelled variable, there will be many hidden attributes that are important for fairness and generalization.
As we do not have labels for these hidden attributes, we {\em cannot} resample over them to make the distribution more similar to $p$.

To explore this in a simplistic setting, we proceed as follows.
We assume we have two variables $a, y$ where $y$ is the label and $a$ some hidden attribute
which we want to be fair over (and generalise over). Again, they follow a Bernoulli distribution: $a \in \{0, 1\}$ and $y \in \{0, 1\}$.

We condition on the label and can sample arbitrarily according to the label, but this gives a {\em fixed} distribution over attributes conditioned on the label.  More specifically, we can sample arbitrarily to create $\bar{t}(y)$ and $\bar{p}(y)$ (provided $t(y), \hat{p}(y) > 0$) but $\bar{t}(a|y) = t(a|y)$ and $\bar{p}(a|y) = \hat{p}(a|y)$ is fixed.
Again, the question is how to optimally combine $t(y, a)$ and  $\hat{p}(y, a)$.
So now 
\begin{align}
p'(y, a) &= t(a|y)\bar{t}(y) + \hat{p}(a|y)\bar{p}(y) \\
\sum_y \bar{t}(y) + \bar{p}(y) &= 1 
\end{align}
We want to minimize the KL divergence between these two distributions, which can be factorized into the sum of the KL divergence between the marginal distribution over $a$ and conditional distribution over $y$.
\begin{align}
KL(p'(a, y) || p(a, y)) = KL(p'(y) || p(y)) + KL(p'(a|y) || p(a|y))
\end{align}
This is hard to solve analytically but we can find the optimal distributions of $\bar{t}, \bar{q}$ for a given set of distributions experimentally.

\begin{figure}
    \centering
    \includegraphics[width=0.5\linewidth]{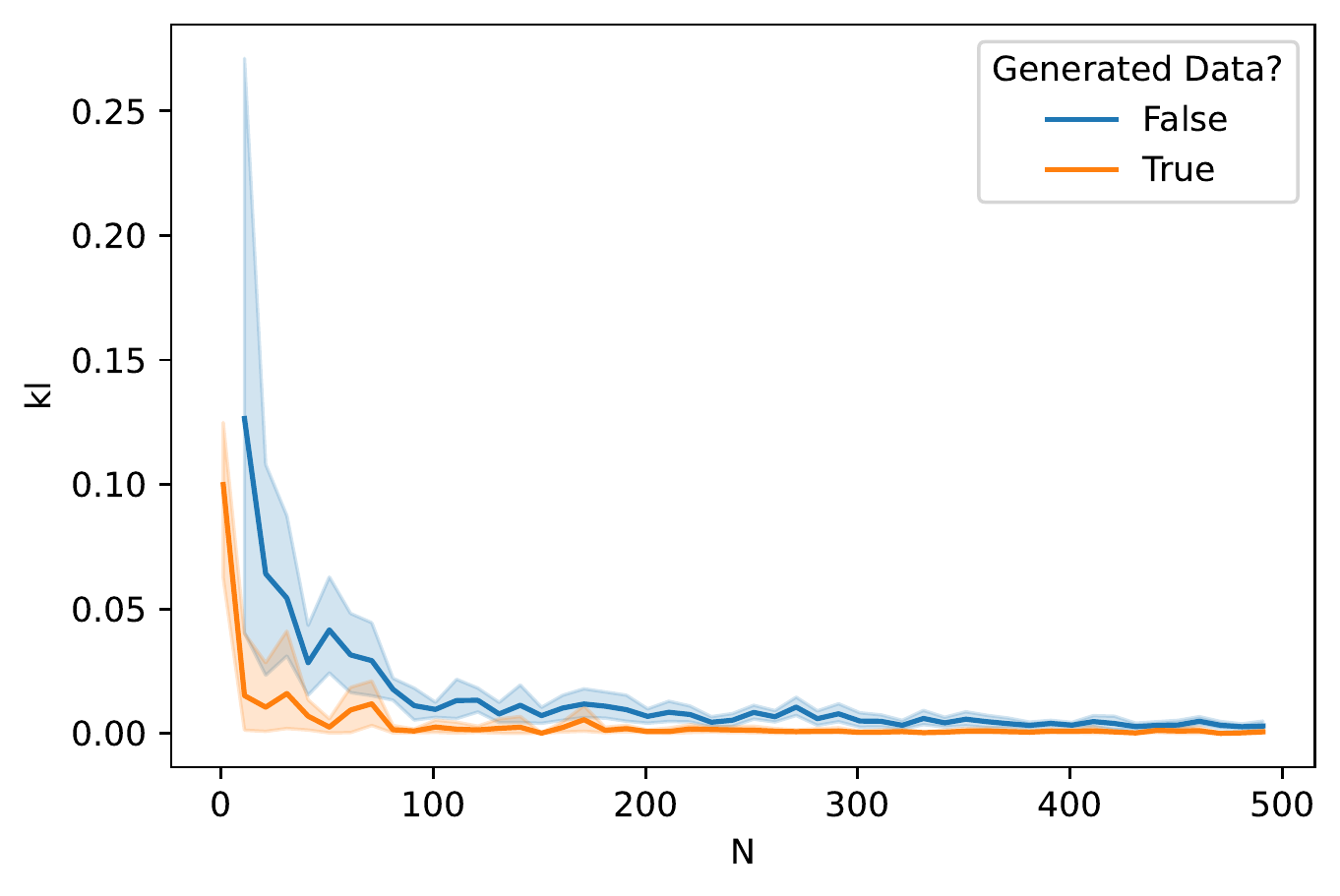}
    \caption{A Bernoulli distribution of two variables.  We vary the number of samples $N$ and experimentally find the optimal $\bar{p}, \bar{t}$. We then compare the KL divergence when using and not using the generated data. We repeat each experiment $10$ times for each value of $N$ to obtain the $95\%$ confidence intervals. }
    \label{fig:bernoulli_hidden}
\end{figure}

We randomly choose $p = \begin{pmatrix} 0.35 & 0.4 \\ 0.1 & 0.15 \end{pmatrix}$, $p'= \begin{pmatrix} 0.7 & 0.2 \\ 0.05 & 0.05 \end{pmatrix}$ and plot the KL with and without generated data for varying $N$ in \autoref{fig:bernoulli_hidden}.
For small values of $N$, the data distribution is not able to capture the full distribution of $p$, giving a KL divergence of $\infty$.
However, even for large values of $N$, using the additional generated data improves the KL divergence.
Again, this toy example demonstrates that there is value in leveraging generated data in a low data setting {\em even} when the training set is sampled from the same distribution as the fair distribution used for evaluation.

\subsection{A high dimensional Gaussian Mixture Model (GMM)}
\label{sec:toy_example}
Here we demonstrate that additional unlabelled data can improve classification performance in a simple setting.
We assume that points come from an underlying GMM where each mixture in the GMM corresponds to a different class.
This GMM can have a varying number of dimensions (the number of features) and components (the number of classes).
We demonstrate that using a generative model fit to some number of unlabelled points in combination with the labelled points can lead to improved performance of a trained classifier.

Given some number of labelled and unlabelled points, the aim is to train a classifier that performs well on a held out validation set.
We do this by fitting a GMM to the unlabelled points and then using the labelled points to determine what class each mode belongs to by minimizing the classification error on the labelled points.
We then combine the labelled points and some number of generated points by sampling from the GMM to form the training set. 
This is used to train the classifier, which is evaluated on the validation set.

\begin{figure}[t]
\centering
\begin{overpic}[width=0.24\textwidth]{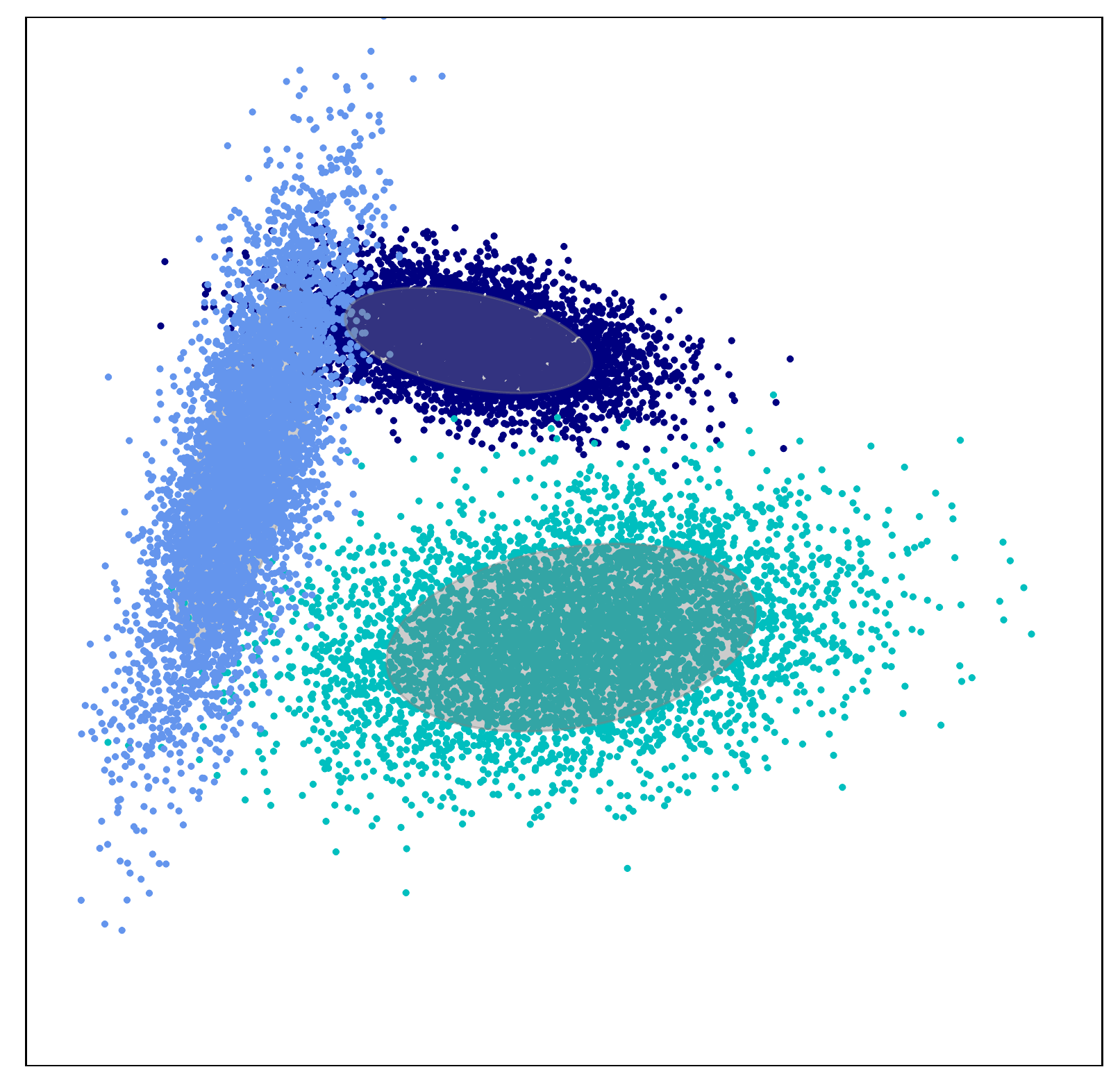}
\put(35,98){\tiny True Distribution}
\end{overpic}
\begin{overpic}[width=0.24\textwidth]{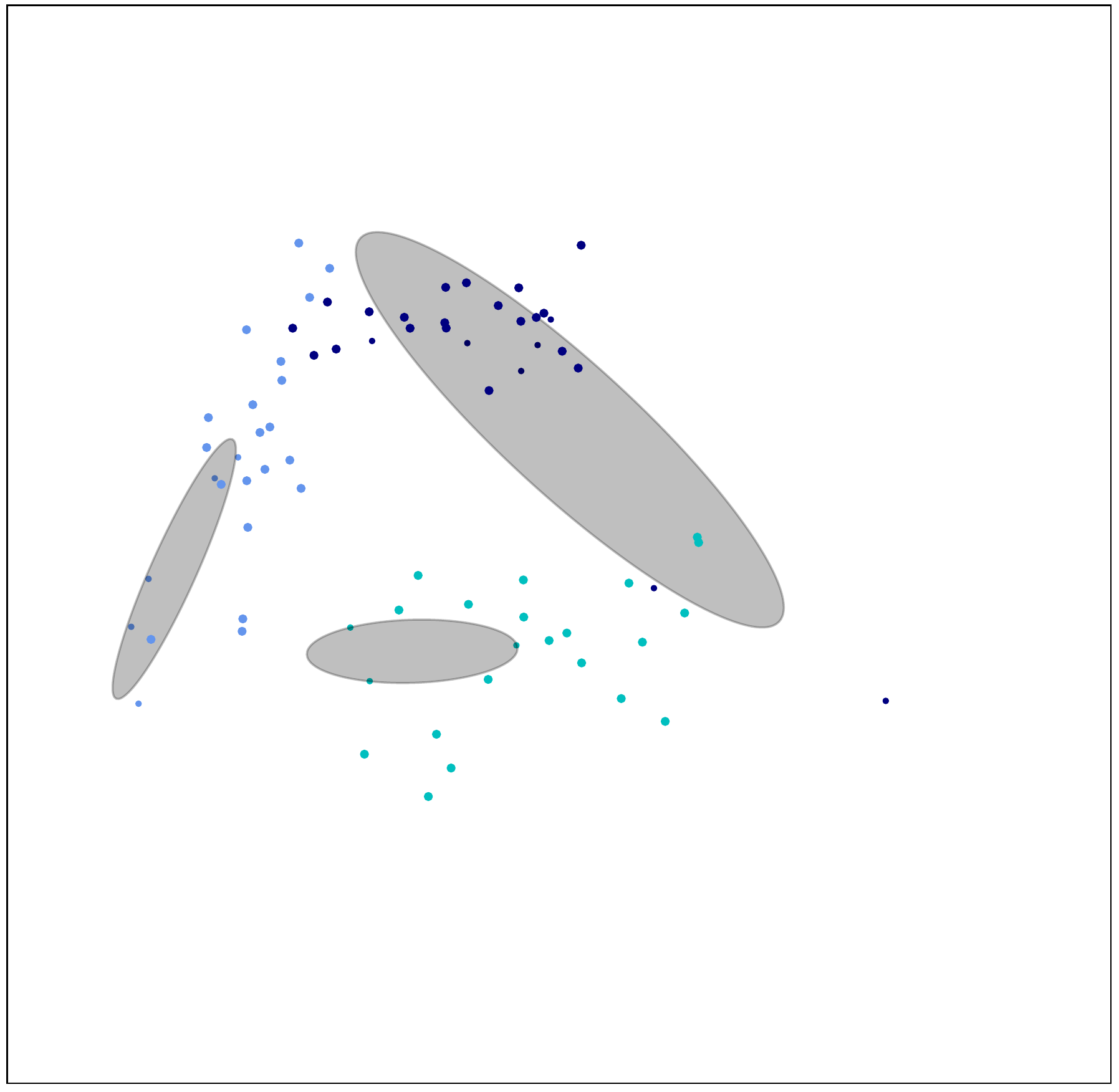}
\put(32,98){\tiny 5 Unlabelled Points}
\end{overpic}
\begin{overpic}[width=0.24\textwidth]{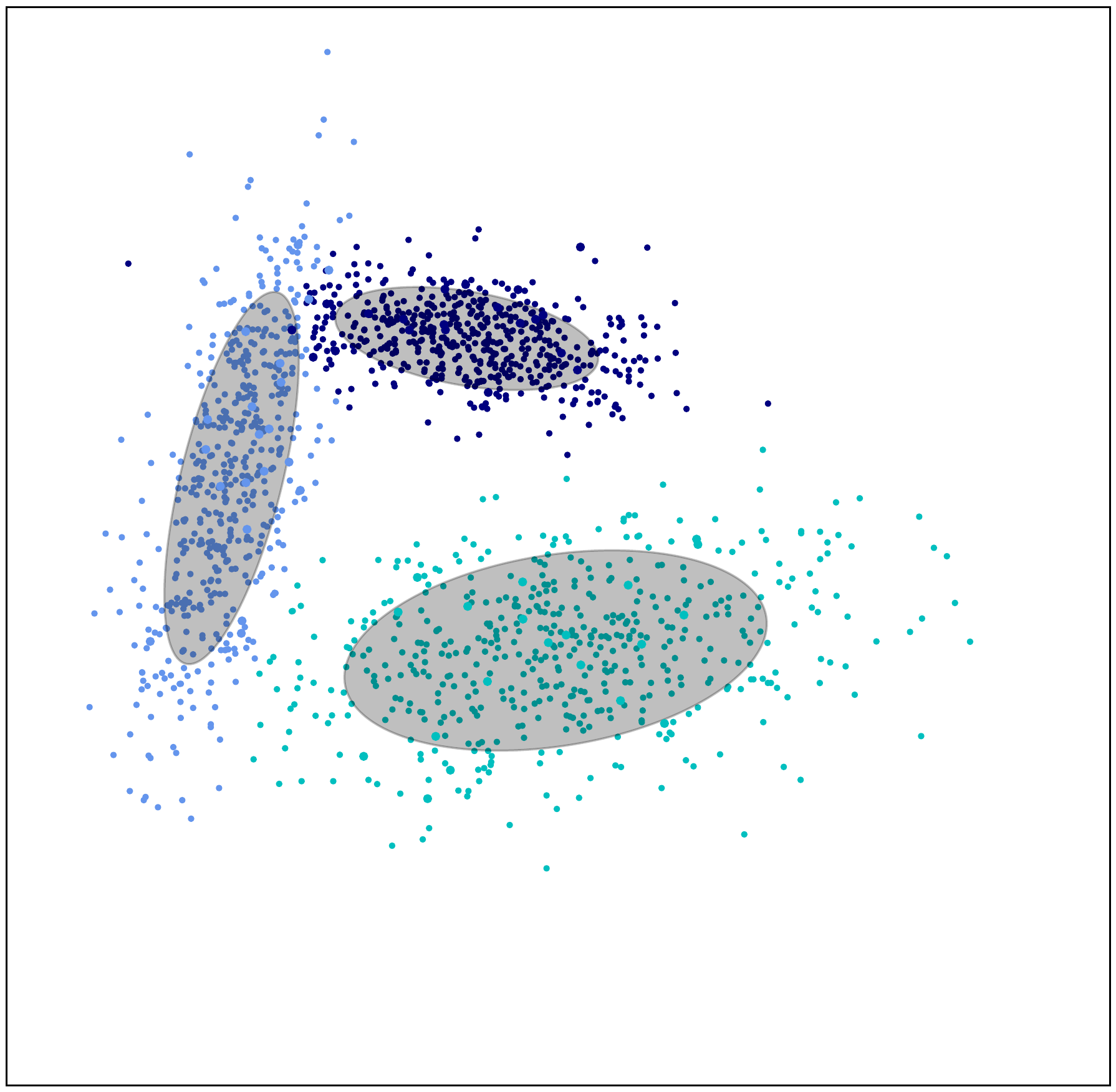}
\put(27,98){\tiny 500 Unlabelled Points}
\end{overpic}
\begin{overpic}[width=0.24\textwidth]{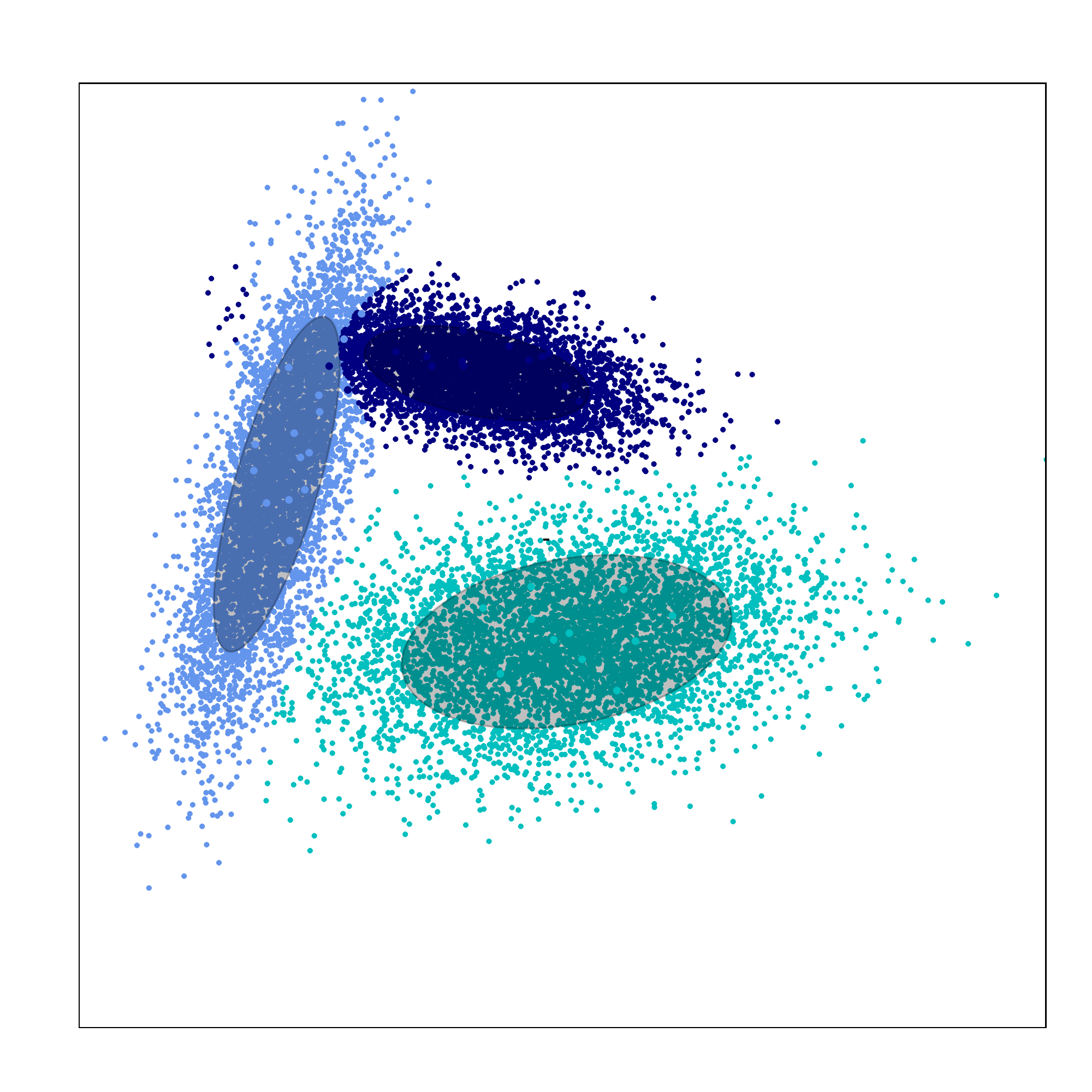}
\put(27,98){\tiny 5000 Unlabelled Points}
\end{overpic}
\caption{Visualising the fit GMM model for varying numbers of unlabelled points. Large points are labelled points, small points are unlabelled ones. Given only 500 unlabelled points, the fit distribution is similar to the true and with 5000 points, it matches the original. }
\label{fig:gmm_visualisation}
\end{figure}

We plot the evolution of the GMM being fit to the unlabelled points as the number of unlabelled points varies next to the true, underlying GMM in \autoref{fig:gmm_visualisation}.
We plot the fitted GMM for varying numbers of unlabelled points.
With only 500 unlabelled points, the fitted GMM closely matches the true distribution.

\begin{figure}[t]
\centering
\begin{overpic}[width=0.32\textwidth]{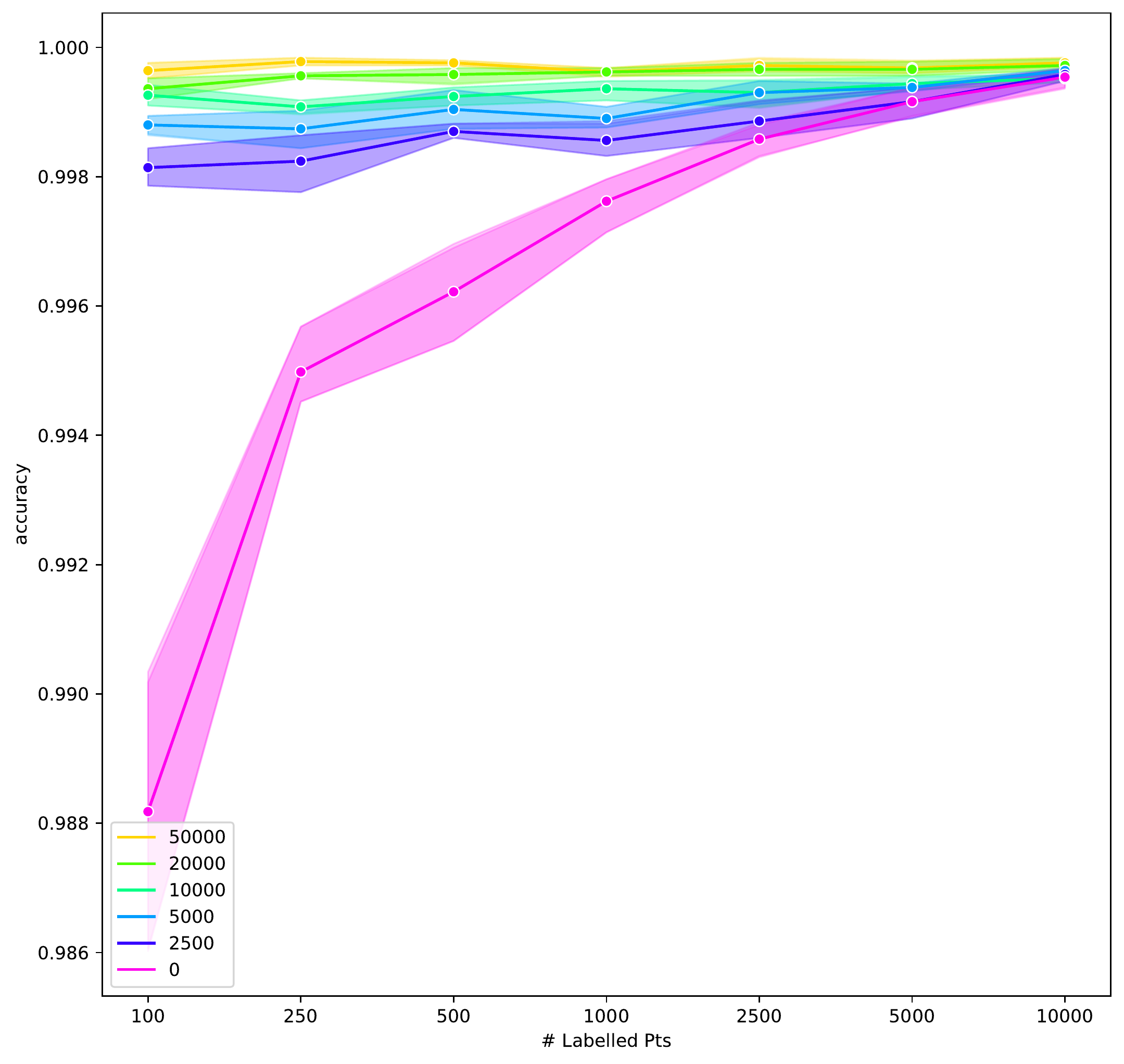}
\put(35,97){\tiny 2 Components, 64 Dimensions}
\end{overpic}
\begin{overpic}[width=0.32\textwidth]{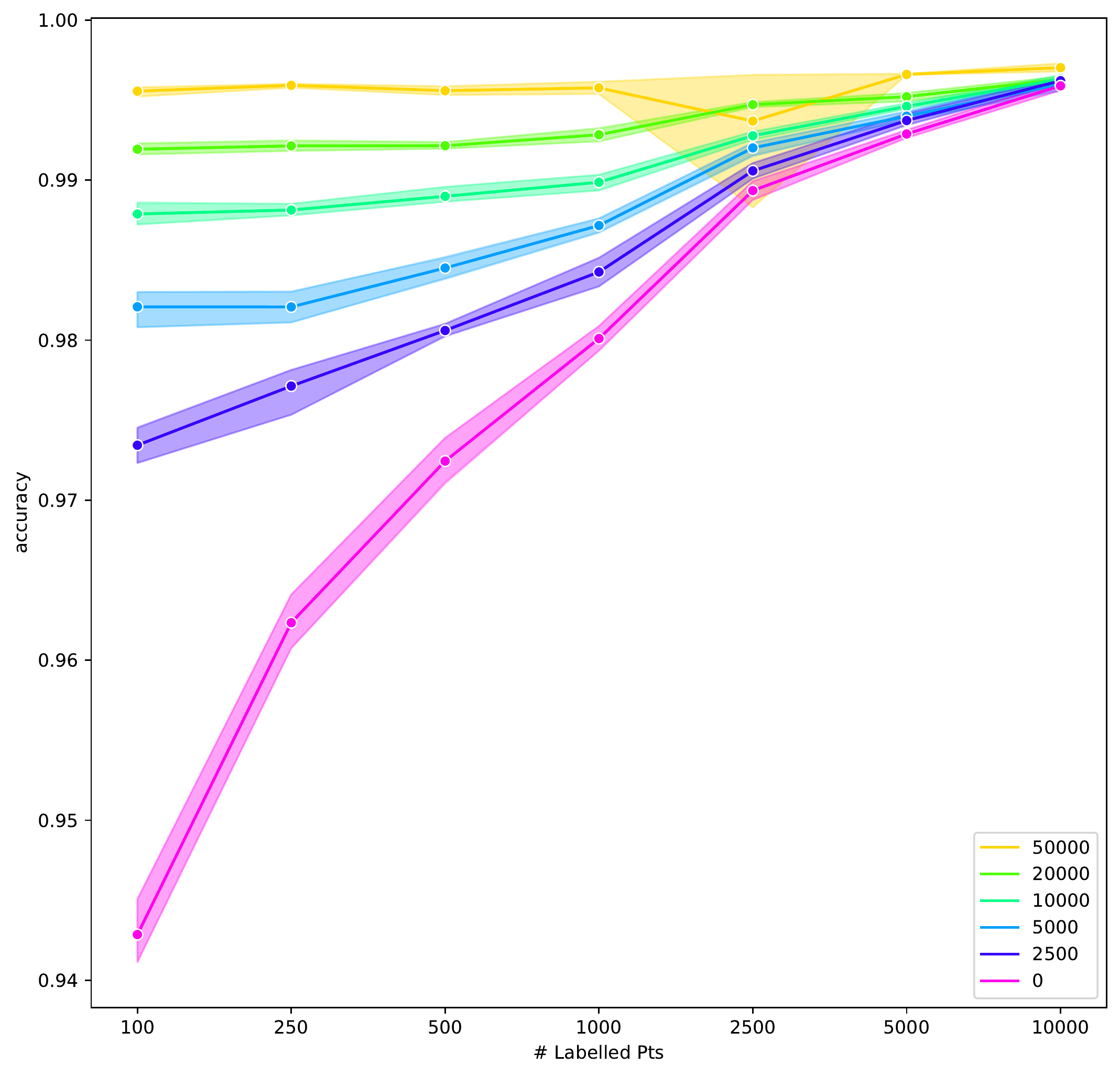}
\put(32,97){\tiny 5 Components, 64 Dimensions}
\end{overpic}
\begin{overpic}[width=0.317\textwidth]{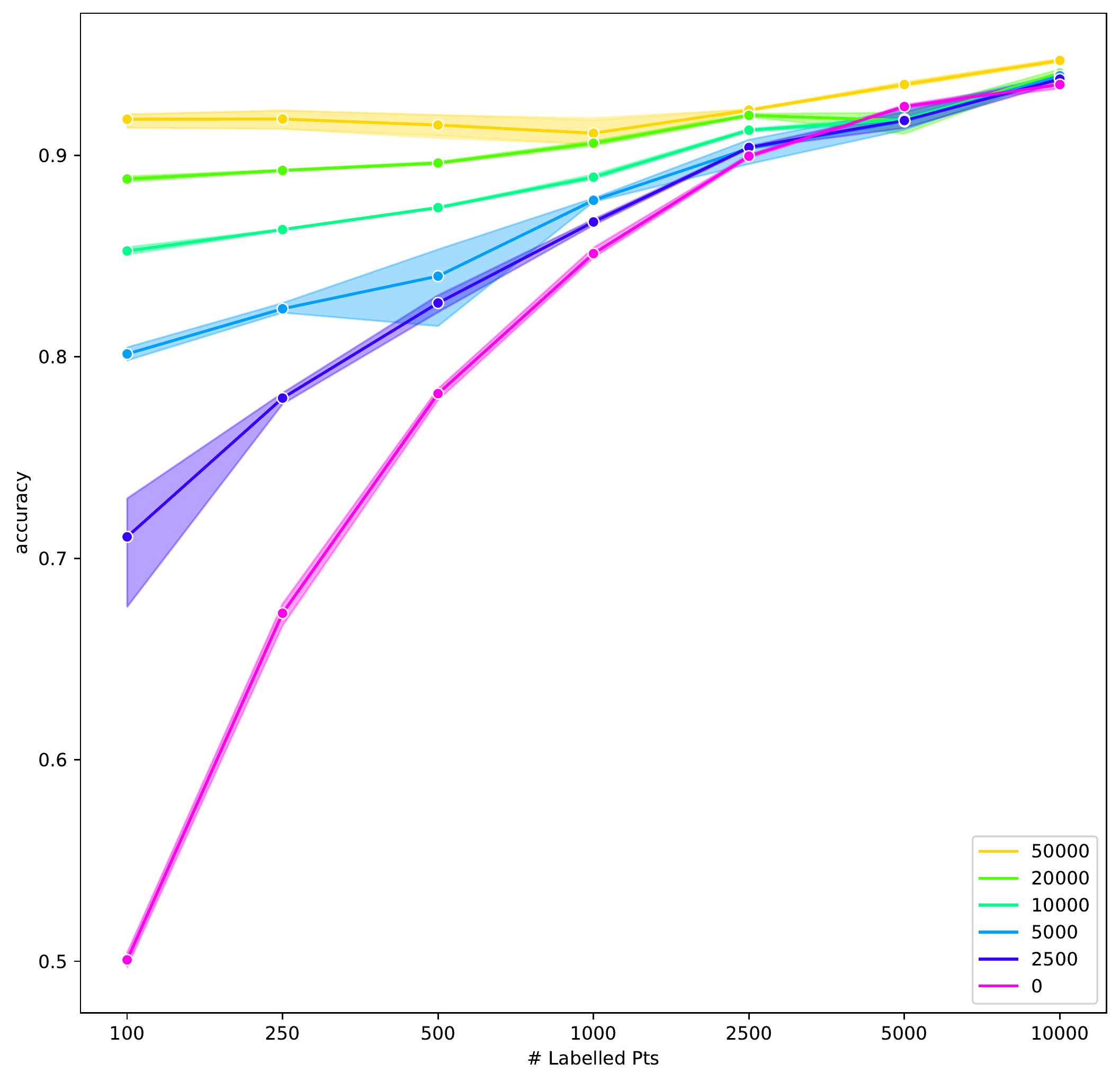}
\put(27,97){\tiny 10 Components, 1024 Dimensions}
\end{overpic}
\caption{Results for a downstream classifier when using a varying number of generated points from the generative model (which are shown in different colours) and labelled points, which varies along the x-axis. We can see that using additional generated points helps in all these settings (which vary in terms of the number of components and dimensions).}
\label{fig:gmm_results}
\end{figure}

We then run the full pipeline including the downstream classifier for varying numbers of labelled (sampled points), generated points, components, and dimensions.
We always use 10K unlabelled points.
To create the GMM, we randomly sample means from the range $[0, 1]$ for each dimension and we create a full covariance metric under a normal distribution with a scale of $0.1$.
We plot the results in \autoref{fig:gmm_results}. 
In these three cases, we can see that using additional generated data improves performance.
However, this approach provides limited benefits if (1) the classifier already performs near perfectly using the labelled data;  or (2) the mixtures overlap so much that fitting the generative model gives a poor estimate of the underlying distribution.

\section{Models}

\begin{table}[h!]
\subtable[Classifier hyperparameters]{
\begin{tabular}{cccc}
\hline
\textbf{Hyperparameter}        & \textbf{Histopathology}       & \textbf{Radiology} & \textbf{Dermatology} \\ \hline
Pretraining dataset & None & JFT & JFT \\
ResNet depth & 152 & 152 &	101 \\
ResNet width & 1 & 2 & 3 \\
Optimizer & Adam & Adam & Momentum \\ 
Init. learning rate & $1e^{-4}$ & $1e^{-5}$ & 0.01 \\
Learning schedule & constant & cosine & cosine \\
Training steps & 200K & 50K & 10K \\
Batch size & 32 & 64 & 96 \\
Weight decay & 0.0 & $1e^{-5}$ & $1e^{-4}$ \\
Dropout & 0.0 & 0.0 & 0.0 \\
\hline
\end{tabular}
\label{tab:classifier_params}
}
\hfill%
\subtable[Generative model hyperparameters]{
\begin{tabular}{cccc}
\hline
\textbf{Hyperparameter}        & \textbf{Histopathology}       & \textbf{Radiology} & \textbf{Dermatology} \\ \hline
Learned variance & No & No & No \\
$\#$ Channels & 128  & 192 & 192 \\
$\#$ Head channels & 64 & 64 & 64 \\
$\#$ Residual blocks & 2 & 3 & 3 \\ 
Channel multipliers & $(1, 1, 2, 3, 4)$ & $(1, 2, 3, 4)$ & $(1, 2, 3, 4)$ \\
Attention resolutions & $(32, 16, 8)$ & $(32, 16, 8)$ & $(32, 16, 8)$ \\
$\#$ Diffusion timesteps & 1000 & 1000 & 1000 \\
Learning rate & 1$e^{-4}$ & 1.5$e^{-5}$ & 1.5$e^{-4}$ \\
Batch size & $64$ & 128 & 128 \\
\hline
\end{tabular}
\label{tab:generative_params}
}
\hfill%
\subtable[Upsampler hyperparameters]{
\begin{tabular}{ccc}
\hline
\textbf{Hyperparameter}        & \textbf{Radiology} & \textbf{Dermatology} \\ \hline
Learned variance & No & No \\
Output resolution & 224 & 256 \\
$\#$ Channels & 128  & 128 \\
$\#$ Head channels & 64 & 64 \\
$\#$ Residual blocks & 2 & 2 \\ 
Channel multipliers & $(1, 1, 2, 2, 4, 4)$ & $(1, 1, 2, 2, 4, 4)$ \\
Attention resolutions & $(32, 16, 8)$ & $(32, 16, 8)$ \\
$\#$ Diffusion timesteps & 1000 & 1000 \\
Learning rate & 1$e^{-4}$ & 1$e^{-4}$ \\
Batch size & $32$ & $32$ \\
\hline
\end{tabular}
\label{tab:upsampler_params}
}
\caption{Model hyperparameters for classifier and generative (low-resolution \& upsampler) models.}
\label{tab:model_hyperparams}
\end{table}

\paragraph{Upsampler preprocessing.} Whenever we require an upsampler (i.e. in radiology and dermatology), we train it by preprocessing the original images with the following steps:

\begin{enumerate}
    \item upsample images from 64x64 input resolution to the desired output resolution with bilinear interpolation and use anti-alias with 0.5 probability
    \item add random gaussian noise with 0.2 probability and $\sigma=4.0$ (in the $[0, 255]$ range)
    \item apply random gaussian blurring with a $7\times7$ kernel and $\sigma_{mean}=0.0$, $\sigma_{std}=0.2$
    \item quantize image to 256 bins
    \item normalize image to $[-1, 1]$ range
\end{enumerate}

\paragraph{Dealing with missing labels.} For both the generative model and the upsampler we fill the conditioning vectors with zeros (indicating an invalid vector) for the unlabelled data. This allows us to use classifier-free guidance~\citep{Ho2022classifier} in order to make images more ``canonical" with respect to a given label or property.

In this section, we describe the exact model architecture used for the trained diffusion models and classifiers, as well as the hyperparameters used for the presented results. We note that hyperparameters were selected based on the baseline model performance on the respective in-distribution validation sets and held constant for the remaining methods. This means that we do not fine-tune hyperparameters for each method (other than the baseline) separately. We use the DDPM diffusion model as presented by~\cite{Ho20, Nichol21,Ho22} for the generation and the upsampler (only for radiology and dermatology do we require higher resolution images). The backbone model is always a UNet architecture. The hyperparameters used for the cascaded diffusion models were based on the standard values mentioned in the literature with minimal modifications. We present all hyperparameters in~\autoref{tab:model_hyperparams}.



\subsection{Standard augmentations} 

\paragraph{Histopathology.} For this modality augmentations include brightness, contrast, saturation and hue jitter. Hue and  saturation were found to be sufficient to achieve high quality results by \cite{Tellez19}.

\paragraph{Chest Radiology.}
The heuristic augmentations considered for this modality include: random horizontal flipping, random crop to $202 \times 202$ resolution, resizing to $224 \times 224$ with bilinear interpolation and antialias, random rotation by 15 degrees, shifting luminance by a value sampled uniformly from the $[-0.1, 0.1]$ range and shifting contrast using a value uniformly sampled from the $[0.8, 1.2]$ range (i.e. pixel values are multiplied by the shift value and clipped to remain within the $[0, 1]$ range.

\paragraph{Dermatology.} For this modality we use the following heuristic augmentations: random horizontal and vertical flipping, adjust image brightness by a random factor (maximum $\delta=0.1$), adjusting image saturation by a random factor (within the $[0.8, 1.2]$ range), adjusting the hue by a random factor (maximum $\delta=0.02$), adjusting image contrast by a random factor (within the $[0.8, 1.2]$ range), random rotation within the $[-150, 150]$ range and random gaussian blurring with standard deviation uniformly sampled from the following values $\{0.001, 0.01, 0.1, 1.0, 3.0, 5.0, 7.0\}$.

\subsection{Baselines}
\label{subsec:baselines}

\paragraph{Histopathology.}
For this modality, all models use the same ResNet-152 backbone. We compare \textbf{(1)} a baseline using no augmentation (\textit{Baseline}) and \textbf{(2)} one using standard color augmentations (\textit{Color augm}) as applied in standard ImageNet training. This augmentation includes brightness, contrast, saturation and hue jitter. Hue and  saturation were found to be sufficient augmentations to achieve the highest quality results by \cite{Tellez19}, hence we do not evaluate other heuristic augmentations. Our baseline does not use pretraining, as it was previously found to not yield any benefits on this particular dataset by~\cite{Wiles21}. We also compare the models to those applying heuristic colour augmentations on top of the synthetic data.

\paragraph{Chest Radiology.}
All models use the same BiT-ResNet-152 backbone.~\citep{Kolesnikov20}
We consider baselines that use \textbf{(1)} different pretraining, \textbf{(2)} different heuristic augmentations and combinations thereof, and \textbf{(3)} use the focal loss.
We investigate using JFT \citep{Sun17} and ImageNet21K \citep{Deng09} for pretraining to explore how much different pretraining datasets impact final results.
We investigate using RandAugment \citep{Cubuk20}, ImageNet Augmentations as described above, and RandAugment + ImageNet Augmentations  to determine how much performance we can gain by utilizing heuristic augmentations.
Finally, we consider using the focal loss \citep{Lin17} which was developed to improve performance on imbalanced datasets.

\paragraph{Dermatology.}
All models use the same BiT-ResNet backbone.~\citep{Kolesnikov20}
We consider baselines that use \textbf{(1)} different pretraining, \textbf{(2)} different heuristic augmentations, \textbf{(3)} resample the dataset and \textbf{(4)} use the focal loss.
We investigate using JFT \citep{Sun17} and ImageNet21K \citep{Deng09} for pretraining.
We investigate using RandAugment \citep{Cubuk20}, ImageNet Augmentations, and RandAugment + ImageNet Augmentations.
We then resample the dataset so that the distribution over attributes is even (we upsample samples from low data regions so that they occur more frequently in the dataset).
Finally, we consider using the focal loss \citep{Lin17} which was developed to improve performance on imbalanced datasets.

\section{Evaluation details}

\subsection{Experimental setup}
In order to account for potential variations with respect to model initialization, we evaluate all versions of our model and baselines with 5 different initialization seeds and report the average and standard deviation across those runs for all metrics. We run all experiments on Tensor Processing Units (TPUs).

\subsection{Fairness metrics.} Different definitions of fairness have been proposed in the literature, which are often at odds with each other~\citep{Ricci22}. In this section we discuss our choice of fairness metrics for each modality. In histopathology, we use the gap between the best and worst performance amongst the in-distribution hospitals. For radiology, we consider \textit{AUC parity}, namely the parity of the area under the ROC for different demographic subgroups identified by the sensitive attribute $A$, which can be seen as the analogous of equality of accuracy~\citep{castelnovo2022clarification}. For this modality, we therefore report AUC gap between males and females in~\autoref{fig:eval_auc_fairness_gap}. We consider this most relevant given that the positive / negative ratio of samples across all conditions is very imbalanced.

In dermatology, we report the gap between the best subgroup and worst subgroup performance, where subgroups are defined based on the sensitive attribute axis under consideration in~\autoref{fig:eval_high_risk_fairness}. We also report the central best estimate for the a posteriori estimate of performance (i.e. top-3) difference between a group and its outgroup. The steps to obtain the values plotted in~\autoref{fig:fairness_results} are the following:

\begin{enumerate}
\item We define a group (and its matching outgroup) as the set of instances that are characterised with a particular value of a sensitive attribute $A=a$, i.e. $\textrm{group} = \{(\vx_i, c_i) | \va_i = a\}$ and $\textrm{outgroup} = \{(\vx_i, c_i) | \va_i \neq a\}$. Here $A \in \{\textrm{sex}, \textrm{skin type}, \textrm{age}\}$.
\item We assume a uniform, beta distribution $Beta(1, 1)$, as a prior for the performance difference between $top_3^{\textrm{group}} - top_3^{\textrm{outgroup}}$ and fit to the observed data.
\item We sample $n=100,000$ samples from the estimated posterior differences $\hat{top}_3^{\textrm{group}} - \hat{top}_3^{\textrm{outgroup}}$ and report the spread, i.e. the standard deviation of the maximum a-posteriori estimates, which can be interpreted as the central best estimate for fairness.
\end{enumerate}

\subsection{Set-up for distribution shift estimation}
\label{subsec:dist_shift}
We compute domain mismatches considering the space where decisions are performed, i.e. the output of the penultimate layer of each model. We thus project each data point from the input space of size $\mathbb{R}^{64 \times 64}$ to a representation of size $\mathbb{R}^{6144}$ and then compute the Maximum Mean Discrepancy (MMD) between two distributions (i.e. datasets). Given two distributions $\mathcal{U}$ and $\mathcal{Z}$, their respective samples $\hat{\mathcal{U}}=\{u_1, \ldots, u_N\}$ and $\hat{\mathcal{Z}}=\{z_1, \ldots, z_N\}$, and a kernel $K$, we considered the MMD empirical estimate as defined below:  

\begin{equation} 
\widehat{\text{MMD}}^2(\mathcal{U}, \mathcal{Z}) = \frac{1}{N(N-1)} \sum_{i,j=1}^N K(u_i, u_j) + \frac{1}{N(N-1)} \sum_{i,j=1}^N K(z_i, z_j) - \frac{2}{N^2} \sum_{i,j=1}^N K(u_i, z_j).
\label{eq:mmd_empirical}
\end{equation}

We use a cubic polynomial kernel in order to minimize the number of hyperparameters to be selected, as well as to capture mismatches between up to the third-order moments of each distribution. We compute $S=30$ estimates of MMD between all pairs of domains using representations from the different models considering samples of size $N=300$. The Mann-Whitney U test under a significance level of $95\%$ is then carried out to test for the hypothesis that, for a fixed pair of distributions, the data augmentation strategy had a significant effect on the estimate MMD values. Importantly, we highlight that models were trained under the same experimental conditions so that our analysis is capable of isolating the effect of the data augmentation protocol on the estimated pairwise distribution shifts.

\section{Additional results}

\begin{figure}
    \centering
    \subfigure[Histopathology - healthy cells]{ \includegraphics[width=0.42\linewidth]{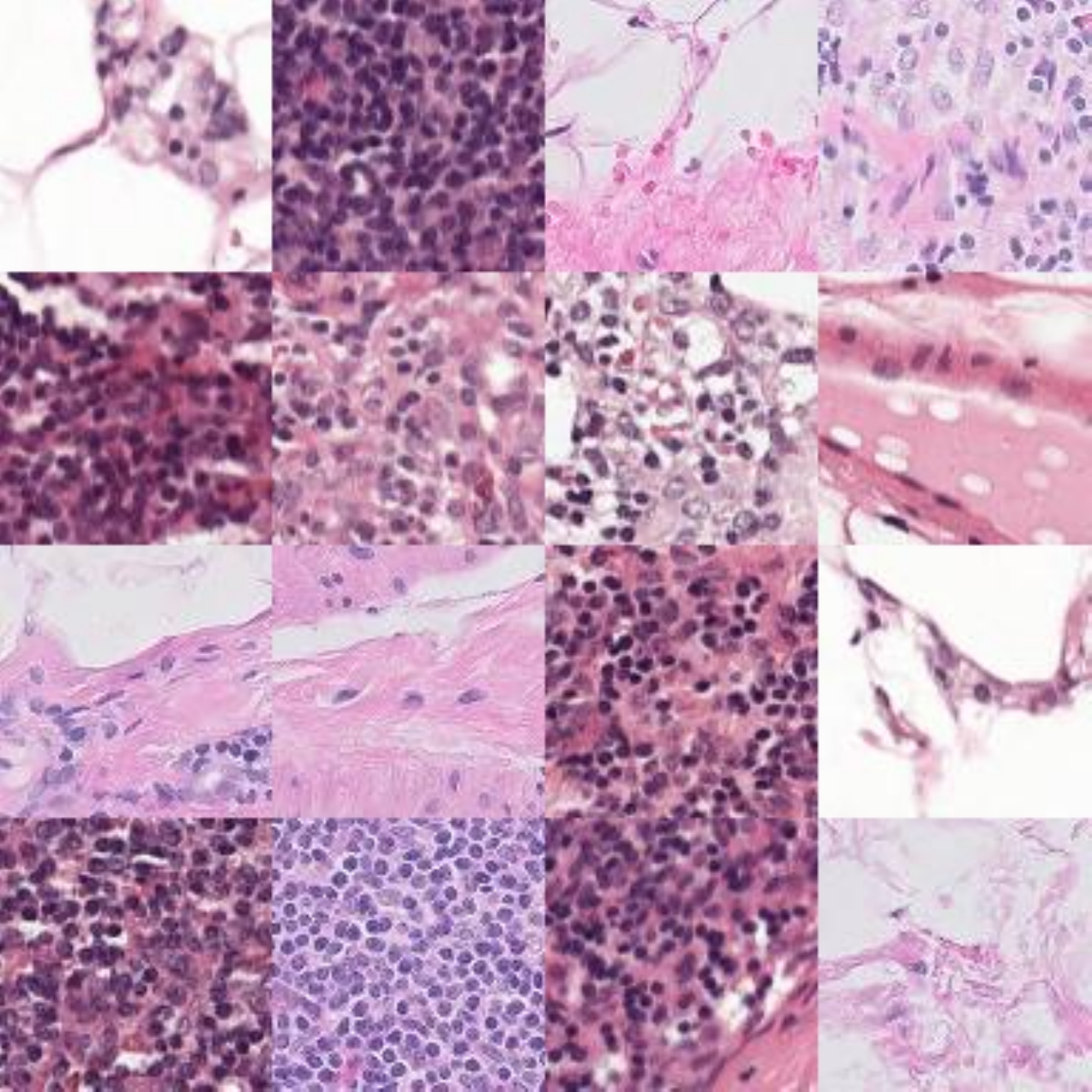}}\hspace{1.0cm}
    \subfigure[Histopathology - cancerous cells]{ \includegraphics[width=0.42\linewidth]{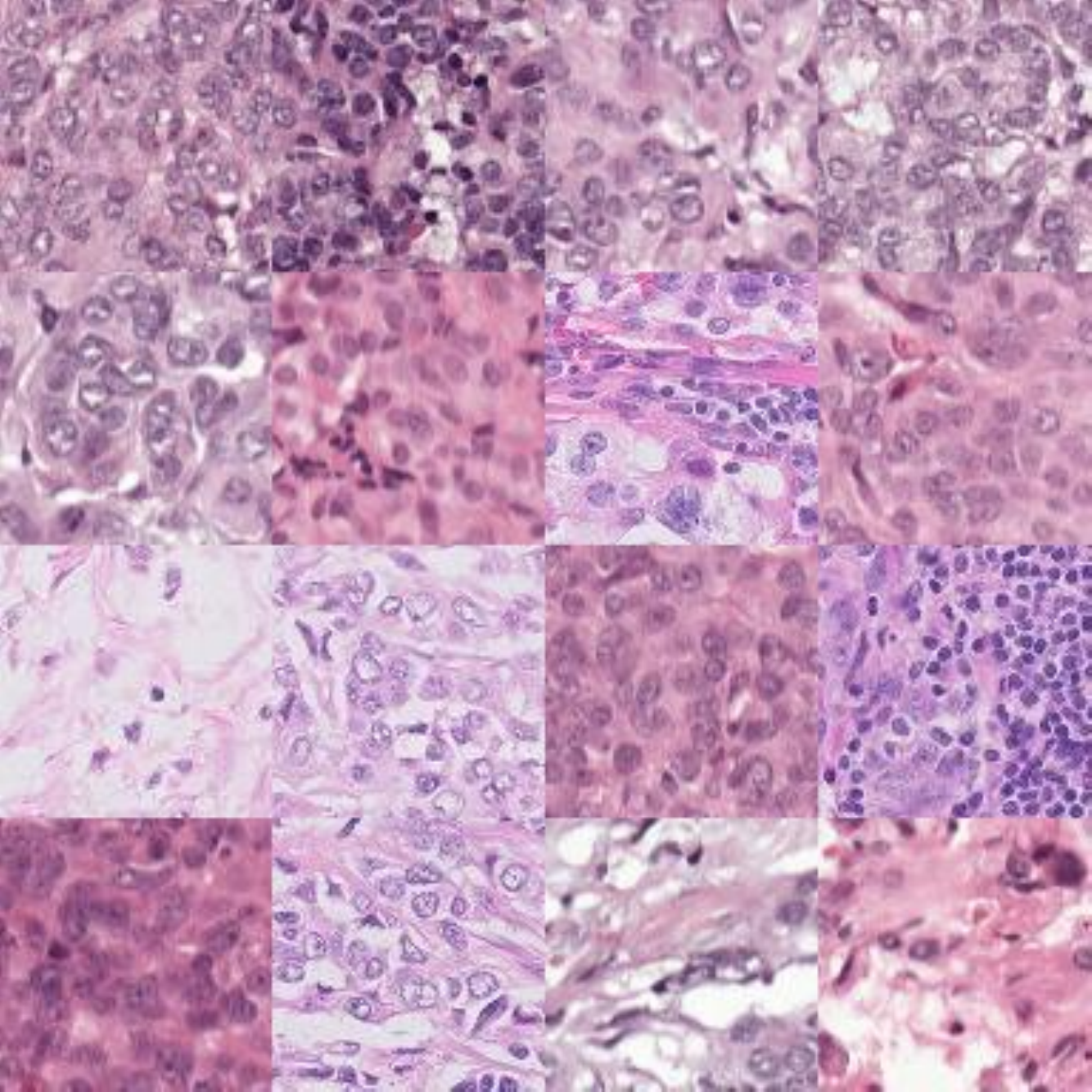}}
    \subfigure[Radiology - healthy chest X-rays]{ \includegraphics[width=0.95\linewidth]{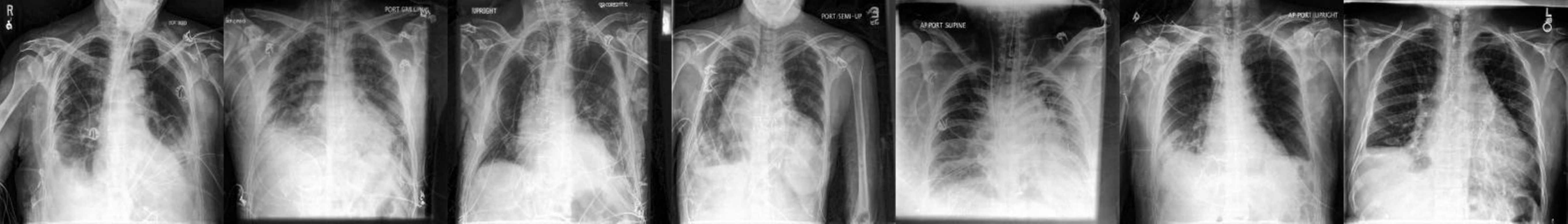}}
    \subfigure[Radiology - atelectasis, cardiomegaly, consolidation, pleural effusion, pulmonary edema]{ \includegraphics[width=0.95\linewidth]{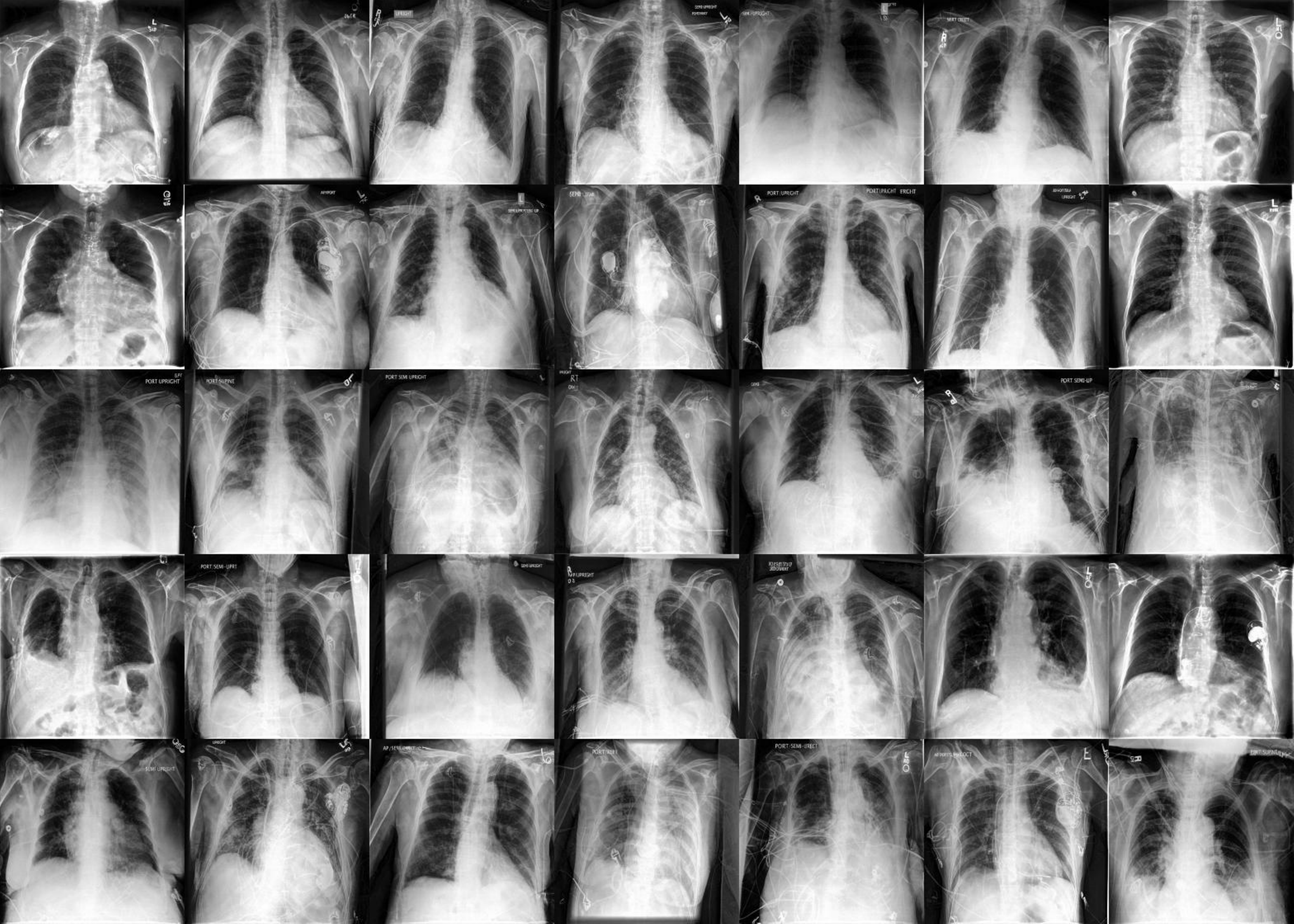}}
    \caption{Generated images in simple setting.}
    \label{fig:generated_image_clinical}
\end{figure}

\subsection{Histopathology}
\label{subsec:hist_label_eff}

\subsubsection{Generated samples}

\autoref{fig:generated_image_clinical} presents some examples of generated images by the class-conditioned diffusion models for: (a) healthy and (b) abnormal whole-slide images of histological lymph node sections.

\begin{figure}
    \centering
    \subfigure[in distribution results]{ \includegraphics[width=0.49\linewidth]{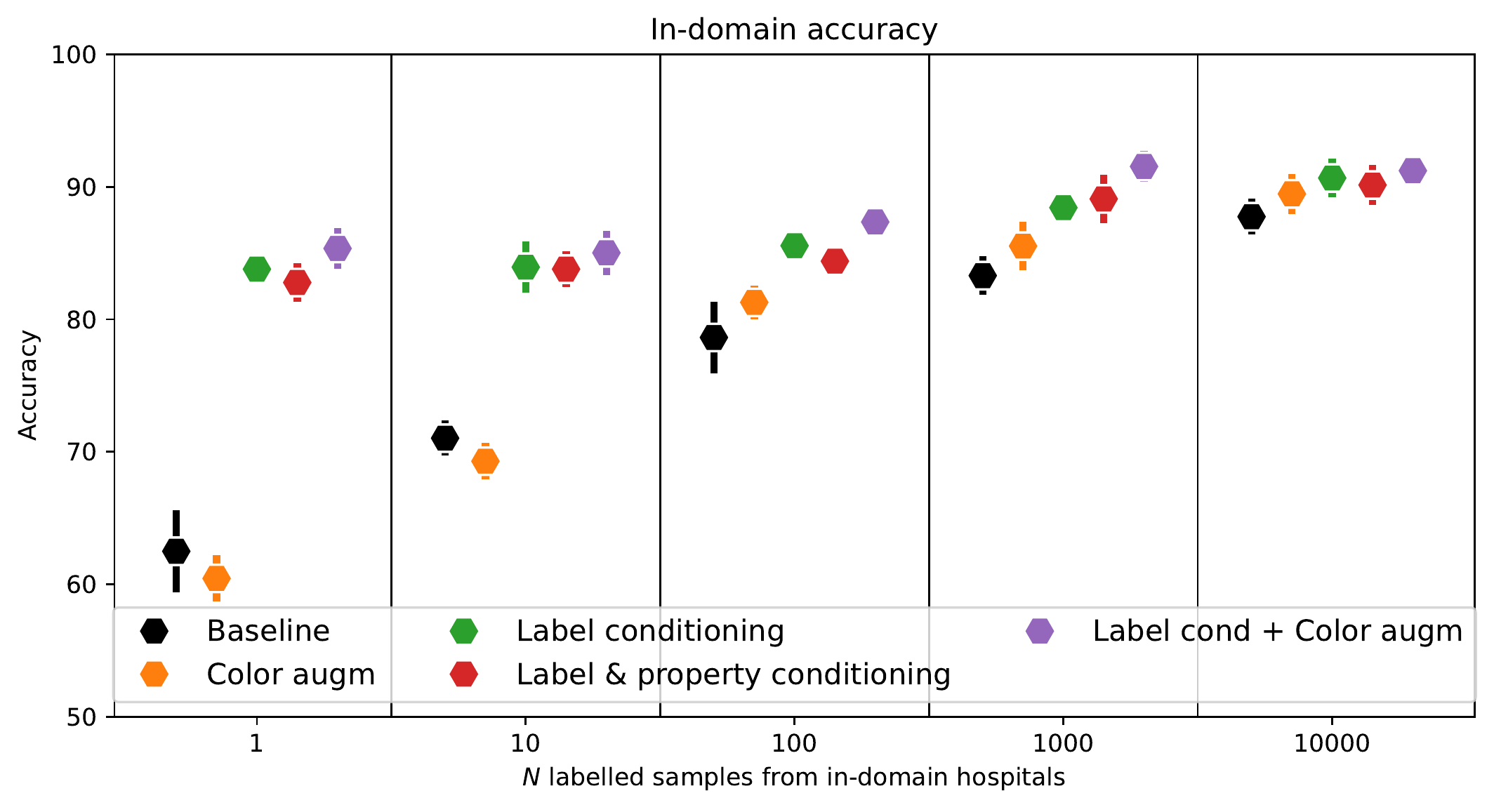}}
    \subfigure[out-of-distribution results]{ \includegraphics[width=0.49\linewidth]{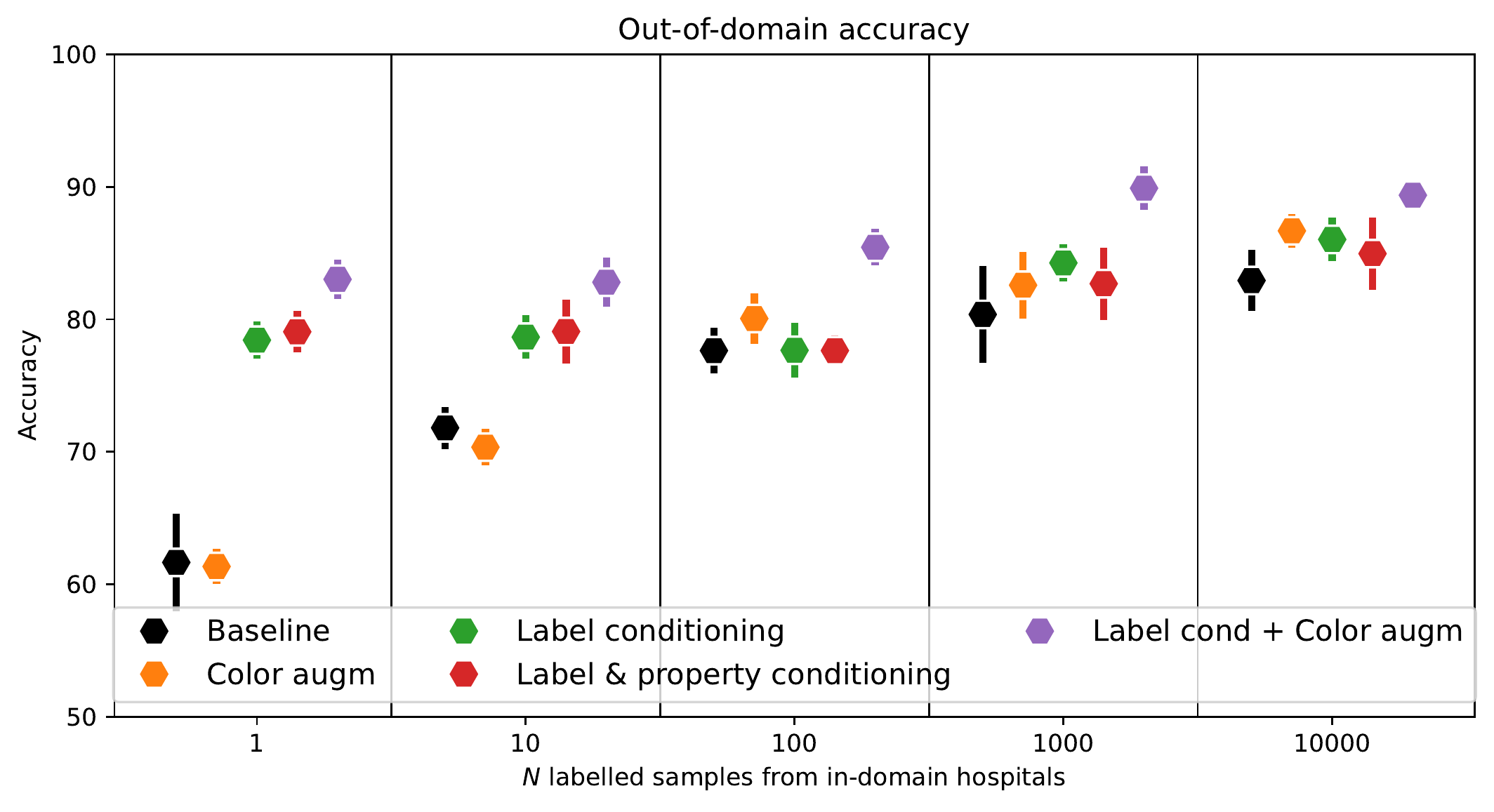}}
    \caption{Prediction accuracy for the presence of breast cancer metastases in whole-slide images of histological lymph node sections. In these experiments we vary the number of labelled samples $N$ used to train the diffusion model from in-distribution datasets on (a) in-distribution and (b) out-of-distribution hospitals. We compare the following methods: the \textit{Baseline} model that does not use any augmentations; \textit{Color augm} for a model with the same architecture that uses color augmentations; \textit{Label conditioning} for our proposed approach of a generative model conditioned on the diagnostic label; \textit{Label \& property conditioning}  for our proposed approach of a generative model conditioned on both the diagnostic label and the hospital id; \textit{Label cond + Color augm} for additionally applying color augmentations on the images generated with a diffusion model. It is worth noting that apart from \textit{Baseline} and \textit{Color augm}, all models are trained purely on synthetic images. Performance improvements are particularly striking in the low data regime (i.e., small values of $N$). Combining color augmentation with synthetic data performs best across all settings (in and out-of-distribution).}
    \label{fig:camelyon_results_label_eff}
\end{figure}

\subsubsection{Label efficiency}
The histopathology dataset is balanced, so it does not demonstrate whether synthetic data is useful in the presence of data imbalance.
In order to understand the impact of the number of labelled examples on both in- and out-of-distribution (OOD) generalization, we create different variants of the labelled training set, where we vary the number $N$ of samples from two of the training hospitals. The number of labelled examples from one hospital remains constant. For each value of $N$ we train a diffusion model using the labelled and unlabelled dataset.
We consider two settings when conditioning the diffusion model: (1) we use only the diagnostic label when available and (2) we use the diagnostic label together with the hospital id.
We subsequently sample synthetic samples from the diffusion model and train a downstream classifier that we evaluate on the held out in-distribution and out-of-distribution datasets.
We train the downstream classifier with five seeds and plot the mean and standard deviation in \autoref{fig:camelyon_results_label_eff}.
We find that using synthetic data outperforms both baselines consistently over varying $N$ in-distribution. The same holds for the low-data regime in the out-of-distribution setting. It is worth noting that using our approach can achieve the performance that the baseline model achieves with 1000 labelled samples using only 1-10 samples (yielding \textbf{3x} better label efficiency in terms of the low-data regions).
We can also perform color augmentation on top of the generated samples and find that this generalizes best overall, leading to $\sim 10\%$ absolute improvement OOD over the baseline model in the high-data regime (1,000-10,000 samples) and a striking $\sim 24\%$ in the low-data regime (1 labelled sample).

\subsection{Chest Radiology}

\subsubsection{Generated samples}
\autoref{fig:generated_image_clinical} presents examples of generated images by the class-conditioned diffusion models for: healthy chest X-rays (c) and those with thoracic conditions (d).

\subsubsection{Results per condition} 
\begin{figure}
    \centering
    \includegraphics[width=\linewidth]{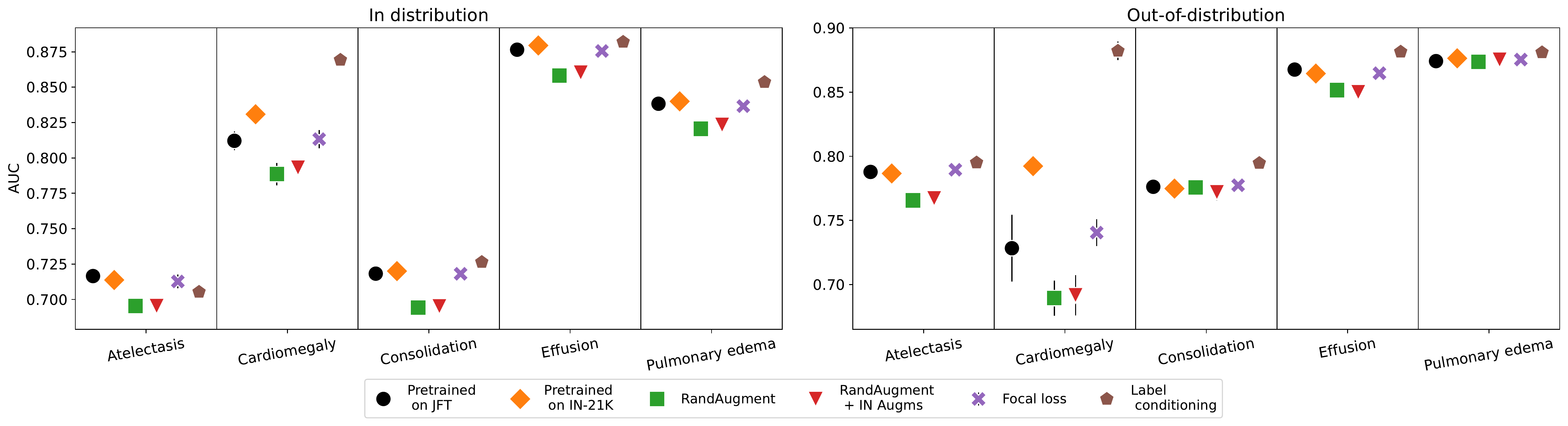}
    \caption{AUC per condition for chest-X ray in- and out-of-distribution datasets.}
    \label{fig:radiology_per_condition}
\end{figure}

We show the model's AUC values across method in- and out-of-distribution in~\autoref{fig:radiology_per_condition}. We observe that there are some conditions, i.e. \textit{cardiomegaly}, that benefit significantly from synthetic data, while others, e.g. \textit{effusion}, benefit more out-of-distribution than in-distribution. Finally for \textit{atelectasis} we observe that synthetic images are only marginally beneficial OOD. 

\subsection{Dermatology}
\subsubsection{Multiple metrics across datasets}\label{sec:supp_different_splits_derm}
\begin{figure}
    \centering
    \subfigure[Sex]{ \includegraphics[width=\linewidth]{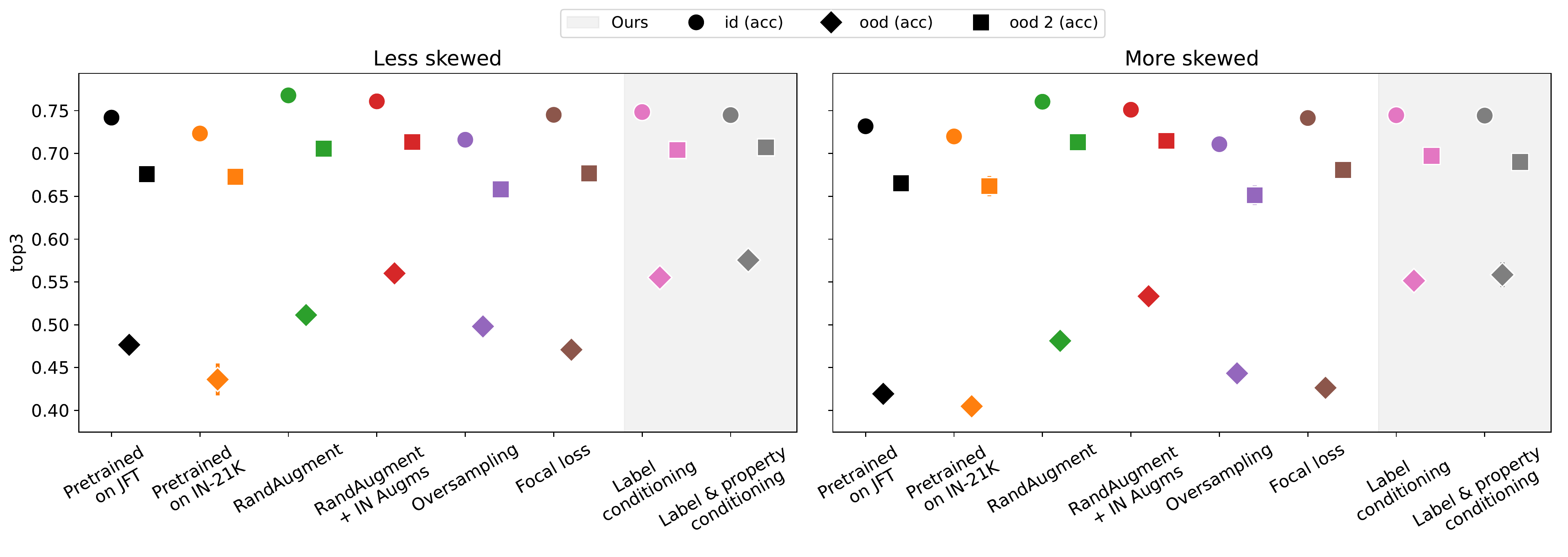}}
    \subfigure[Skintone]{ \includegraphics[width=\linewidth]{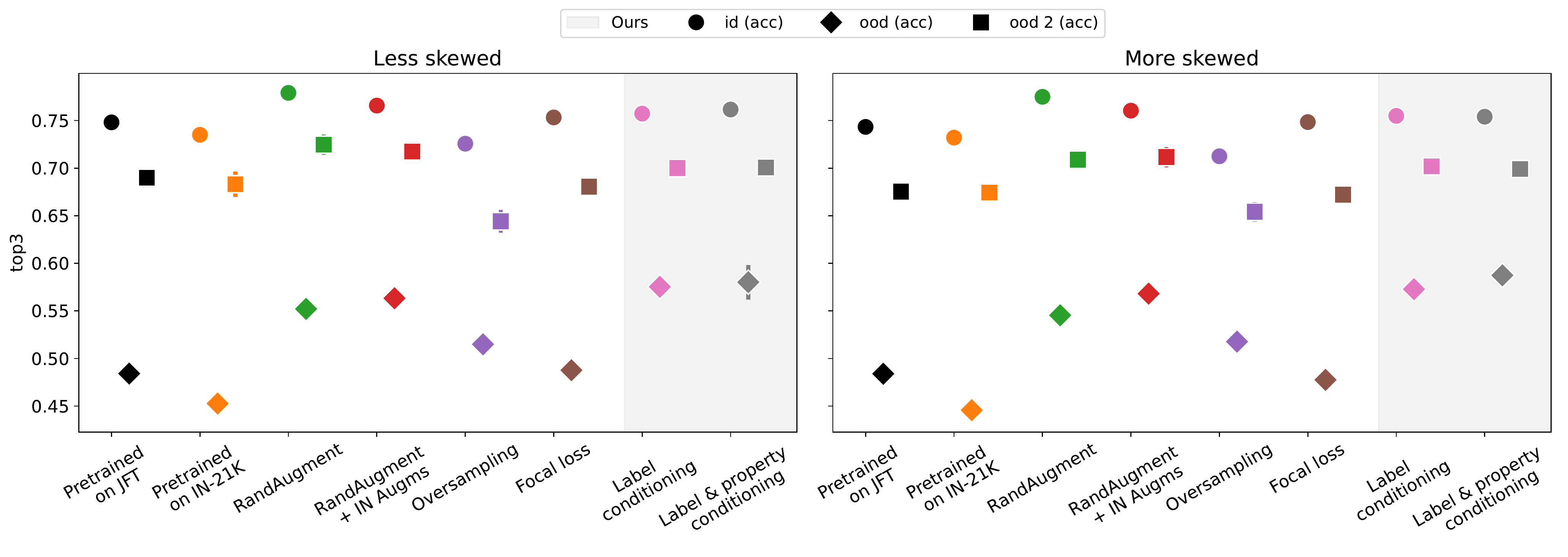}}
    \subfigure[Age]{ \includegraphics[width=\linewidth]{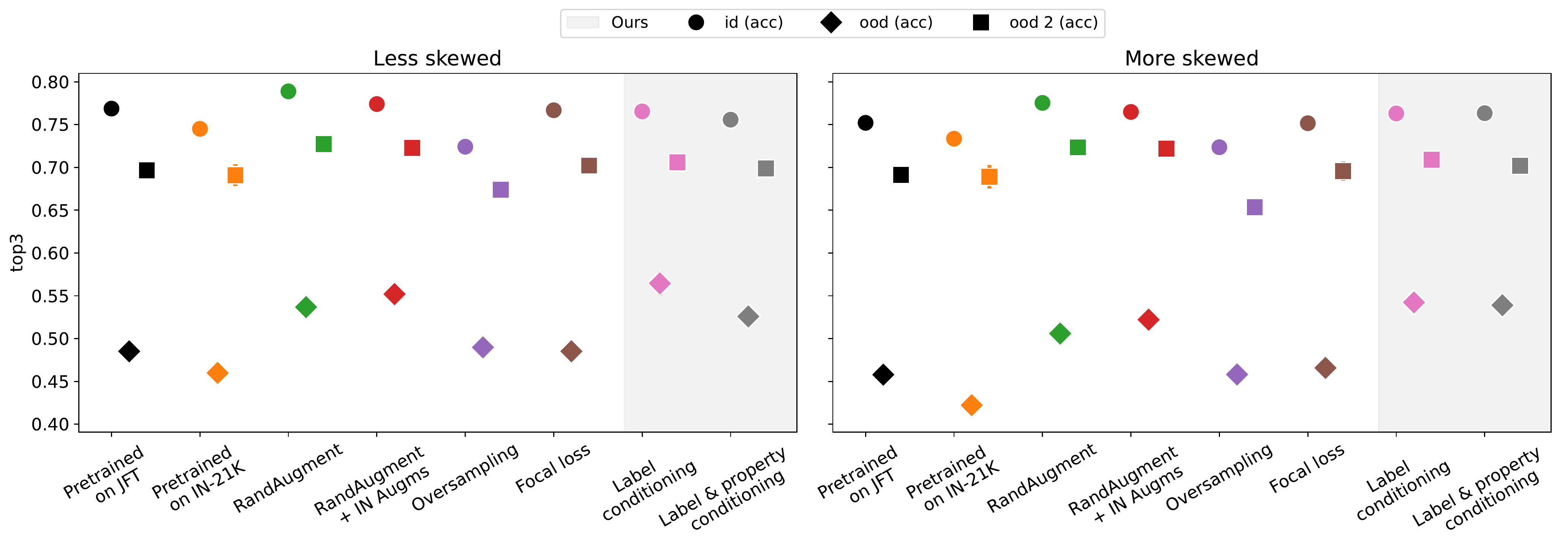}}
    \caption{Top-3 accuracy metrics on dermatology. Higher is better.}
    \label{fig:accuracy_results}
\end{figure}

\begin{figure}
    \centering
    \subfigure[Sex]{ \includegraphics[width=\linewidth]{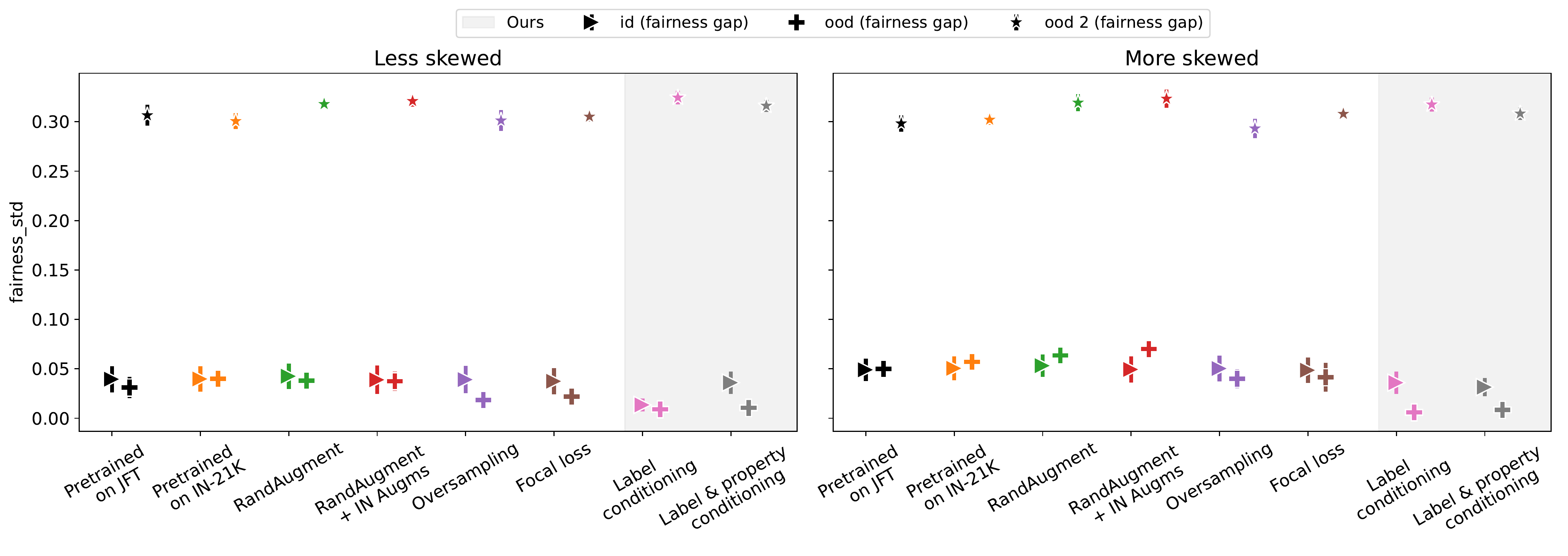}}
    \subfigure[Age]{ \includegraphics[width=\linewidth]{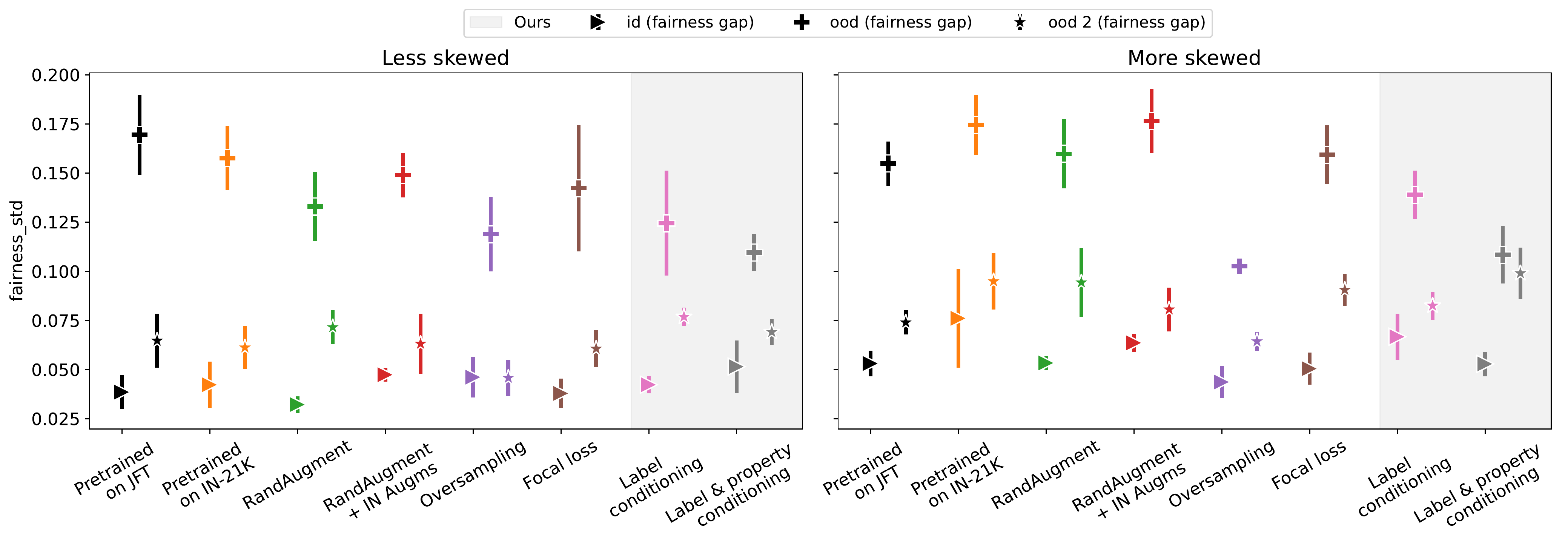}}
    \caption{Central best estimate for fairness in dermatology. Lower is better.}
    \label{fig:fairness_results}
\end{figure}

\begin{figure}
    \centering
    \subfigure[Sex]{ \includegraphics[width=\linewidth]{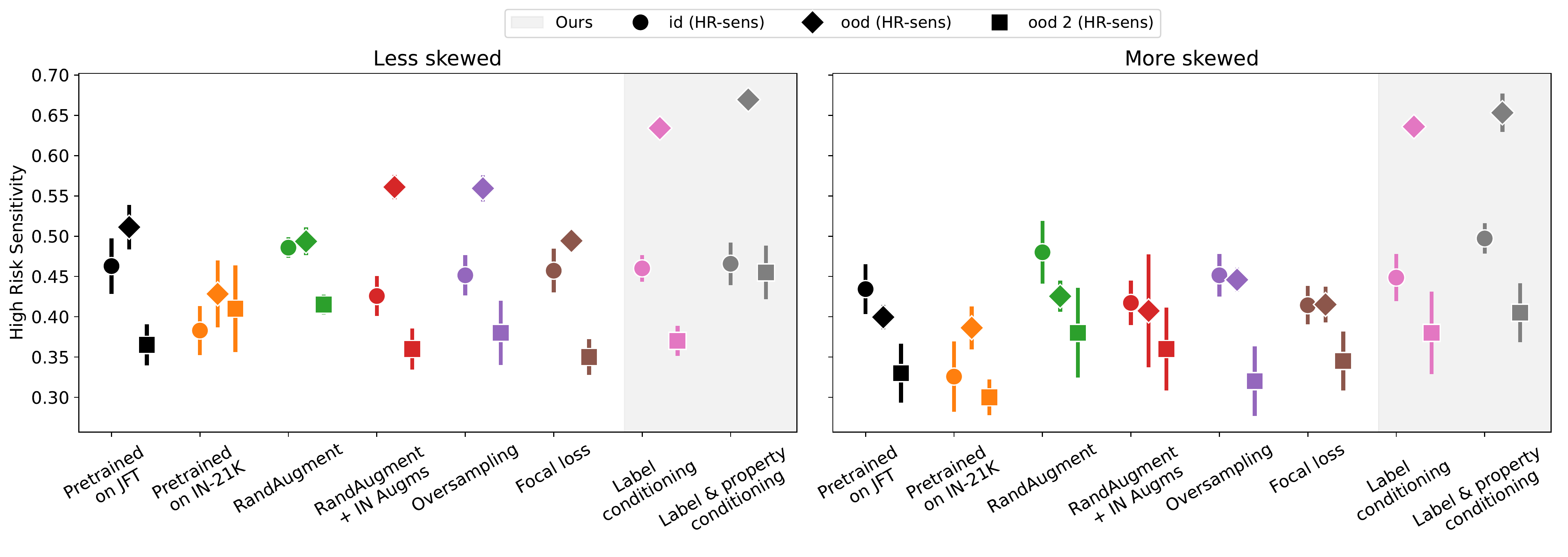}}
    \subfigure[Skintone]{ \includegraphics[width=\linewidth]{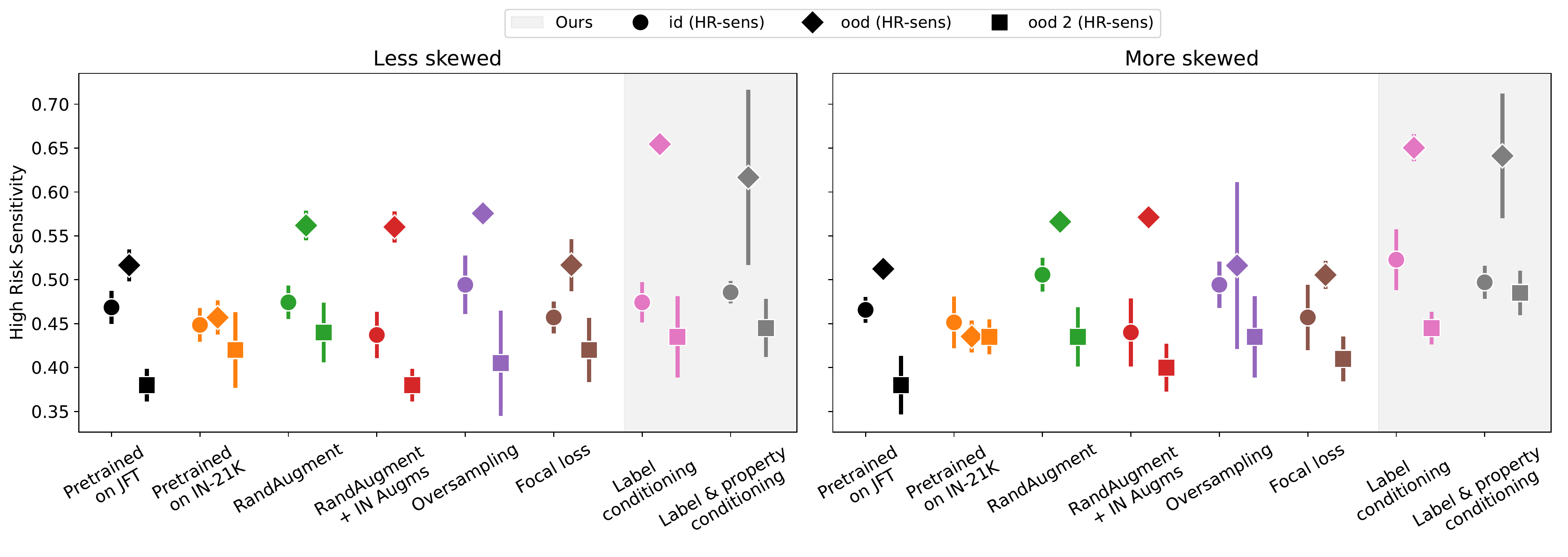}}
    \subfigure[Age]{ \includegraphics[width=\linewidth]{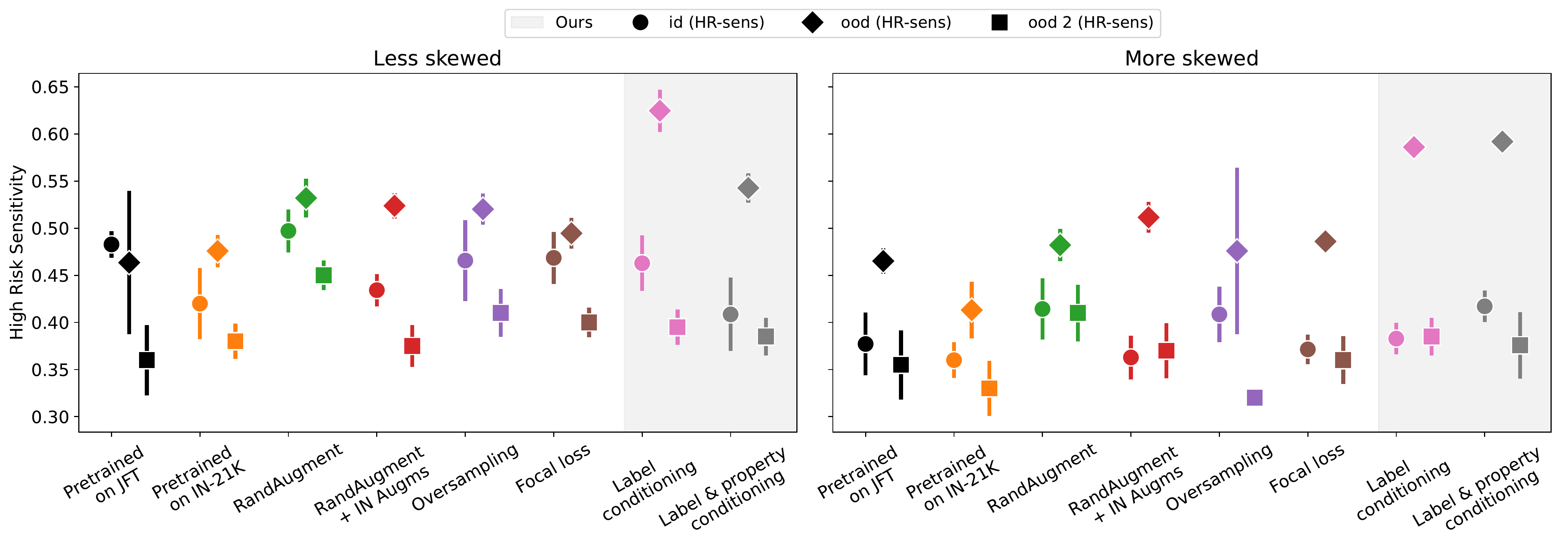}}
    \caption{High risk sensitivity for dermatology. Higher is better.}
    \label{fig:risk_results}
\end{figure}

For each sensitive attribute and distribution shift, we run all baselines with five random seeds.
We then train a diffusion model at $64\times64$ (for faster iteration) using the labelled and unlabelled data for that specific shift and combine synthetic and real data.
We consider conditioning either only on the label or on the label and sensitive attribute.
We plot both the accuracy and balanced accuracy on the in distribution datasets and out of distribution datasets in \autoref{fig:accuracy_results}.
We plot the fairness metrics in \autoref{fig:fairness_results} and the high risk sensitivity in \autoref{fig:risk_results}.
For both accuracy and fairness, we plot the normalised metric (we plot the improvement over the baseline, where we use Pretrained on JFT as the baseline).

First, we discuss results on the accuracy metrics. Across all distribution shifts and all datasets, using generated data either improves or maintains the accuracy metrics on dermatology.
In particular, generated data seems to help most on the out of distribution dataset that is has a stronger prevalence shift with respect to the training set and on the balanced accuracy metric.
We can see that using heuristic augmentation is also helpful here, in particular RandAugment consistently improves over the baseline.
The other methods (oversampling and focal loss) give minimal improvements.

Next, we investigate results on the fairness metrics in \autoref{fig:fairness_results}.
Here, we see that using heuristic augmentation leads to no consistent improvement over the baseline.
However, for sex, skintone, and age, our approach of using generated data consistently improves on or maintains the performance of the baseline model.
We find that this is true {\em even on out of distribution datasets}, but more so for those characterised by stronger shifts in comparison to the in-distribution dataset (i.e., OOD 2 is much more similar to the in-distribution dataset compared to OOD 1 where we observe the strongest improvements).
This is impressive as \cite{Schrouff22} demonstrated that improving fairness on in-distribution datasets does not guarantee performance improvements on out-of-distribution datasets.
(Note that there are no skin tone labels for the OOD datasets, so for skintone we only report results on the in-distribution dataset.)

Finally, we investigate how using synthetic data impacts high risk sensitivity in \autoref{fig:risk_results}.
In diagnostics, it is imperative to not miss someone with a high risk condition.
As a result, we investigate whether using synthetic data negatively or positively impacts the model's ability to correctly identify images of a high risk condition.
Of the $27$ classes, four of them are identified as high risk conditions: \texttt{Basal Cell Carcinoma}, \texttt{Melanoma}, \texttt{SCC/SCCIS}, and \texttt{Urticaria}.
By adding additional data we want to improve (or at least not harm) high risk sensitivity.
We investigate the high risk sensitivity on both the training dataset (held out part of it) and the two out of distribution datasets.
We find that across distribution shifts and datasets, using the additional synthetic data either maintains or improves high risk sensitivity, most notably on the most out of distribution dataset.
Moreover, synthetic data is consistently similar or better than heuristic augmentation on this metric.

We find that on dermatology, using synthetic data has a host of benefits. 
While it can to some extent improve balanced accuracy while maintaining overall accuracy, additional synthetic data can improve fairness metrics both in and out of distribution and high risk sensitivity for both in and out of distribution datasets.
This demonstrates that using synthetic data as an augmentation tool has promise for improving fairness and the diagnosis of high risk conditions.

\begin{figure}
    \centering
    \subfigure[Sex]{ \includegraphics[width=\linewidth]{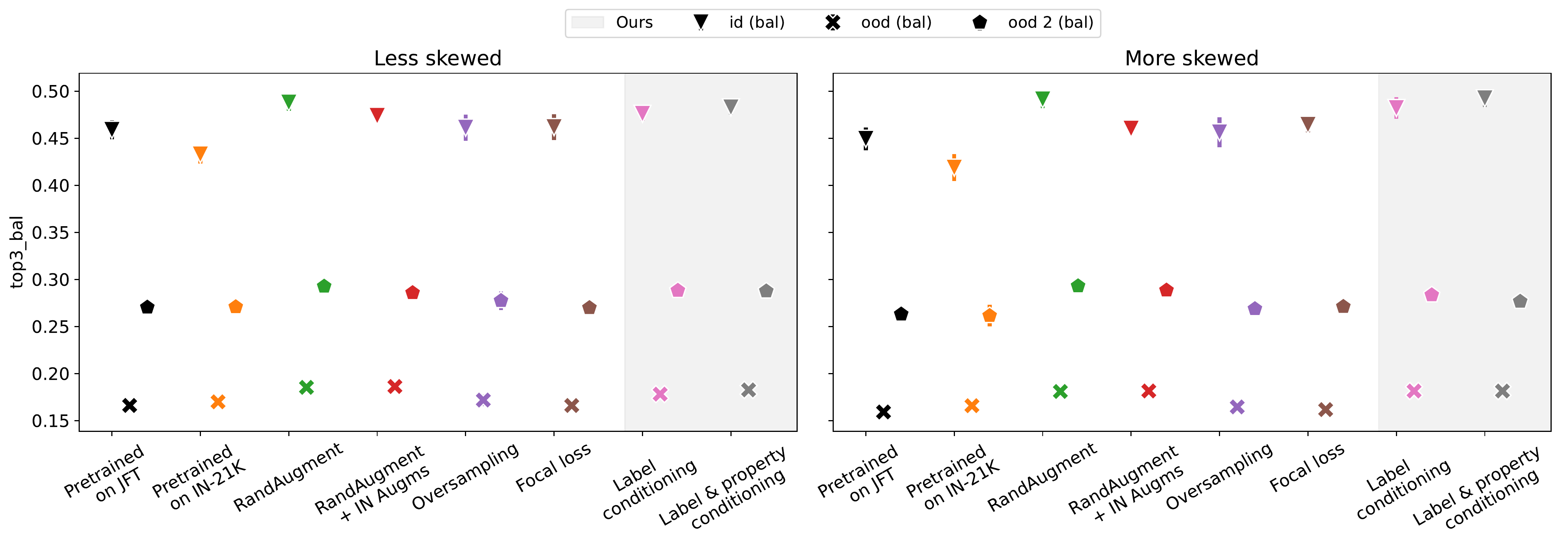}}
    \subfigure[Skintone]{ \includegraphics[width=\linewidth]{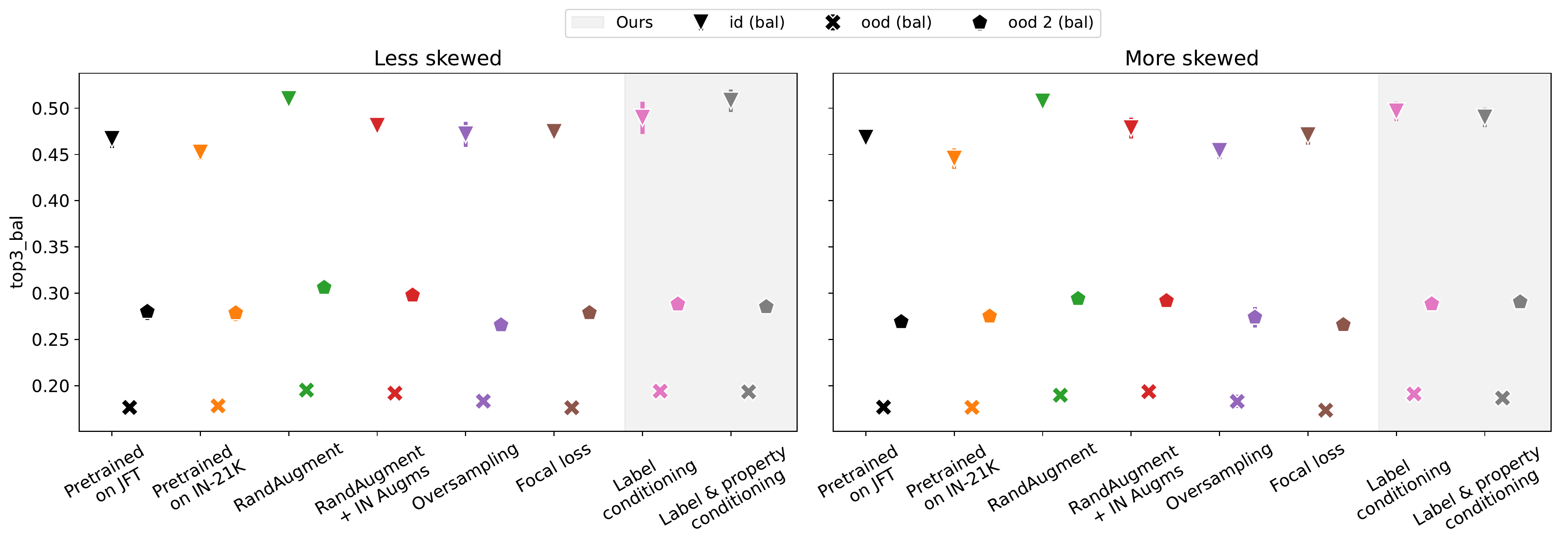}}
    \subfigure[Age]{ \includegraphics[width=\linewidth]{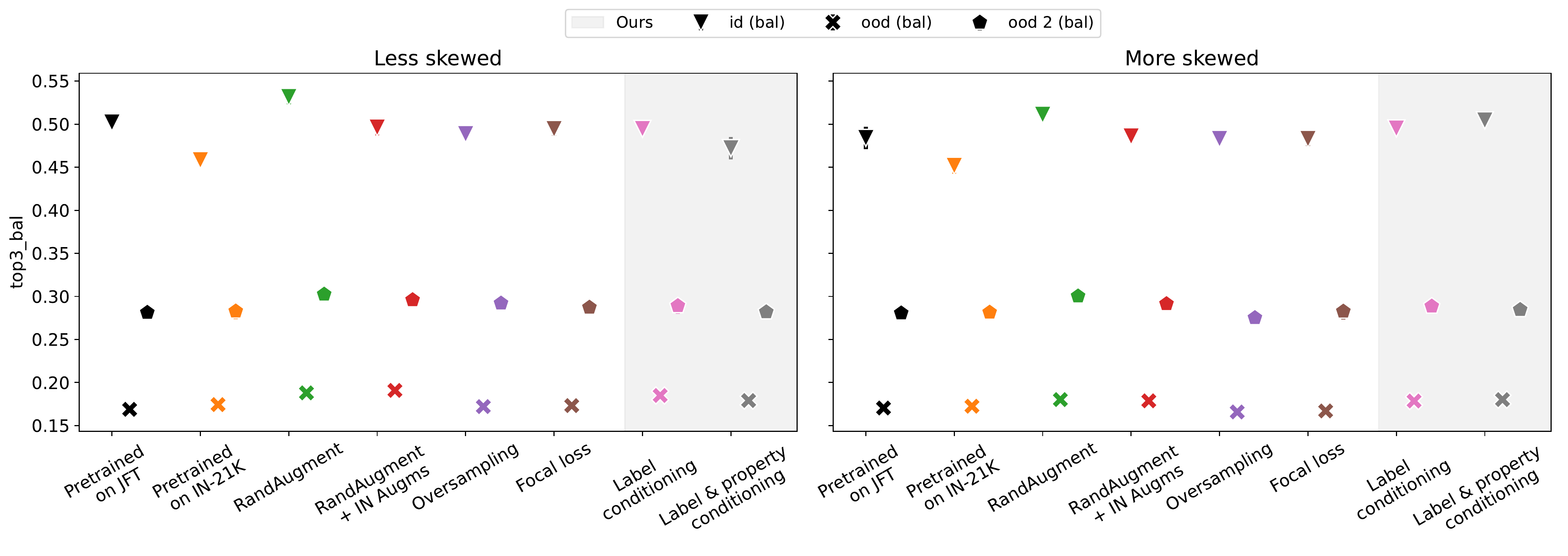}}
    \caption{Balanced top-3 accuracy for dermatology. Higher is better.}
    \label{fig:bal_accuracy_results}
\end{figure}

\subsubsection{Distribution shift estimation}
We compute domain mismatches considering the space where decisions are performed, i.e., the output of the penultimate layer of each model. We, thus, project each data point from the input space to a representation. We compute multiple estimates ($S$) of MMD between all pairs of domains using representations from the different models considering samples of size $N$. Models were trained under the same experimental conditions so that our analysis is capable of isolating the effect of data augmentation on the estimated pairwise distribution shifts. In addition to the heuristic augmentation discussed in the main text, we further include models trained with RandAugment in this analysis. All findings are summarized in~\autoref{tab:mmd_domains}.

\begin{table}[h]
\begin{tabular}{ccccc}
\hline
\textbf{Domain 1}        & \textbf{Domain 2}       & \textbf{Learned augms.} & \textbf{Heuristic augms.} & \textbf{RandAugment} \\ \hline
\textbf{ID train} & \textbf{OOD}          & 0.6279 $\pm$ 0.0823     & 0.7778 $\pm$ 0.0930  & 0.4748 $\pm$ 0.0395     \\
\textbf{ID train} & \textbf{ID eval} & 0.0121 $\pm$ 0.0047  & 0.0209 $\pm$ 0.0079 & 0.0128 $\pm$ 0.0054      \\
\textbf{ID train} & \textbf{Generated}      & 0.2472 $\pm$ 0.0190     & 0.3580 $\pm$ 0.0524 & 0.1262 $\pm$ 0.0204      \\
\textbf{Generated}       & \textbf{OOD}          & 0.6992 $\pm$ 0.0706     & 0.7682 $\pm$ 0.0915 & 0.4151 $\pm$ 0.0384      \\
\textbf{ID eval}  & \textbf{OOD}          & 0.6107 $\pm$ 0.0731     & 0.7596 $\pm$ 0.0859 &  0.4554 $\pm$ 0.0376     \\
\textbf{ID eval}  & \textbf{Generated}      & 0.2374 $\pm$ 0.0175     & 0.3052 $\pm$ 0.0472 & 0.1037 $\pm$ 0.0160     \\ \hline
\end{tabular}
\caption{Maximum mean discrepancy (MMD) values between pairs of domain distributions with learned and heuristic augmentations. \textit{ID}: in distribution; \textit{OOD}: out-of-distribution. Differences for all comparisons are statistically significant based on a Mann-Whitney U test with a significance level of 95\%.}
\label{tab:mmd_domains}
\end{table}

We observe that from the three considered augmentation schema, RandAugment yields representations that are more aligned in comparison to the learned and heuristic augmentations for all pairs of domains. We hypothesize this augmentation strategy promotes better in-distribution generalization by allowing domain-specific cues to be removed at the expense of learning spurious correlations. Evidence to support this hypothesis can be found in~\autoref{fig:accuracy_results}, which shows that models trained with RandAugment yielded improved performance in-distribution and in the OOD 2 domain, which is more similar to the training distribution than OOD I (\textit{c.f.}~\autoref{fig:eval_condition_stats}).

\subsubsection{Principal component analysis for spurious correlations}

\begin{figure}[h]
\includegraphics[width=0.6\textwidth]{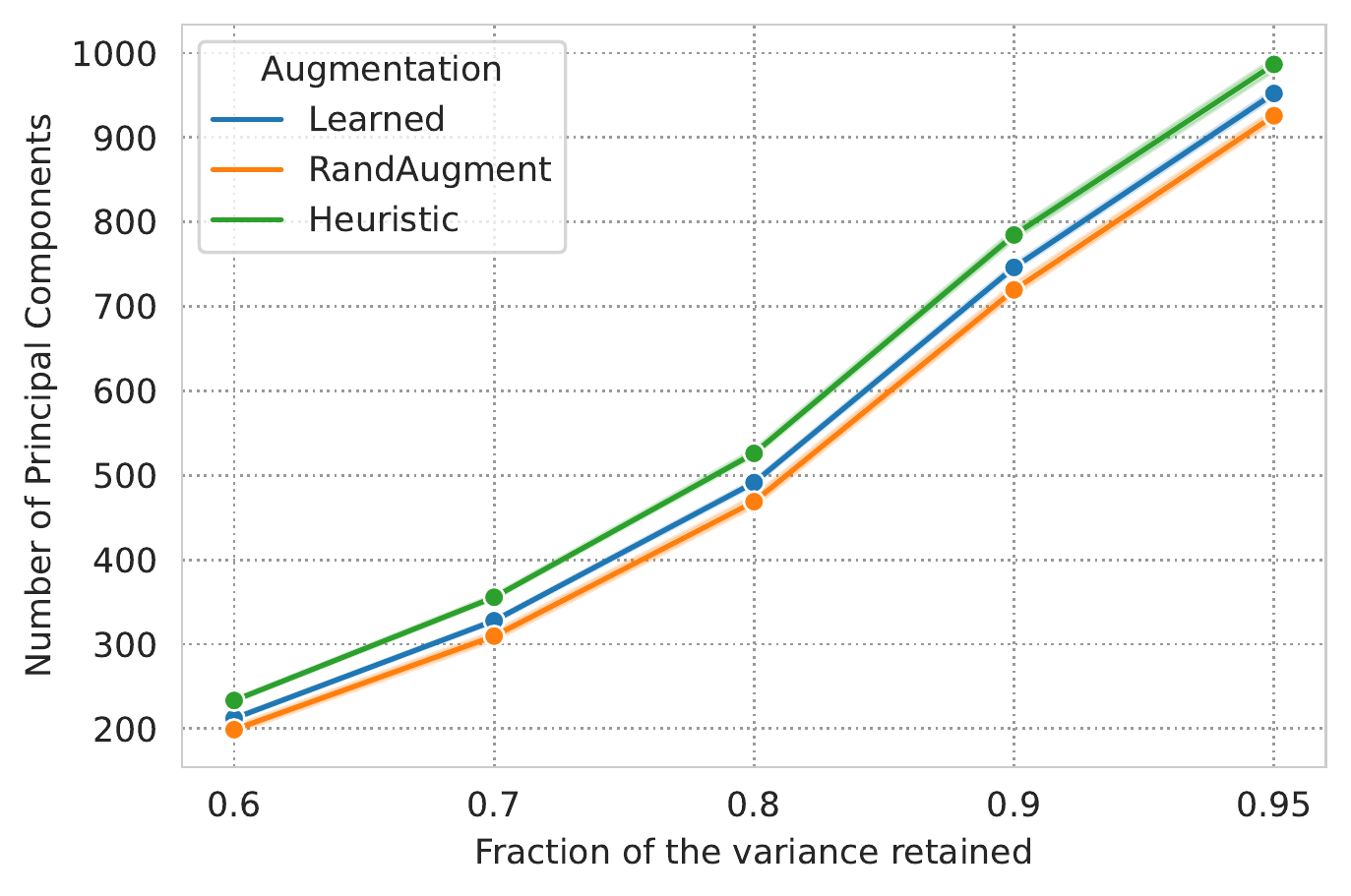}
\caption{Number of principal components (PCs) required to explain $X$ fraction of the variance after projecting real samples onto the latent space of models trained with heuristic vs. learned and RandAugment augmentations.}
\label{fig:pca_analysis}
\end{figure}

In order to further compare the effect of different augmentation schemes on the features learned by the downstream classifier and investigate why learned augmentations promote better OOD generalization, we design an experiment to account for the simplicity bias \citep{Shah2020pitfalls} within features learned with each augmentation strategy. This phenomenon is linked to the underlying mechanisms that lead models models to rely on spurious correlations to make predictions \citep{Shah2020pitfalls}. In practice, we project $N$ randomly sampled instances from each dataset to the feature space learned by each model and apply the Principal Component Analysis algorithm~\citep{abdi2010principal}. We then adopt the number of principal components required to represent different fractions of the variance across all instances projected to the feature spaces as a proxy measure for simplicity in the decision space learned by a model. These feature spaces are induced by models obtained with heuristic and learned augmentations. In~\autoref{fig:pca_analysis}, we show the average number of components required to retain $\{0.60, 0.70, 0.80, 0.90, 0.95\}$ of the variance of the projected data obtained across five different model initializations for RandAugment, learned, and heuristic augmentations. We observe that for a fixed dataset, features from models trained with learned augmentations and RandAugment require fewer principal components to retain the same fraction of variance for all considered values, indicating that compressing features via removing domain specific cues (\textit{c.f.} ~\autoref{tab:mmd_domains}) yields improved performance. Notably, our findings suggest that learned augmentations enforce models to capture features that strike a better balance between simplicity and predictive power out-of-distribution, i.e. they are simultaneously not overly complex, therefore generalizing well in-distribution, and not too simplistic as in the case of RandAugment, which yield models that generalize \emph{too well} in the vicinity of the training distribution via relying on spurious correlations.


\subsubsection{Individuals underserved by models} 

\begin{figure}[t]
    \centering
    \subfigure[In-distribution high-risk females]{ \includegraphics[width=0.49\linewidth]{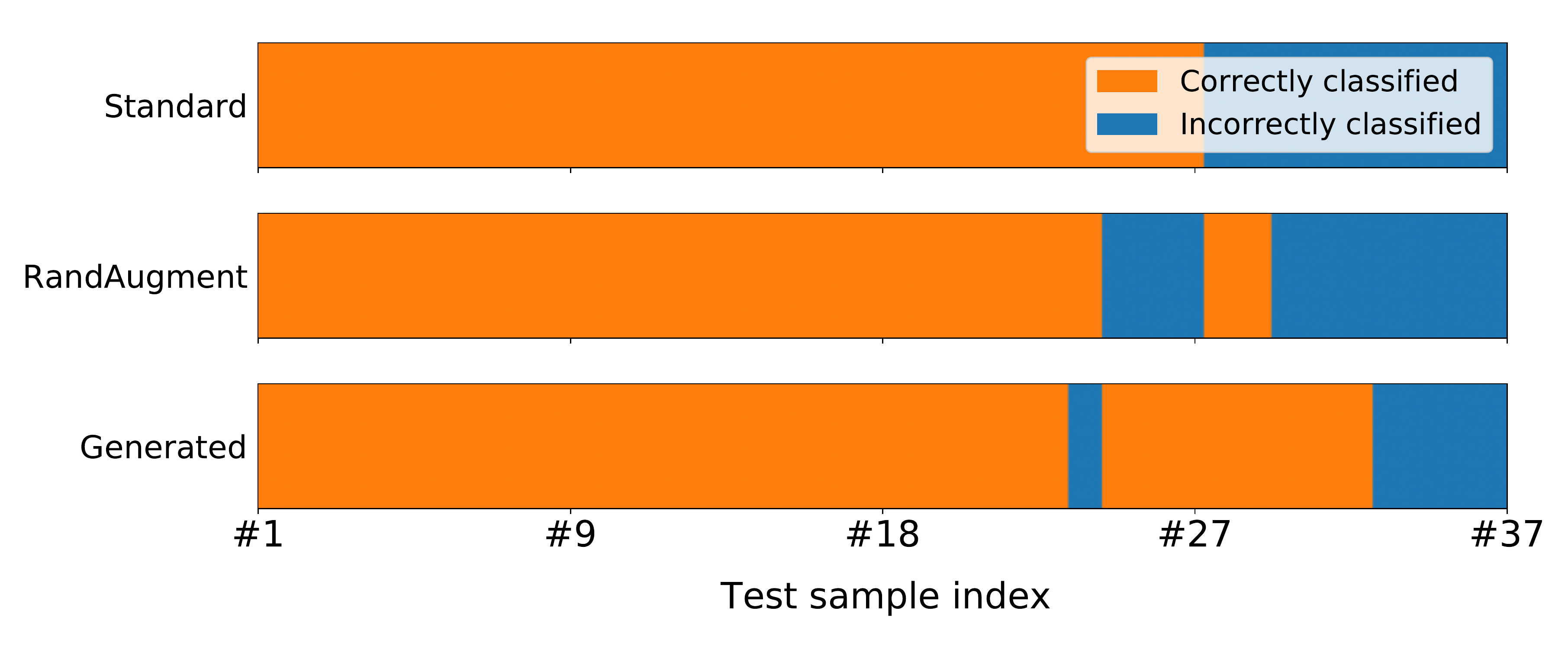}}
    \subfigure[In-distribution high-risk males]{ \includegraphics[width=0.49\linewidth]{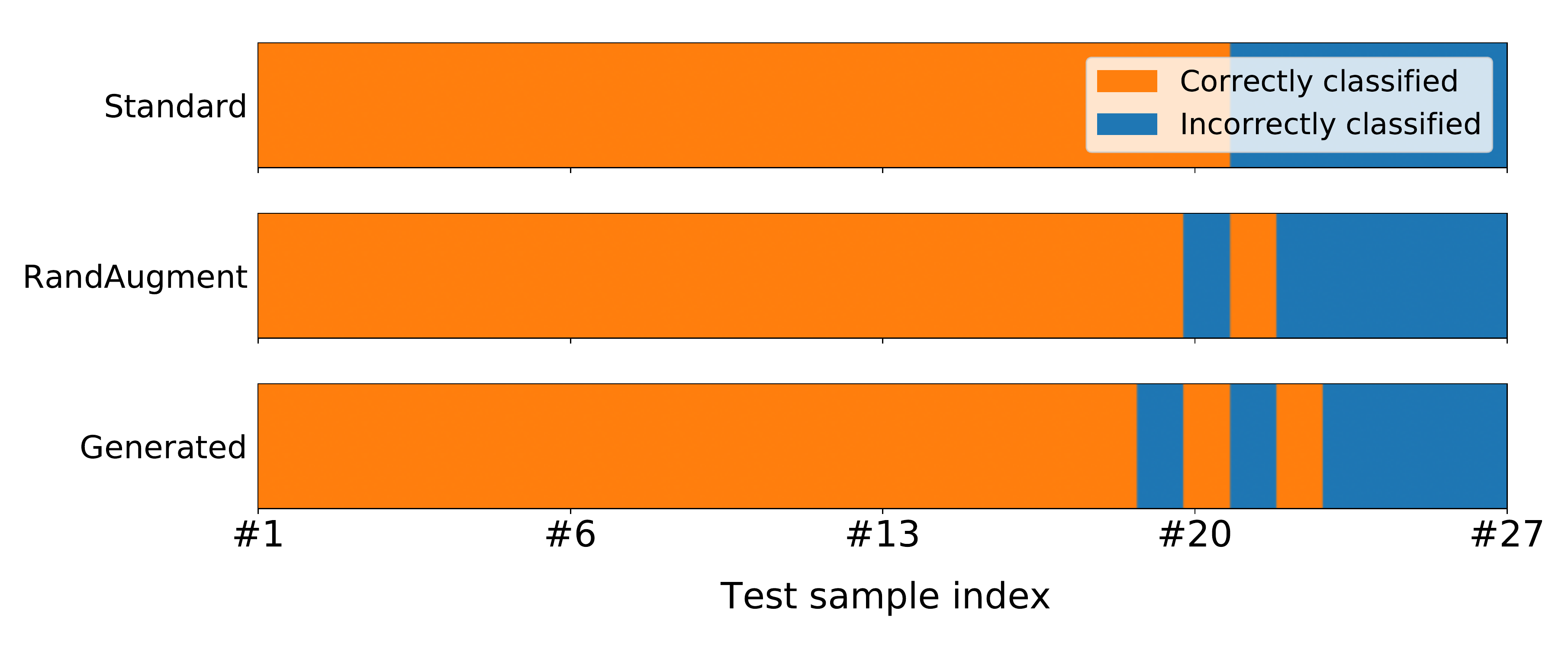}}
    \subfigure[Out-of-distribution high-risk females]{ \includegraphics[width=0.49\linewidth]{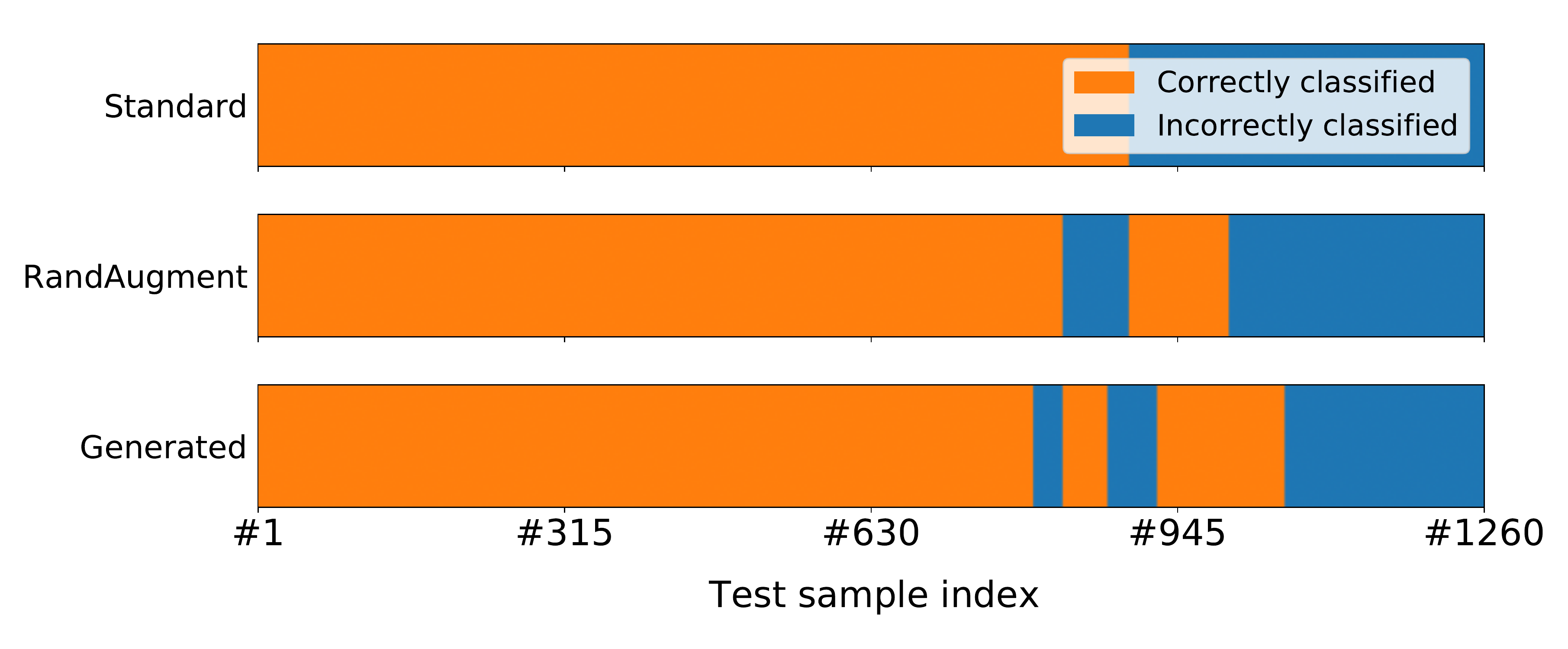}}
    \subfigure[Out-of-distribution high-risk males]{ \includegraphics[width=0.49\linewidth]{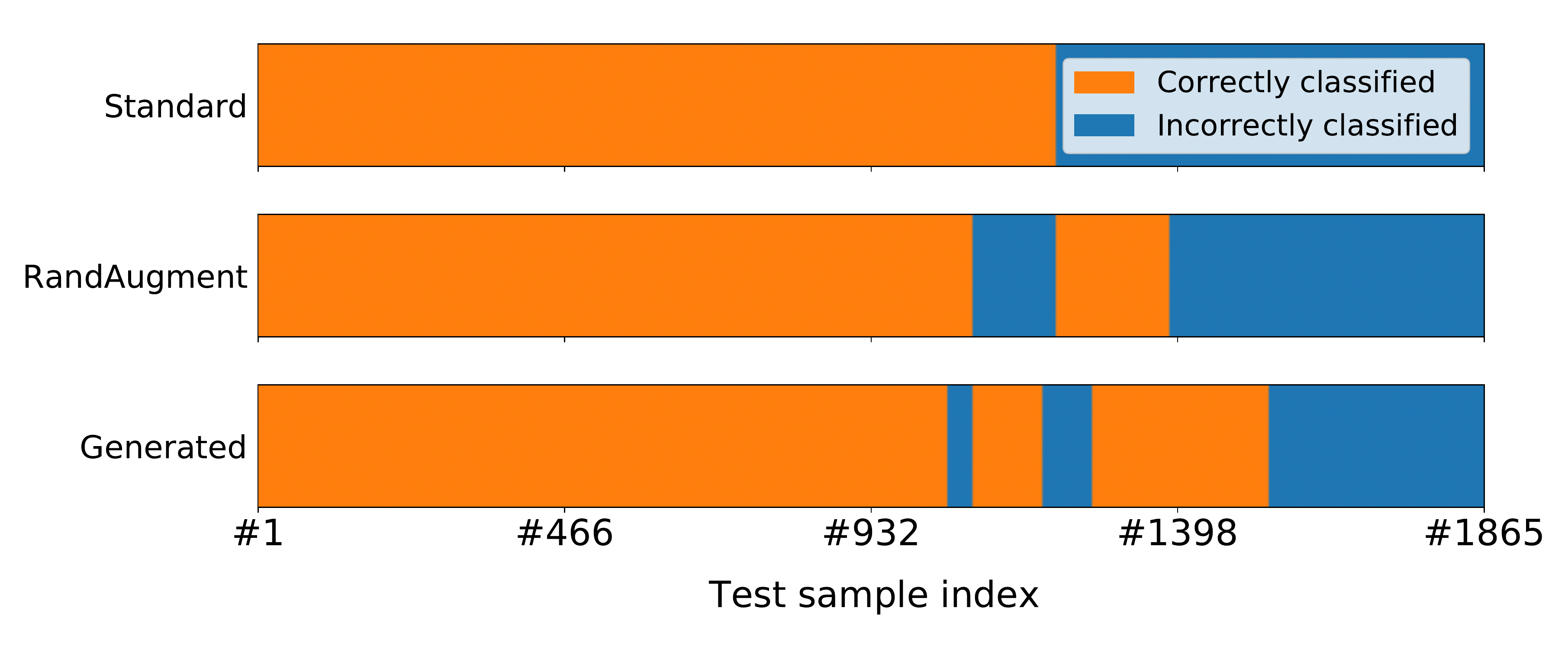}}
    \caption{Mis-classification analysis on individuals with high-risk conditions depending on their sex attribute. We investigate the impact of a skewed training dataset with respect to the underrepresented sex on correctly detecting high-risk conditions for that subgroup in- and out-of-distribution. In each plot, we show the test sample index on the x-axis to investigate which individuals are consistently misclassified by all approaches. Sample indices are re-ordered such as to form contiguous blocks of correctly and incorrectly classified samples. It is worth noting that the source domain is more skewed towards females (see~\autoref{tab:sens_attr_splits_eval} for details).}
    \label{fig:individuals_high_risk}
\end{figure}

Inspired by a recent study by~\cite{Bommasani2022picking} looking at how often the same individuals are underserved by machine learning models that have been trained on the same data, we investigate whether the same individuals with high-risk conditions are consistently misclassified.  In~\autoref{fig:individuals_high_risk} we illustrate for all sample IDs across the in- and out-of-distribution evaluation datasets whether there are particular individuals within each demographic subgroup (male or female) that benefited more from the generated data than from other augmentation techniques. For each of the three setups (1) standard ImageNet augmentations, (2) RandAugment and (3) generated data, we perform 5 training runs and consider a test sample as incorrectly classified for a setup if it has been consistently misclassified by its 5 trained models. For better comparison, we re-order the sample indices such as to form contiguous blocks of correctly and incorrectly classified samples. While a majority of the individual predictions are the same between setups, we still note that each setup enables some samples to be correctly classified which the other setups cannot. Particularly, in panels (a) and (d) training with generated data significantly reduces the number of consistently misclassified samples compared to standard ImageNet augmentations or RandAugment. We see that even though the training dataset is more skewed towards females, out-of-distribution males with high-risk conditions in panel (d) were more often correctly classified for a model trained with generated data. Hence, using generated data reduces the number of underserved individuals compared to standard augmentation techniques which only apply basic transforms to the original data. Finally, we observe that these training setups are complementary as each of them has its own set of well classified samples, so this could open new research directions for model ensembling to create new models which would benefit from this diversity in individual predictions.

\subsubsection{Discussion on sampling schemes}
\label{app:samplingschemes}

In this section, we further discuss and validate the choice of sampling function.
First, we motivate different sampling functions within the vanilla loss function (\autoref{eq:loss}) and then evaluate their performance on the downstream classifier to validate the setting used in our main experiments. In particular, we consider variations that aim to reduce the discrepancy between the training data distribution $p_{\texttt{train}}$ and the target ``fair'' distribution $p_{f}$ by weighting the losses of training examples. Specifically, the extended loss function is given by
\begin{equation}
\label{eq:loss_reweighted}
    R_{\alpha} (\theta) := \alpha \sE_{(\vx, \va, y) \sim p_{\texttt{train}}} \left[ w_1( y, \va) \cdot L \left( f_\theta(\vx), \va, y\right) \right] + (1 - \alpha) \sE_{(\vx, \va, y) \sim \hat{p}} \left[ w_2( y, \va) \cdot L\left(f_\theta(\vx), \va, y \right) \right] 
\end{equation}
where the weighting functions $w_1$ and $w_2$ adjust the contributions of the real and synthetic examples, respectively. As previously defined, $p_{\texttt{train}}(\vx, \va, y)$ denotes the training distribution over the image, class and attributes, $\hat{p}(\vx, \va, y)$ denotes the approximation of the target fair distribution $p_{f}(\vx, \va, y)$ based on the generative model, and the steering probability $\alpha \in [0, 1]$ determines the proportion of real data. Let us unpack the loss function term by term. 

First, the weighting function $w_1(y, \va)$ reweighs the losses of the real training examples by the ratio of distributions over sensitive attributes conditioned on the skin condition between the training and the target fair distribution:
\begin{equation}
w_1(y, \va):=\left(\frac{p_{f}(\va\vert y)}{p_{\texttt{train}}(\va \vert y)}\right)^{l}
\end{equation}
where $l \geq 0$ (``equality level'') controls the degree of penalty. When $l = 1.0$, the first term in ~\autoref{eq:loss_reweighted} equates to the importance weighting \citep{horvitz1952generalization,shimodaira2000improving} with respect to the target fair distribution $p_{f}$. Intuitively, because we desire the sensitive attributes to be uniformly distributed per condition in the fair distribution i.e., $p_{f}(\va\vert y)$ is constant, this weighting function $w_1$ essentially upweights the losses for examples from under-represented subgroups in the training data (i.e., small $p_{\texttt{train}}(\va|y)$), thereby attenuating the discrepancy between $p_{\texttt{train}}$ and $p_{f}$. On the other hand, when $l = 0.0$, we recover the default unweighted loss. Introducing this extra hyper-parameter allows us to interpolate between the default loss function and its importance weighted version by choosing $0.0 < l < 1.0$ and also ``extrapolate'' (i.e., over-penalise the underrepresented groups) by choosing $l > 1.0$. 

Second, we introduce the function $w_2(y, \va)$ that acts as a filtering mechanism for selecting which generated samples to include in the training loss. Our dermatology dataset is dominated by the most prevalent 4 conditions as illustrated in Figure~\ref{fig:eval_condition_stats}, and we suspect that the benefits of the synthetic data are most pronounced for the remaining less frequent skin conditions. To test this hypothesis, we define $w_2(y, \va)$ as the indicator function that yields 0.0 if the generated samples belong to the dominant 4 conditions, and otherwise outputs 1.0. 

We now investigate the effects of the weighting functions $w_1$ and $w_2$ on the overall and fairness performance of the downstream classifiers. To generate synthetic data, as in Sec.~\ref{sec:supp_different_splits_derm}, we use the same diffusion model trained on the dermatology dataset at $64\times64$ resolution for all settings. We use a lower resolution than in the main results for faster iteration, but we expect results to hold at higher resolution too. 

\begin{figure}[t]
\includegraphics[width=0.7\textwidth]{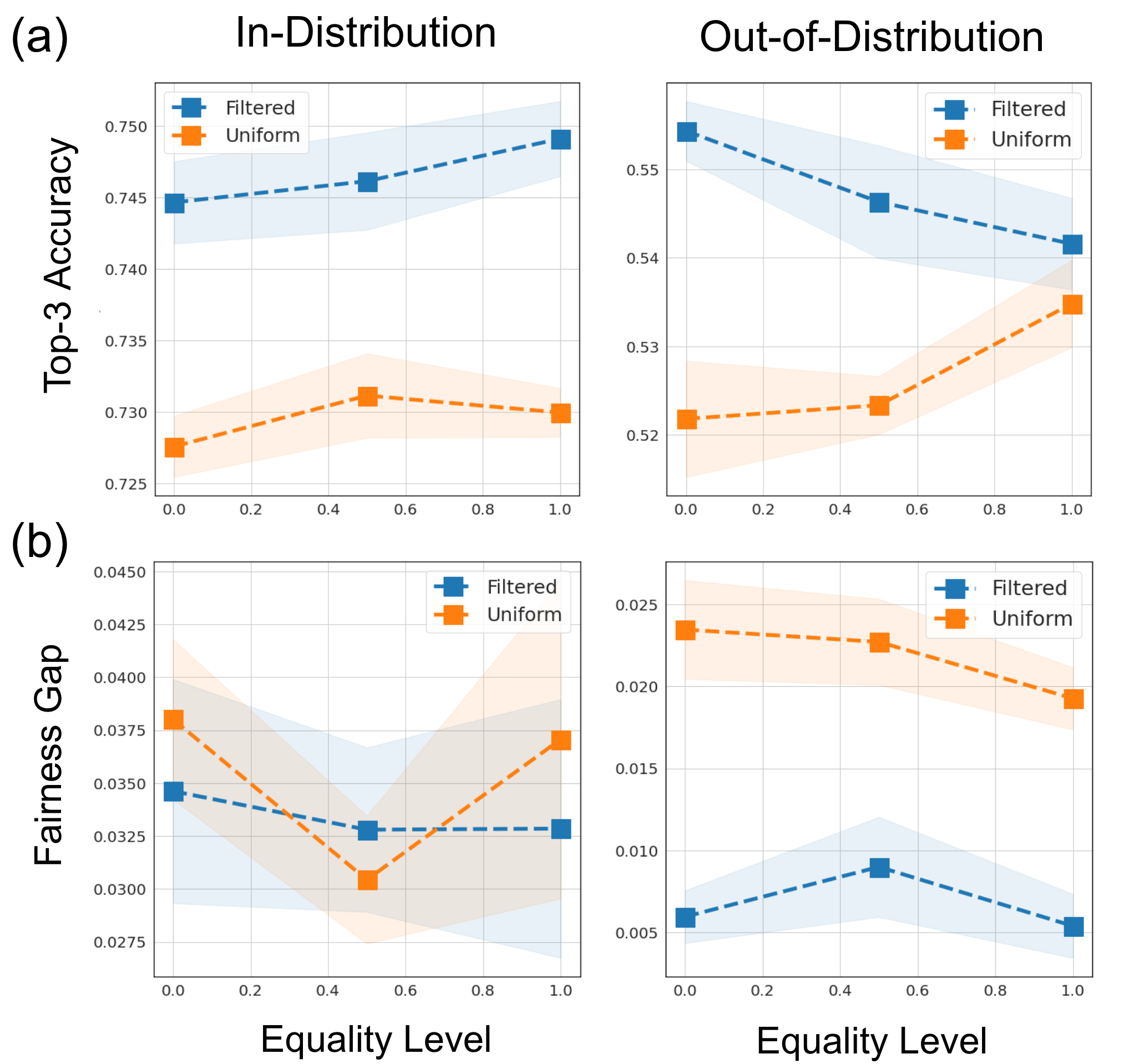}
\caption{Comparison of different training schemes on our dermatology dataset in terms of (a) the overall top-3 accuracy and (b) the fairness gap with respect to sex. The fairness gap here is measured as the difference in top-3 accuracy between the male and the female subgroups. In particular, we measure the effects of the weighting terms $w_1$ and $w_2$ in both in-distribution and out-of-distribution settings. To measure the effects of the weighting function $w_1$ for the real data, we vary its strength by changing the equality level, $l$ that is on the x-axis (note that when it's zero, it corresponds to no $w_1$ term, which is the default setting). Secondly, we compare the scenarios with (``\textit{filtered}'') and without (``\textit{uniform}'') the filtering function $w_2$ for the synthetic data. The mean and standard deviations of the metrics are calculated over five runs with different random seeds. 
}
\label{fig:comparison_sampling_schemes}
\end{figure}

Figure~\ref{fig:comparison_sampling_schemes} shows that, in the absence of any filtering mechanism $w_2$ (the orange lines), the inclusion of importance weights (equality level = 0.5, 1.0) consistently increases the overall top-3 accuracy and reduces the fairness gap with respect to the sex attribute in both in-distribution and out-of-distribution test datasets. While the improvement in fairness scores through penalisation of under-represented subgroups is expected, the concurrent gains in the overall accuracy comes as a surprise. One possible explanation is that sex and some skin conditions are spuriously correlated, and thus reducing this correlation through importance weighting helps the model learn more generalisable features. 

However, when combined with the filtering function $w_2$ (the blue lines in Figure~\ref{fig:comparison_sampling_schemes}), the effects of the importance weighting are more complicated. As before, the fairness gap narrows in most cases. On the other hand, the overall top-3 accuracy only improves in the in-distribution setting but worsens in the out-of-distribution setting. This suggests a subtle interaction between the importance weighting and the filtering. 

Finally, we find that the filtering function $w_2$ (the blue lines) increases the overall top-3 accuracy and lessens the fairness gap in the majority of settings. This result shows that the benefits of the synthetic data in terms of both overall performance and fairness are largely concentrated on the under-represented classes. Additionally, the improvements due to $w_2$ alone are generally considerably larger than those due to the importance weighting term $w_1$ on its own. Notably, the best top-3 OOD accuracy is attained with the filtering function alone. 

Based on the above analysis, in our experiments, we decided to use the loss function with the filtering function $w_2$ on the synthetic data but without the importance weighting term $w_1$ on the real ones for the dermatology datasets. This is because the filtering function $w_2$ alone confers considerable improvements in both overall and fairness scores. Although the fairness score can be marginally improved by adding the importance weighting $w_1$, this comes at the expense of the overall OOD accuracy, which is an important metric in our applications. For the other datasets, as they are more balanced over the label distribution (and histopathology is balanced), we do not use either term.

\end{document}